\documentclass[letterpaper,twocolumn,10pt]{article}
\usepackage{zhanggroup}

\usepackage{hyperref}
\usepackage{xurl}
\usepackage{tikz}
\usepackage{xspace} 
\usepackage{amsmath}
\usepackage{subcaption}
\usepackage{booktabs}
\usepackage{multirow}
\usepackage{graphicx} 
\usepackage{caption,subcaption}
\usepackage{makecell}
\usepackage[absolute]{textpos}
\usepackage{dsfont}
\usepackage[linesnumbered,ruled]{algorithm2e}
\usepackage{xcolor}
\definecolor{myblue}{HTML}{3B6790}
\definecolor{myred}{HTML}{A94A4A}
\definecolor{mygreen}{HTML}{889E73}
\hypersetup{
    colorlinks=true,
    linkcolor=myblue,
    citecolor=myred,
    urlcolor=mygreen
}

\usepackage{enumitem}
\setlist[itemize]{leftmargin=*}

\newcommand{\refappendix}[1]{\hyperref[#1]{Appendix~\ref*{#1}}}
\newcommand{\mypara}[1]{\noindent{\bf {#1}.} \xspace}
\newcommand{\dtest}{$\mathcal{D}_{\textit{test}}$\xspace}
\newcommand{\dt}{$\mathcal{D}_{\textit{target}}$\xspace}
\newcommand{\dtr}{$\mathcal{D}_{\textit{target}}^\textit{real}$\xspace}
\newcommand{\dts}{$\mathcal{D}_{\textit{target}}^\textit{syn}$\xspace}
\newcommand{\dta}{$\mathcal{D}_{\textit{target}}^\textit{aux}$\xspace}
\newcommand{\dr}{$\mathcal{D}_{\textit{ref}}$\xspace}
\newcommand{\drr}{$\mathcal{D}_{\textit{ref}}^{\textit{real}}$\xspace}
\newcommand{\drs}{$\mathcal{D}_{\textit{ref}}^{\textit{syn}}$\xspace}
\newcommand{\dra}{$\mathcal{D}_{\textit{ref}}^\textit{aux}$\xspace}
\newcommand{\tone}{$\mathcal{T}_{\mathcal{C}_1}$\xspace}
\newcommand{\ttwo}{$\mathcal{T}_{\mathcal{C}_2}$\xspace}
\newcommand{\tthree}{$\mathcal{T}_{\mathcal{C}_3}$\xspace}
\newcommand{\tfour}{$\mathcal{T}_{\mathcal{G}_1}$\xspace}
\newcommand{\tfive}{$\mathcal{T}_{\mathcal{G}_2}$\xspace}
\newcommand{\tsix}{$\mathcal{T}_{\mathcal{P}_1}$\xspace}
\newcommand{\tseven}{$\mathcal{T}_{{\mathcal{P}_2}}$\xspace}
\newcommand{\sone}{$\mathcal{S}_1$\xspace}
\newcommand{\stwo}{$\mathcal{S}_2$\xspace}
\newcommand{\sthree}{$\mathcal{S}_3$\xspace}
\newcommand{\know}{$\mathcal{K}$\xspace}
\newcommand{\qs}{$\mathcal{Q}_{\textit{syn}}$\xspace}
\newcommand{\qr}{$\mathcal{Q}_{\textit{real}}$\xspace}
\newcommand{\qt}{$\mathcal{Q}_{\phi}$\xspace}
\newcommand{\qa}{$\mathcal{Q}_{\textit{aux}}$\xspace}
\newcommand{\at}{$\mathcal{A}_{\textit{target}}$\xspace}
\newcommand{\ct}{$\mathcal{C}_{\textit{target}}$\xspace}
\newcommand{\ctr}{$\mathcal{C}_{\textit{target}}^{real}$\xspace}
\newcommand{\gt}{$\mathcal{G}_{\textit{target}}$\xspace}

\newcommand{\mcc}{$\mathcal{M}_{\omega_1}$\xspace}
\newcommand{\mpc}{$\mathcal{M}_{\omega_2}$\xspace}

\newcommand{\cres}{$\mathcal{C}_{\textit{ref}}^{syn}$\xspace}
\newcommand{\cress}{$\{\mathcal{C}_{\textit{ref},1}^{\textit{syn}}, \mathcal{C}_{\textit{ref},2}^{\textit{syn}}, \dots, \mathcal{C}_{\textit{ref},k}^{\textit{syn}}\}$\xspace}
\newcommand{\deltacress}{$\Delta({\mathcal{C}_{\textit{ref}}^{\textit{syn}}})$\xspace}
\newcommand{\deltacrers}{$\Delta(\mathcal{C}_{\textit{ref}}^{\textit{real}})$\xspace}
\newcommand{\crer}{$\mathcal{C}_{\textit{ref}}^{real}$\xspace}
\newcommand{\crers}{$\{\mathcal{C}_{\textit{ref},1}^{\textit{real}}, \mathcal{C}_{\textit{ref},2}^{\textit{real}}, \dots, \mathcal{C}_{\textit{ref},k}^{\textit{real}}\}$\xspace}
\newcommand{\deltactss}{$\Delta(\mathcal{C}_{\textit{target}}^{\textit{syn}})$\xspace}
\newcommand{\deltactrs}{$\Delta(\mathcal{C}_{\textit{target}}^{\textit{real}})$\xspace}

\newcommand{\gress}{$\{\mathcal{G}_{\textit{ref},1}^{\textit{syn}}, \mathcal{G}_{\textit{ref},2}^{\textit{syn}}, \dots, \mathcal{G}_{\textit{ref},k}^{\textit{syn}}\}$\xspace}
\newcommand{\grers}{$\{\mathcal{G}_{\textit{ref},1}^{\textit{real}}, \mathcal{G}_{\textit{ref},2}^{\textit{real}}, \dots, \mathcal{G}_{\textit{ref},k}^{\textit{real}}\}$\xspace}
\newcommand{\deltagress}{$\Delta({\mathcal{G}_{\textit{ref}}^{\textit{syn}}})$\xspace}
\newcommand{\deltagrers}{$\Delta(\mathcal{G}_{\textit{ref}}^{\textit{real}})$\xspace}
\newcommand{\deltagtss}{$\Delta(\mathcal{G}_{\textit{target}}^{\textit{syn}})$\xspace}
\newcommand{\deltagtrs}{$\Delta(\mathcal{G}_{\textit{target}}^{\textit{real}})$\xspace}
\newcommand{\pres}{$\mathcal{P}_{\textit{ref}}^{syn}$\xspace}

\newcommand{\press}{$\{\mathcal{P}_{\textit{ref},1}^{\textit{syn}}, \mathcal{P}_{\textit{ref},2}^{\textit{syn}}, \dots, \mathcal{P}_{\textit{ref},k}^{\textit{syn}}\}$\xspace}
\newcommand{\prers}{$\{\mathcal{P}_{\textit{ref},1}^{\textit{real}}, \mathcal{P}_{\textit{ref},2}^{\textit{real}}, \dots, \mathcal{P}_{\textit{ref},k}^{\textit{real}}\}$\xspace}
\newcommand{\deltaptss}{$\Delta(\mathcal{P}_{\textit{target}}^{\textit{syn}})$\xspace}
\newcommand{\deltaptrs}{$\Delta(\mathcal{P}_{\textit{target}}^{\textit{real}})$\xspace}
\newcommand{\deltapress}{$\Delta({\mathcal{P}_{\textit{ref}}^{\textit{syn}}})$\xspace}
\newcommand{\deltaprers}{$\Delta(\mathcal{P}_{\textit{ref}}^{\textit{real}})$\xspace}

\begin{document}

\begin{textblock}{13}(1.5,1)
\centering
To Appear in the 34th USENIX Security Symposium, August 13-15, 2025.
\end{textblock}

\title{\Large \bf Synthetic Artifact Auditing: Tracing LLM-Generated Synthetic Data Usage in Downstream Applications}

\author{
\rm Yixin Wu\textsuperscript{1}\ \
Ziqing Yang\textsuperscript{1}\ \
Yun Shen\textsuperscript{2}\ \
Michael Backes\textsuperscript{1}\ \
Yang Zhang\textsuperscript{1}\thanks{Yang Zhang is the corresponding author.}
\\
\\
\textsuperscript{1}\textit{CISPA Helmholtz Center for Information Security}\ \ \ 
\textsuperscript{2}\textit{Netapp}\ \ \
}

\date{}

\maketitle


\begin{abstract}

Large language models (LLMs) have facilitated the generation of high-quality, cost-effective synthetic data for developing downstream models and conducting statistical analyses in various domains.
However, the increased reliance on synthetic data may pose potential negative impacts.
Numerous studies have demonstrated that LLM-generated synthetic data can perpetuate and even amplify societal biases and stereotypes, and produce erroneous outputs known as ``hallucinations'' that deviate from factual knowledge.
In this paper, we aim to audit artifacts, such as classifiers, generators, or statistical plots, to identify those trained on or derived from synthetic data and raise user awareness, thereby reducing unexpected consequences and risks in downstream applications.
To this end, we take the first step to introduce synthetic artifact auditing to assess whether a given artifact is derived from LLM-generated synthetic data.
We then propose an auditing framework with three methods including metric-based auditing, tuning-based auditing, and classification-based auditing.
These methods operate without requiring the artifact owner to disclose proprietary training details.
We evaluate our auditing framework on three text classification tasks, two text summarization tasks, and two data visualization tasks across three training scenarios.
Our evaluation demonstrates the effectiveness of all proposed auditing methods across all these tasks.
For instance, black-box metric-based auditing can achieve an average accuracy of $0.868 \pm 0.071$ for auditing classifiers and $0.880 \pm 0.052$ for auditing generators using only 200 random queries across three scenarios.
We hope our research will enhance model transparency and regulatory compliance, ensuring the ethical and responsible use of synthetic data.\footnote{Our code is available at \url{https://github.com/TrustAIRLab/synthetic_artifact_auditing}.}

\end{abstract}

\section{Introduction}

\begin{figure}[!t]
\centering
\includegraphics[width=0.9\columnwidth]{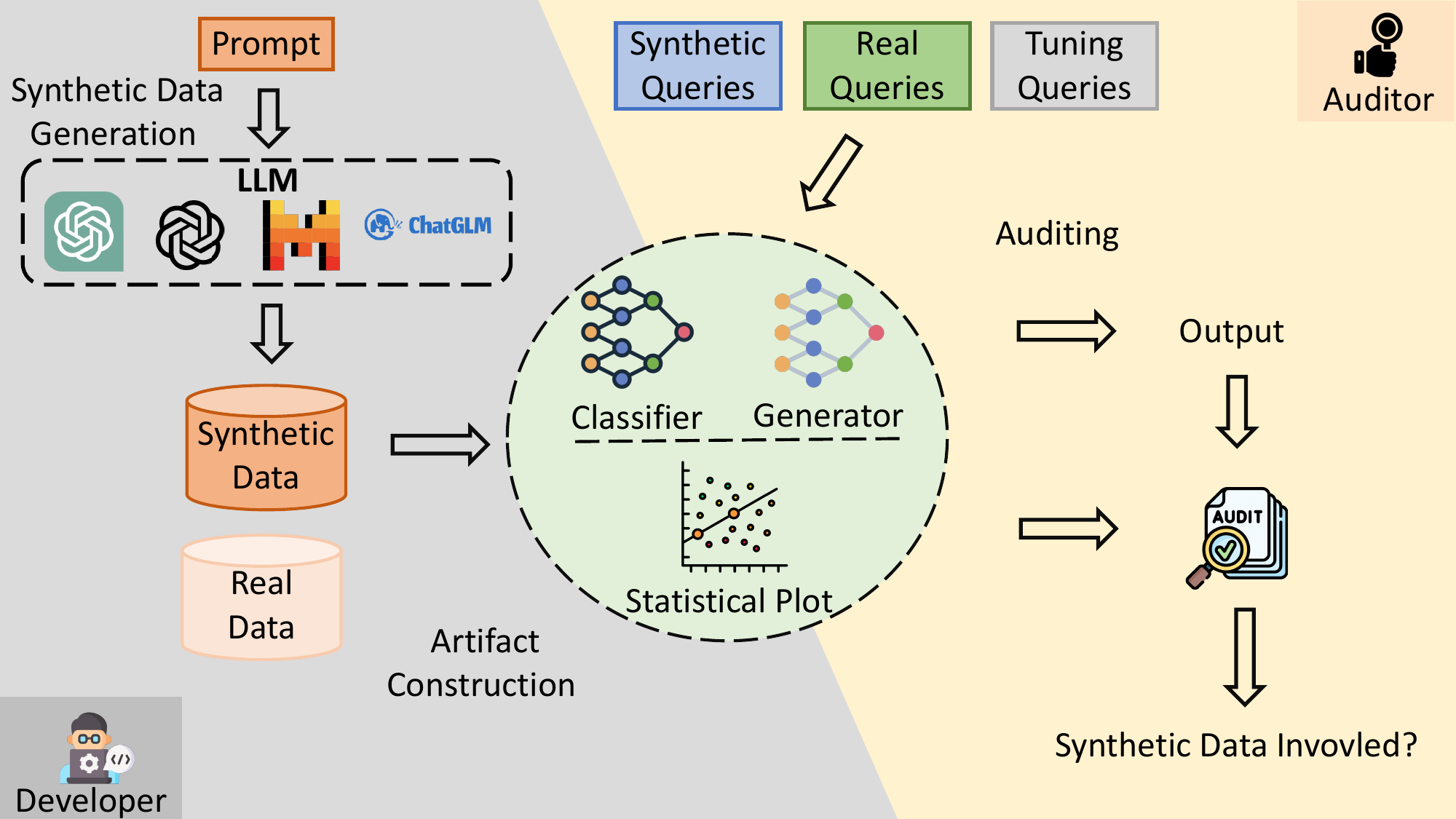}
\caption{Overview of the synthetic artifact auditing.
The auditing targets are: classifiers (\autoref{section:classifier_audit}), generators (\autoref{section:generator_audit}), and statistical plots (\autoref{section:plot_audit}).}
\label{figure:overview}
\end{figure}

Large language models (LLMs) are revolutionizing data acquisition for natural language processing (NLP) tasks.
Collecting high-quality data has always been a challenge due to its labor-intensive and time-consuming nature.
With the advent of the LLM era, the explosive growth in training scales has rapidly increased the corresponding demand for data, further escalating these challenges.
LLMs, known for their ability to generate high-quality data cost-effectively, have spurred a surge in leveraging them for synthetic data generation~\cite{ZGWCW24,DQZLLCXHLJ24,GC24}.

This approach allows for tailoring training data, especially for low-resource NLP tasks like medicine and healthcare~\cite{TTEGTT23,PYCSPCMFZMLMOAHSGBW23,THJH23}.
It can also strategically enhance training data by rebalancing under-represented classes~\cite{HRK23, C23}, and mitigate the privacy risks associated with privacy-concerned data sharing and analysis~\cite{HRK23,HEM19}.
Building on these benefits, synthetic data has rapidly gained widespread adoption across academia and industry.
It facilitates developing NLP models and conducting statistical analyses across various domains, from healthcare~\cite{PYCSPCMFZMLMOAHSGBW23,THJH23,AYFL23,WYMOYY23} and law~\cite{LCLYW24} to education~\cite{NFYDTO23} and scientific discovery~\cite{ZKJNMWP23,HTK23} and marketing~\cite{MDZMB24}.
Frameworks such as Microsoft's AgentInstruct~\cite{MCZMRCLCVRSKLA24} and Hazy's synthetic data generator (acquired by SAS)~\cite{HAZY} are already being actively used in real-world settings.
In practice, Microsoft post-trains Mistral-7b using synthetic data, achieving notable performance gains on multiple benchmarks\cite{MCZMRCLCVRSKLA24}, while OpenAI employs synthetic data generated by o1-preview to fine-tune Canvas~\cite{CANVAS}.

The increasing reliance on LLMs for generating synthetic data, however, raises significant concerns regarding potential adverse impacts~\cite{DLMKLKHOHPKPTFRLK24}.
Numerous studies have demonstrated that LLMs inherently perpetuate or even amplify societal biases and stereotypes related to race, sex, and culture in the uncurated training data, neglecting perspectives from other regions of the world~\cite{GRBTKDYZA23,KDS23,WSP24}.
LLMs can further produce erroneous outputs known as ``hallucinations,'' generating fictional text misaligned with factual knowledge~\cite{JLFYSXIBMF23,RSD23,MLG23}.
With the persistent engagement of LLM-generated synthetic data in downstream training and analysis processes, the dissemination of exaggerated biases and inaccurate information could significantly undermine the reliability of decision-making processes and erode user trust.
Although various studies aim to detect and mitigate bias and hallucination, existing approaches predominantly depend on the self-correction mechanisms of models~\cite{LXM24} and external classifiers~\cite{YFZXWSSLSC23,WBK23}.
Both strategies exhibit limitations in their correction and detection capabilities, making them insufficient to fully eliminate bias and hallucination.
Additionally, concerns have also arisen over the potential unauthorized use of LLM-generated data, with reports~\cite{BYTEDANCE_REPORT} indicating instances where competitors leverage such data to develop competing products in violation of usage terms~\cite{OPENAI_BUSINESS_LICENSE,LLAMA_LICENSE}.
Worse yet, given the rapid development of LLMs, it is highly conceivable that unforeseen issues may emerge in the future.

These issues highlight the necessity for a deeper investigation to determine whether LLM-generated synthetic data was involved in the construction processes of a given artifact.
In response, this paper aims to audit artifacts to label those trained on or derived from synthetic data and raise users' awareness, thereby reducing unexpected consequences and risks in downstream applications.
We first introduce the concept of \textit{synthetic artifact auditing} and frame it as a binary classification task.
Artifact owners typically only reveal the trained models and the analysis results to their users, while keeping the details of training data confidential.
Considering that, we propose an auditing framework with three methods: metric-based auditing, tuning-based auditing, and classification-based auditing.
These methods obviate the need for disclosure of proprietary training specifics.
We currently focus on three types of artifacts that commonly appear in real-world applications: classifiers, generators, and statistical plots (\autoref{figure:overview}).

We evaluate our auditing framework on three text classification tasks, two text summarization tasks, and two data visualization tasks with four LLMs across three scenarios.
In general, it can achieve good auditing performance across all tasks and scenarios.
With black-box access and limited resources, it achieves an average accuracy of $0.868 \pm 0.071$ for auditing classifiers and $0.880 \pm 0.052$ for auditing generators.
Meanwhile, it can also achieve $0.966 \pm 0.003$ average accuracy for auditing statistical plots.
We attribute the high performance to the fact that these downstream synthetic artifacts can learn unique patterns from synthetic data and capture the relationships between them, the target labels, and the reference texts.
In this manner, for example, synthetic classifiers trained on synthetic data exhibit more confidence than classifiers trained on real data in making predictions for synthetic data, thereby aiding in distinguishing between the two.

\mypara{Contributions}
We summarize our contributions as follows:
\begin{itemize}
\item We introduce the concept of synthetic artifact auditing.
Given an artifact, it determines whether it is trained on or derived from LLM-generated synthetic data.
\item We propose an auditing framework with three methods that require no disclosure of proprietary training specifics: metric-based auditing, tuning-based auditing, and classification-based auditing.
This framework is extendable, currently supporting auditing for classifiers, generators, and statistical plots.
\item We evaluate our auditing framework on three text classification tasks, two text summarization tasks, and two data visualization tasks across three training scenarios.
The evaluation demonstrates the effectiveness of all proposed auditing methods across all these tasks.
\end{itemize}

\mypara{Impact}
Our work has a real-world impact, particularly in promoting the responsible use of synthetic data.
Regulatory and governmental bodies are increasingly prioritizing data governance and transparency in the development of AI systems.
For instance, the UK's ICO requires documentation of synthetic data creation and its properties~\cite{ICO}.
Similarly, California recently passed Law AB 2013~\cite{AB2013}, mandating the disclosure of training datasets, including the use of synthetic data~\cite{AB2013_NEWS}.
Our framework provides a practical means for third parties to audit artifacts without requiring the disclosure of proprietary training details by artifact owners.
This supports compliance with data governance and transparency requirements, enhances alignment with regulatory and legal standards, and facilitates responsible and accountable AI practices.
We will open-source our code to facilitate further research.

\section{Synthetic Data Generation}

Synthetic data generation~\cite{BTSLKLCF24} provides a variable solution to scenarios where real data is limited due to high costs~\cite{PKS21}, privacy constraints~\cite{HEACR22}, and biased distributions~\cite{KESGMV19}.
The fundamental objective of synthetic data generation is to produce data that is both plausible and representative of the underlying distribution observed in real data.
The evolution of synthetic data generation models has been closely tied to advancements in machine learning research.
Early approaches predominantly employed statistical methods such as (hidden) Markov chains~\cite{OS14}, and n-grams~\cite{CAC12}, which excel at capturing token co-occurrences but struggle to capture nuanced semantic meanings, resulting in lower-quality generated data.
With the advent of Deep Learning~\cite{GBC16}, synthetic data generation models adopt deep sequential models, such as RNN~\cite{MB13} and (Variational) Autoencoder~\cite{HYLSX17}, and more recent GANs~\cite{ZXLZWHM19}.
These models, trained on larger datasets, can better comprehend token meanings and subsequently generate more realistic data.
The introduction of Transformer architecture~\cite{VSPUJGKP17} has further revolutionized these efforts by utilizing attention mechanisms to model token relationships.
Recent advancements in large language models (LLMs) lead to a surge in synthetic data generation for NLP tasks, such as time series~\cite{ZCGS24}, text~\cite{LZLY23}, and code~\cite{LXWZ23}.

\section{Problem Statement}

\mypara{Auditing Scenario} 
We consider an auditing scenario in which auditors aim to determine whether given artifacts, such as classifiers, generators, and statistical plots, are trained on or derived from LLM-generated synthetic data.
An overview of this auditing scenario is shown in~\autoref{figure:overview}.
Developers prompt LLMs to create synthetic datasets tailored to specific NLP tasks and then use the synthetic data or a combination of real and synthetic data to train models and conduct statistical analyses, such as data visualization.
The auditor can be third-party regulatory agencies, downstream users, or LLM service providers investigating whether certain artifacts are derived from synthetic data.

\mypara{Auditor's Capability}
We consider capabilities as follows:
\begin{itemize}
\item \textit{Access to the target artifact.}
Auditors possess either black-box or white-box access to the auditing target.
Black-box access allows querying the model via an API with input data and receiving outputs.
In contrast, white-box access grants more knowledge including model architecture and parameters.
These capabilities align with previous studies~\cite{WWBBHSZ24,WYLBZ22,SSSS17,HSSDYZ21,SM21}.
For artifacts like statistical plots, auditors have direct access to the targets.
\item \textit{Access to a reference real dataset.}
Auditors leverage their understanding of the target artifact, including its functions and input data, to independently gather a reference real dataset and perform the same task.
We primarily assume that the reference real dataset and the target real dataset originate from a similar distribution.
In~\autoref{section:discussion}, we demonstrate our auditing methods still work well when they come from different distributions.
\item \textit{Access to a synthetic dataset.}
Auditors can instruct LLMs to generate a synthetic dataset tailored to the functions of auditing targets.
We primarily assume that the reference synthetic dataset and the target synthetic dataset are generated from the same source LLM.
In~\autoref{section:discussion}, we demonstrate that our auditing methods still work well when they can come from different source LLMs.
\end{itemize}

\noindent We stress that the auditor has no direct access to the training datasets used to develop the target artifact.
They also do not have access to certain training hyperparameters (e.g., epochs) considered proprietary to the artifact owners.
Furthermore, the reference real data collected by auditors remains entirely independent from the target artifact.
Our experimental setup reflects the above settings (see \autoref{section:classifier_eval_setup} for details).

\mypara{Synthetic Artifact Auditing}
Formally, we formulate our auditing as a binary clarification problem.
That is, given a target artifact \at and external knowledge \know of an auditor, synthetic artifact auditing can be defined as follows:
\begin{equation}
   \mathcal{A}_{\textit{target}}, \mathcal{K} \rightarrow \{\mathbf{0}, \mathbf{1}\}, 
\end{equation}
where $\mathbf{1}$ denotes that synthetic data was involved in the target classifiers' and generators' training procedure or used to generate target statistical plots, and $\mathbf{0}$ indicates otherwise.

\mypara{Note}
Building models from scratch for NLP tasks has been uncommon in recent years.
Instead, developers typically fine-tune pre-trained language models (PLMs) for specific tasks.
Therefore, we consider that our artifacts are fine-tuned from PLMs, such as BERT~\cite{DCLT19} and BART~\cite{LLGGMLSZ20}.
Moreover, we do not consider the scenario where auditing targets are LLMs for the following reasons.
First, in real-world applications, efficiency, cost-effectiveness, and customizability are essential.
Smaller and less complex models are not only easier to train and deploy but also enable faster inference.
For example, using an expensive LLM for sentiment analysis may not be financially sustainable.
Therefore, the mainstream approach remains to fine-tune much smaller PLMs tailored to specific downstream tasks~\cite{GC24,JSPW23,MPDA24,HRK23,NFYDTO23}.
Second, recent studies~\cite{LCLYW24,THJH23} show that these fine-tuned PLMs achieve overall even better performance in domain-specific tasks.

\section{Classifier Auditing}
\label{section:classifier_audit}

\begin{figure}[!t]
\centering
\includegraphics[width=0.9\columnwidth]{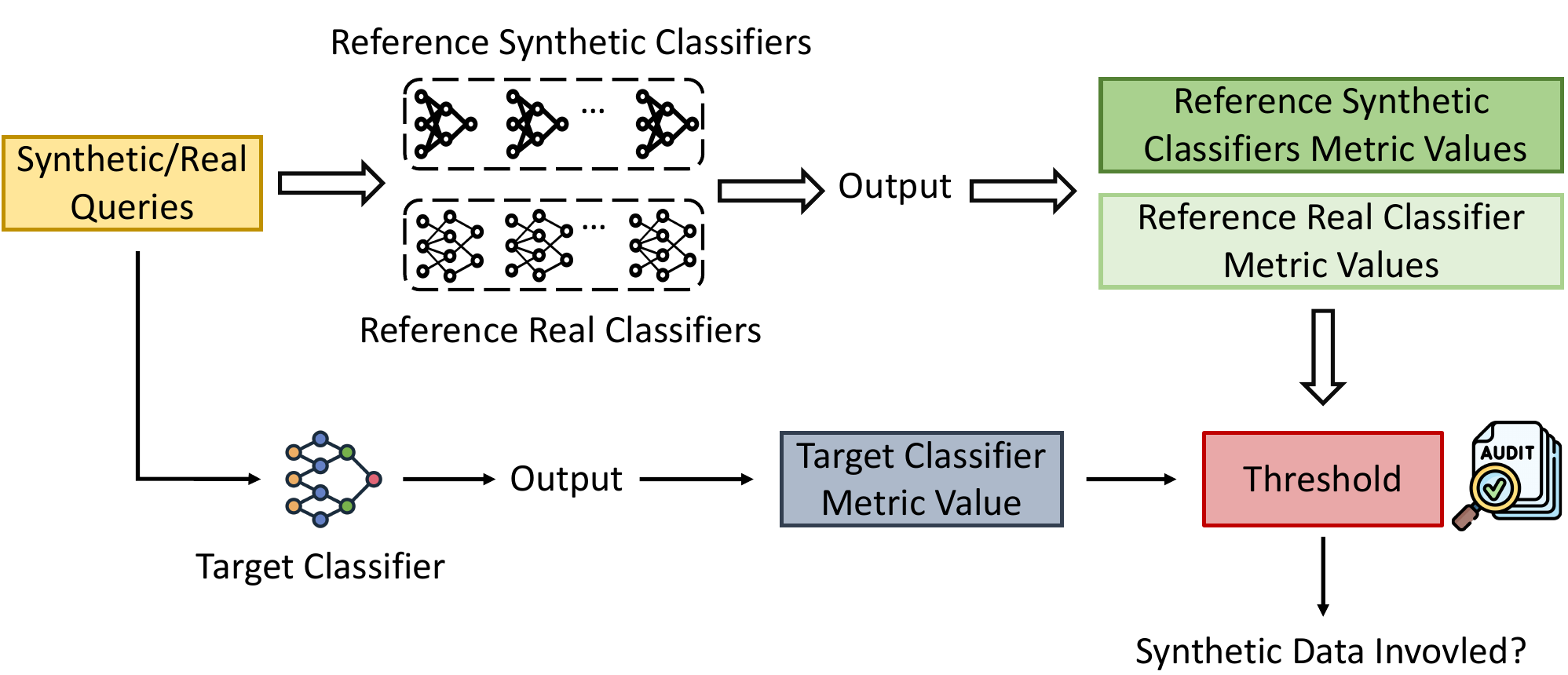}
\caption{Overview of the metric-based auditing.}
\label{figure:metric_based_audit}
\end{figure}

\subsection{Metric-Based Auditing}
\label{section:classifier_metric_audit}

\mypara{Intuition}
We first consider the case where the auditors only have black-box access to the target classifier \ct and query it with input texts to obtain outputs.
Recent studies~\cite{LYLL23, HSCBZ24} show that LLM-generated synthetic data has unique lexical, structural, and semantic features that distinguish it from real data.
Classifiers trained with such data may likely learn to recognize and leverage these distinctive features to predict labels effectively.
Therefore, we hypothesize that classifiers trained with more synthetic data, referred to as \textit{synthetic classifiers}, tend to be more confident when predicting labels for synthetic input texts and less confident with real input texts compared to \textit{real classifiers}.
Conversely, classifiers trained predominantly on real data may exhibit lower confidence when predicting labels for synthetic inputs and higher confidence with real input texts, as these real classifiers have not ``seen'' these distinctive synthetic features during training.

\mypara{Methodology}
Building upon the hypothesis of these behavior disparities, we develop a metrics-based auditing approach.
It involves designing a query set $\mathcal{Q}=\{(x_i, y_i)\}^{n}_{i=1}$, where $x_i$ represents the input text and $y_i$ denotes the target label, and evaluating the classifier's outputs for this query set via a performance metric to conduct auditing.
We present the overview of our metric-based auditing method in~\autoref{figure:metric_based_audit}.
The auditor first queries the target classifier with $\mathcal{Q}$, either \qs consisting of synthetic data or \qr consisting of real data, and obtains outputs.
The auditor then computes the values of the performance metric for all data in the query set and compares the average values of those metrics with a certain threshold to determine whether the target classifier \ct is a \textit{synthetic classifier} or a \textit{real classifier}.
More formally, we define the metric-based auditing using \qs and \qr as follows:
\begin{equation}
   \mathcal{I}_{conf}( \mathcal{C}_{\textit{target}}, \mathcal{Q}_{\textit{syn}}) =  \mathds{1}\{\frac{1}{n} \sum_{i=1}^{n} \mathcal{C}_{\textit{target}}(x_i)_{y_i} > \tau \},
\end{equation}
\begin{equation}
   \mathcal{I}_{conf}( \mathcal{C}_{\textit{target}}, \mathcal{Q}_{\textit{real}}) =  \mathds{1}\{\frac{1}{n} \sum_{i=1}^{n} \mathcal{C}_{\textit{target}}(x_i)_{y_i} < \tau \}.
\end{equation}
We leverage the average confidence score of the query set as an example of the performance metric.
We can also select other performance metrics, such as the average of entropy values and accuracy.
The corresponding definitions are shown in~\refappendix{appendix:equation_for_entropy}.
The auditor empirically determines the threshold $\tau$ based on their reference classifiers.
Specifically, the auditor (1) trains synthetic reference classifiers (\deltacress$=$\cress) using a mix of synthetic and real data, and real reference classifiers (\deltacrers$=$\crers)  exclusively using real data; (2) then leverages the query set (\qs/\qr) to obtain the reference classifiers' outputs and computes performance metric values; (3) establishes an empirical threshold value that achieves the highest accuracy in distinguishing between \cres and \crer.
The auditor inevitably needs to obtain training data for these reference classifiers.
Since the auditor understands the exact downstream task that the target classifier performs, in turn, for \crer, the auditor can independently source corresponding training data (e.g., from the Web).
For \cres, the auditor can prompt LLMs with a task description to generate corresponding synthetic data.
Note that the proportion of synthetic data in the training dataset of different \cres can vary.
This allows the \cres to be trained solely on synthetic data or on a mix of synthetic and real data with random proportions.

\subsection{Tuning-Based Auditing}
\label{section:classifier_tuning_audit}

\mypara{Intuition}
When the auditor has white-box access to the audit target, they gain full visibility into the model's architecture and parameters.
A common approach in such scenarios, as suggested by prior research~\cite{GWYGB18}, may utilize a flattened vector of all parameters as input and train a binary classifier to conduct the audit.
However, our empirical analysis demonstrates that this approach yields results close to random guessing.
Besides the knowledge of the model's architecture and parameters, white-box access further exempts the auditing target from being confined to processing generic text inputs, i.e., discrete tokens, enabling it to accept continuous embeddings as input.
This flexibility allows us to iteratively refine an input query set from a continuous optimization space, thereby yielding more precise outputs to effectively identify behavior discrepancies between synthetic and real targets.

\begin{figure}[!t]
\centering
\includegraphics[width=0.9\columnwidth]{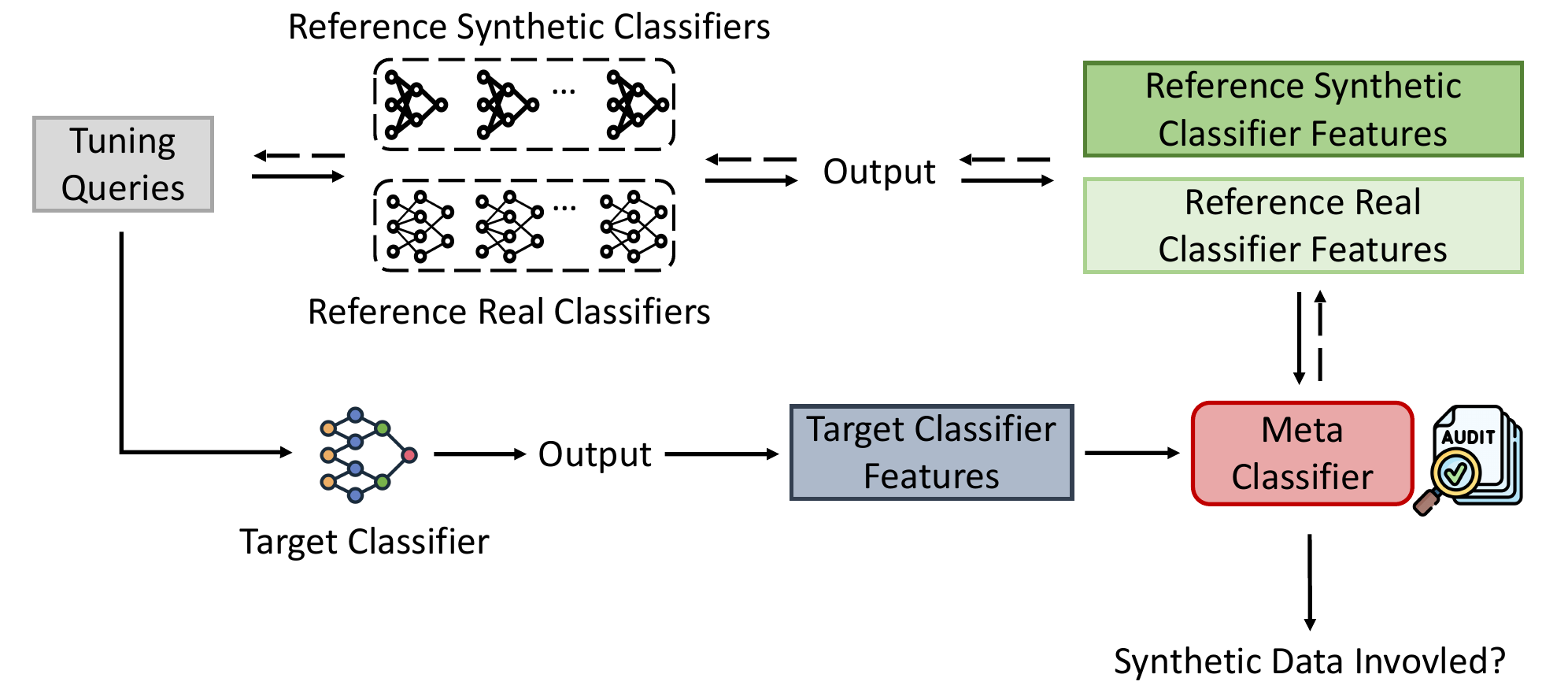}
\caption{Overview of the tuning-based auditing.}
\label{figure:tuning_based_audit}
\end{figure}

\mypara{Methodology}
We present the overview of tuning-based auditing in~\autoref{figure:tuning_based_audit}.
Similar to metric-based auditing, the auditor first trains a small set of synthetic reference classifiers \deltacress and real reference classifiers \deltacrers.
Given these reference classifiers, the auditor leverages a simple gradient-based approach where they directly optimize the query set \qt parameterized by $\phi$ and a \textit{meta-classifier} \mcc parametrized by $\omega_1$ via backpropagation.
The meta-classifier uses the output probabilities (posteriors) from a given classifier to predict its assigned label.
More formally, the auditor aims to maximize the likelihood of the correct label $y \in \mathcal{Y}$, i.e., indicating whether it is a synthetic or real classifier, for the corresponding reference classifiers as follows:
\begin{equation}
\max\limits_{\phi;\omega_1} P_{\omega_1;\theta;\phi} (\mathcal{Y}|\mathcal{M}_{\omega_1}(\mathcal{C}_{ref, \theta}(\mathcal{Q}_{\phi}))),
\end{equation}
where the parameters $\phi$ of the query set and the parameters $\omega_1$ of the meta-classifier are learned via back-propagation and the parameters $\theta$ of the reference classifiers are frozen.
At inference time, the auditor queries the target classifier with learned \qt, and then feeds the outputs into the trained meta-classifier \mcc to make the predictions.
The learned \qt is a format of embedding vectors, the auditor thereby leverages white-box access to the target classifier to feed the embeddings into it.

\subsection{Target and Reference Classifier Setup}
\label{section:classifier_eval_setup}

We mainly consider three text classification tasks: sentiment analysis on the IMDB dataset~\cite{IMDB} (\tone); topic classification on the AG's news (abbreviated as AG) dataset~\cite{ZZL15} (\ttwo); spam detection on the Enron-Spam dataset~\cite{MAP06} (\tthree).
We provide task details in~\refappendix{appendix:eval_setup_task_sup}.
\noindent Our overall setup for target and reference classifiers (see \autoref{figure:overview_eval_setup}) contains three primary steps.
We provide a brief overview of their main objectives below, with further details elaborated in subsequent sections.
\begin{itemize}
\item \textbf{Data Splitting.} 
We prepare the real dataset and the auxiliary dataset.
We ensure that both datasets remain mutually exclusive to maintain the integrity and objectivity of the evaluation process.
\item \textbf{Synthetic Data Generation.}  
We utilize four representative large language models (LLMs) as sources, alongside two prompting strategies, to generate synthetic data.
This guarantees diversity and enhances the quality of the synthetic data produced.
\item \textbf{Training Scenarios.}
We mainly establish three distinct training data composition scenarios for the synthetic classifier.
It is designed to simulate the training process of the auditing target, forming the core setup of our evaluation.
\end{itemize}

\subsubsection{Data Split}
\label{section:data_split}

As illustrated in~\autoref{figure:overview_eval_setup}, we first partition the whole dataset evenly into two disjoint subsets: the target dataset \dt and reference dataset \dr.
\dt is further divided evenly into two disjoint splits as the target real dataset \dtr and the target auxiliary dataset \dta.
We reserve a fixed 1,000 samples in \dr as the testing set \dtest, which is exclusively used to assess classifier performance.
The remaining samples in \dr are further evenly divided into two disjoint subsets \drr and \dra.
We later randomly sample instances from \dtest to construct \qr and use them as reference samples for constructing \qs in a paraphrasing prompt strategy.
All text classification tasks follow the above process for data split, where the specific details for each task are shown in~\refappendix{appendix:eval_setup_data_split_sup}.

\subsubsection{Synthetic Data Generation}
\label{section:synthetic_data_generation}

\mypara{Sources}
Four representative LLMs are employed as the synthetic data generation sources, including \texttt{gpt-3.5-turbo-1106} (GPT-3.5 Turbo)~\cite{GPT-3.5-Turbo}, \texttt{gpt-4-0613} (GPT-4)~\cite{GPT-4-0613}, \texttt{Mistral-7B-Instruct-v0.2} (Mistral)~\cite{Mistral-7B-Instruct-v0.2},  and \texttt{chatglm3-6b} (ChatGLM3)~\cite{ChatGLM3} as the synthetic data generation sources.

\begin{figure}[!t]
\centering
\includegraphics[width=0.9\columnwidth]{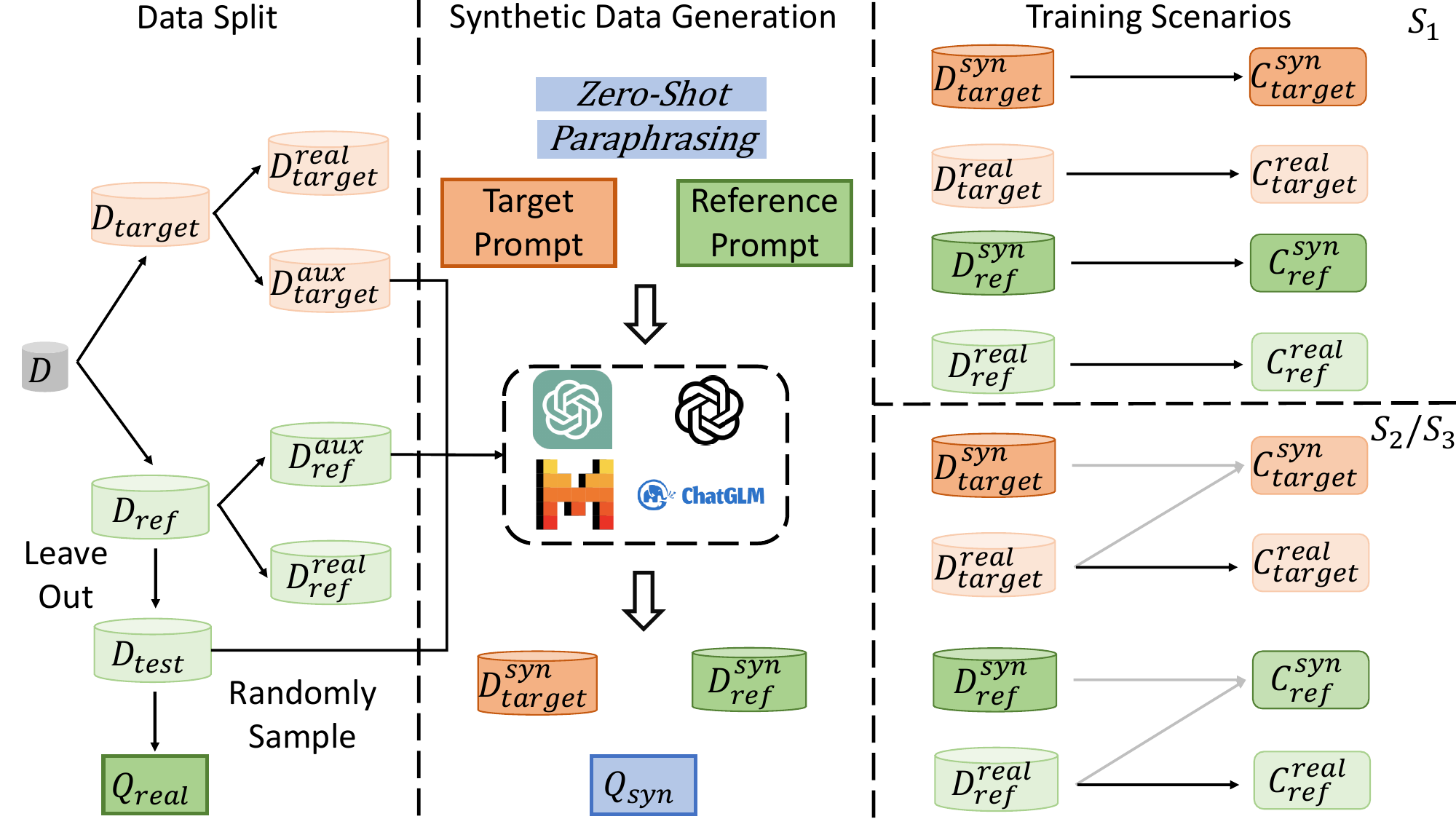}
\caption{Overview of target/reference classifier setup.}
\label{figure:overview_eval_setup}
\end{figure}

\mypara{Strategies}
As illustrated in~\autoref{figure:overview_eval_setup}, we leverage \dta, \dra, and a random subset of \dtest to generate the target synthetic dataset \dts, reference synthetic dataset \drs, and synthetic query set \qs.
All four LLMs are utilized in this process.
Specifically, with \dta, we construct four synthetic datasets \dts containing synthetic data from GPT-3.5, GPT-4, Mistral, and ChatGLM3, respectively.
To accommodate different task requirements, we employ two representative prompting strategies to generate synthetic data~\cite{LYLL23,LZLY23,MPDA24}.
\begin{itemize}
\item \textit{Zero-shot}.
We provide LLMs with labels (and additional information) and instruct them to generate content from scratch.
\item \textit{Paraphrasing}.
We provide the label along with an entire input text as a reference to the LLMs and instruct them to generate new content based on the reference input.
\end{itemize}
For \tone, we adopt the zero-shot strategy by providing movie names and outlines to ensure the quality of the generated review.
The goal is to achieve comparable performance between synthetic and real classifiers.
For \ttwo and \tthree, the zero-shot strategy cannot yield high-quality news articles and email messages.
We thus opt for the paraphrasing strategy.
Note that we do not assume that the auditor knows the prompts used to construct the target synthetic dataset \dts.
We therefore use different prompts to construct \qs and \drs from those used for \dts across all tasks.
We experiment with prompts for synthetic data generation to ensure that synthetic artifacts achieve stable performance, ensuring the reliability of our evaluation.
Further details on synthetic data generation are available in~\refappendix{appendix:eval_setup_syn_data_sup}.
The specific prompts used in our evaluation are detailed in~\refappendix{appendix:prompt}.

\subsubsection{Training Scenarios}
\label{section:training_scenarios}

\mypara{Training Data for Synthetic Classifier}
The use of synthetic data in real-world applications falls into two primary categories: (1) augmenting training data in low-resource scenarios~\cite{CKAK21,MPDA24,HRK23,LRJHH23}, and (2) generating synthetic datasets from scratch to support model training~\cite{JSPW23,LZLY23,THJH23,GPLXYWZLLK23}.
Stemming from these applications, we consider the following three scenarios in our experimental setup:
\begin{itemize}
\item \sone: Training exclusively on synthetic data (100\%) generated from a single LLM.
\item \stwo: Training on a combination of real and synthetic data from a single LLM.
In evaluation, we vary the proportion of synthetic data from 10\% to 100\% in increments of 10\%.
\item \sthree: Training on a mix of real and synthetic data from multiple LLMs (all four LLMs).
The proportions of synthetic data and the sources of synthetic data from different LLMs are randomized.
\end{itemize}

\mypara{Training Data for Real Classifier}
Reference/target real classifiers are exclusively trained on real data.
\ctr is trained solely on \dtr, and \crer is trained solely on \drr.

\noindent We consider all three scenarios for training synthetic classifiers in each task and include more details of the training scenarios in~\refappendix{appendix:eval_setup_training_classifier_sup}.

\mypara{Target/Reference Classifiers}
We utilize BERT~\cite{DCLT19} and DistilBERT~\cite{SDCW19} as base classifiers.
A linear classification layer is tuned on top of these pre-trained models to predict target labels for various tasks.
Our fine-tuning process employs cross-entropy loss and the Adam optimizer with a learning rate set to 2e-5.
We fine-tune the classifiers of \tone for 5 epochs and those of \ttwo and \tthree for 3 epochs.
We develop 50 target synthetic classifiers based on each LLM in both \sone and \stwo.
These target classifiers are used to evaluate the auditing performance.
Overall, we train a total of 50 target real classifiers, 200 target synthetic classifiers in \sone, 200 target synthetic classifiers in \stwo, and 50 target synthetic classifiers in \sthree for each task.
We also train the same number of reference classifiers to determine the threshold values and to train the meta-classifier.
Both target and reference classifier sets are balanced in terms of class distribution.

\mypara{Target Classifier Performance}
The primary metric used to assess classifier performance (i.e., utility) is accuracy on the testing dataset.
We ensure that synthetic classifiers achieve performance comparable to real classifiers.
We provide more details of these results in~\refappendix{appendix:target_classifier_performance}.

\begin{figure}[!t]
\centering
\begin{subfigure}{0.45\columnwidth}
\includegraphics[width=\columnwidth]{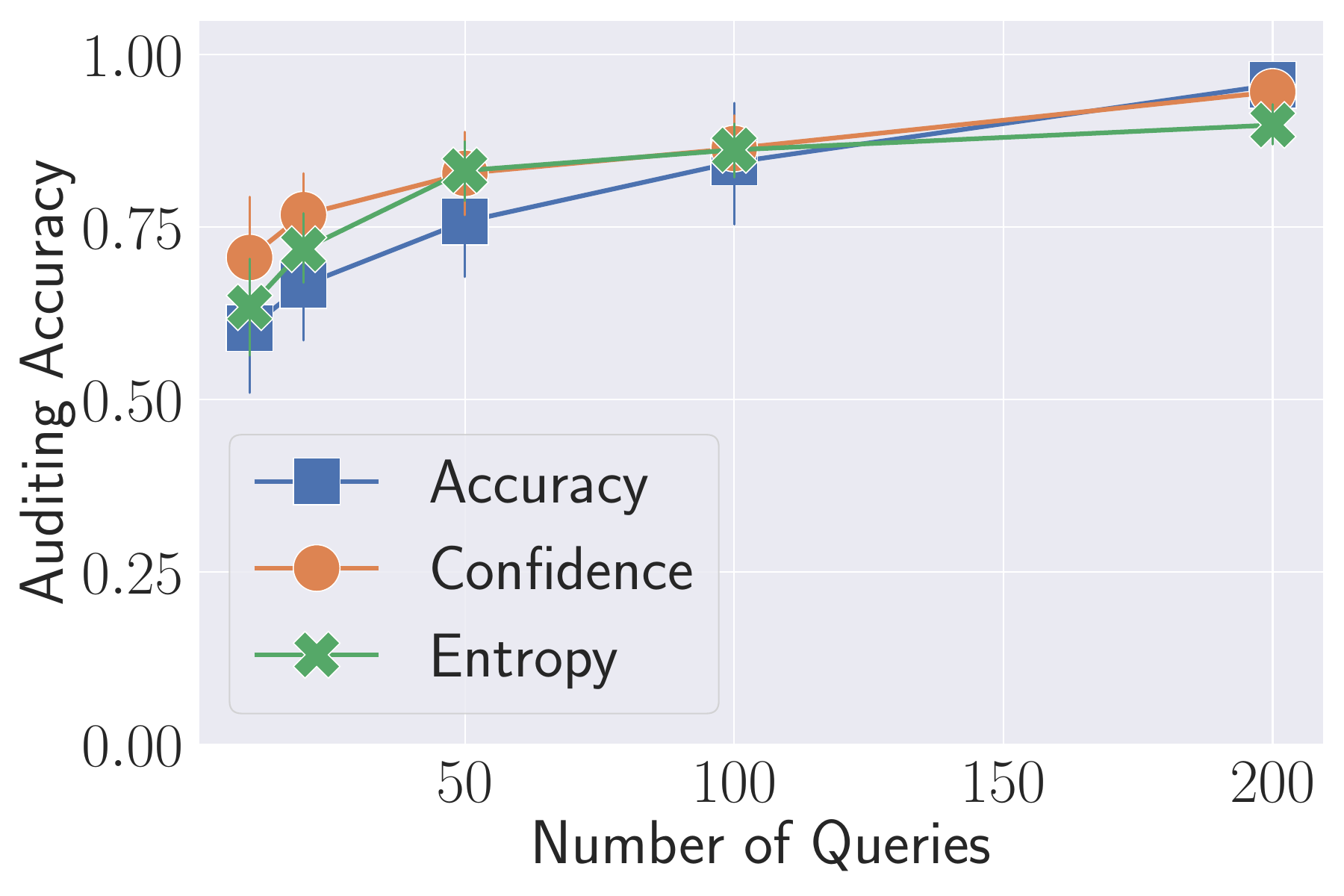}
\caption{\sone}
\label{figure:audit_classifier_x_query_gpt3.5_syn_s1}
\end{subfigure}
\begin{subfigure}{0.45\columnwidth}
\includegraphics[width=\columnwidth]{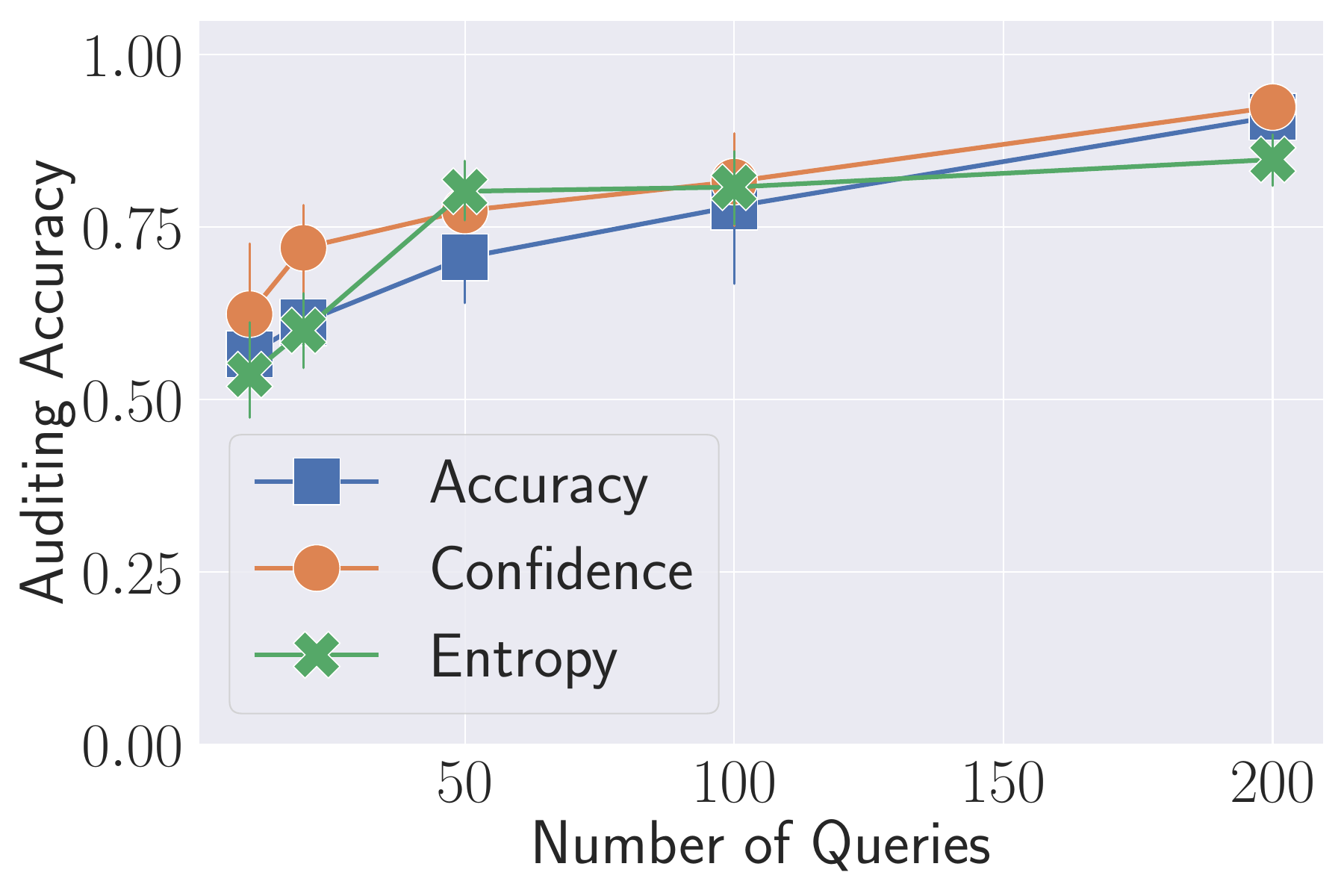}
\caption{\stwo}
\label{figure:audit_classifierr_x_query_gpt3.5_syn_s2}
\end{subfigure}
\caption{Metric-based auditing performance for target classifiers fine-tuned on pre-trained DistilBERT with varying query budgets of \qs $\{10, 20, 50, 100, 200\}$ for \tthree in (a) \sone and (b) \stwo.
The source LLM is GPT-3.5.
The auditing performance of other source LLMs and tasks is shown in~\refappendix{appendix:metric_based_audit_classifier_qs_sup}.}
\label{figure:audit_classifier_x_query_syn}
\end{figure}

\subsection{Auditing Setup}
\label{section:auditing_setup}

\mypara{Auditing Model}
The metric-based auditing calculates a performance metric value for each classifier and uses a reference classifier set to establish a threshold value for auditing targets.
We consider accuracy (Accuracy), average confidence scores (Confidence), and average entropy values (Entropy) on the query set (\qs/\qr) as the performance metrics to enable metric-based auditing.
For tuning-based auditing, a meta-classifier is trained to conduct the auditing, which is a 3-layer MLP model with 32 neurons in the hidden layer.
It directly takes the outputs (posteriors) from the classifiers as input.
We use the cross-entropy loss and optimize it with the Adam optimizer with a learning rate of 1e-3.
The meta-classifier is trained on the reference classifier set for 50 epochs.

\mypara{Auditing Evaluation Protocols}
We ensure balanced class distributions in both the target and reference classifiers.
Consequently, auditing accuracy on the target classifiers is employed as the primary metric.
The target classifiers include 50 target real classifiers and 50 target synthetic classifiers for each scenario.
In \sone, the target synthetic classifiers have a 100\% synthetic proportion.
In \stwo, each of the ten different synthetic proportions corresponds to five classifiers.
In \sthree, the target synthetic classifiers have a random synthetic proportion.
Each experiment is run five times with different seeds and evaluated on all 100 target classifiers.
The final average score is reported alongside its corresponding error bar.

\subsection{Preliminary Investigation}
\label{section:pre_invest}

\begin{table}[!t]
\caption{Average metric-based auditing performance for target classifiers fine-tuned on pre-trained DistilBERT, enabled by three different metrics and different query budgets, across three tasks and two pre-trained models in \sone and \stwo.}
\label{table:different_metrics}
\centering
\scalebox{0.65}{
\begin{tabular}{c | c |c c c}
\toprule
$|\mathcal{Q}|$ &  Query Type & Accuracy & Confidence & Entropy \\
\midrule
\multirow{2}{*}{10} & \qs & $0.617 \pm 0.116$ &  $\mathbf{0.777 \pm 0.139}$ & $ 0.714 \pm 0.126$ \\
 &   \qr & $0.624 \pm 0.120$ & $0.751 \pm 0.142$ & $\mathbf{0.766 \pm 0.094}$\\
 \midrule
\multirow{2}{*}{200} & \qs & $0.765 \pm 0.072$ &  $\mathbf{0.870 \pm 0.070}$ & $ 0.782 \pm 0.075$ \\
  &  \qr & $0.721 \pm 0.098$ & $0.735 \pm 0.120$  & $\mathbf{0.796 \pm 0.065}$\\    
\bottomrule
\end{tabular}}
\end{table}

In this section, we investigate the appropriate query budget, performance metric, and the number of reference classifiers to enable metric-based auditing with \qs (\autoref{section:metric_qs_result}) and \qr (\autoref{section:metric_qr_result}) and tuning-based auditing (\autoref{section:tuning_result}).

\subsubsection{Metric-Based Auditing with \qs}
\label{section:metric_qs_result}

We conduct this investigation in \sone and \stwo.
We initially leverage 10 reference real and 10 reference synthetic classifiers for each scenario.
In \sone, the synthetic classifiers have a 100\% synthetic proportion.
In \stwo, each synthetic classifier has one of ten different synthetic proportions.
We show the metric-based auditing performance with varying query budgets in~\autoref{figure:audit_classifier_x_query_syn}.
The varying query budgets are described as different numbers of queries, and we randomly select synthetic queries each time.
We find that in both scenarios, more synthetic queries help determine a better threshold, resulting in improved auditing performance.
This is reflected not only in higher accuracy but also in a smaller standard deviation.
For example, the metric-based auditing using Confidence achieves only $0.624 \pm	0.164$ with 10 synthetic queries in \stwo, but it increases to $0.924 \pm 0.046$, a significant margin of increase of 0.3, with 200 queries.
Meanwhile, we observe that conducting metric-based auditing with different metrics also leads to varying auditing performance.
To determine which metric yields better auditing performance, we average all results conducted with 200 synthetic queries across three tasks and two pre-trained models in \sone and \stwo.
As reported in~\autoref{table:different_metrics}, the average confidence scores (Confidence) enable the best auditing performance with higher accuracy and lower standard deviation when using \qs.
We speculate that this is because synthetic data is generated based on their labels as conditions, resulting in features that represent the target class and a clear decision boundary between input texts of different classes.
In~\autoref{figure:tsne_plot_word2vec}, it is also demonstrated that synthetic data has clearer boundaries compared to real data, and we defer more discussions in~\autoref{section:audit_plot_eval}.
Consequently, both synthetic and real classifiers tend to make the right predictions for synthetic data, and thus it is difficult to audit based on the accuracy of \qs.
Furthermore, since synthetic classifiers have involved more synthetic data featuring similar characteristics during training, they display higher confidence, thereby enabling them to be distinguished from real classifiers.
Meanwhile, in~\refappendix{appendix:metric_based_audit_classifier_qs_sup}, we demonstrate that 20 reference classifiers are sufficient to launch a successful metric-based auditing with \qs, and the benefit of more reference classifiers is minimal.
We present more results in~\refappendix{appendix:metric_based_audit_classifier_qs_sup} with similar conclusions.

\begin{figure}[!t]
\centering
\begin{subfigure}{0.45\columnwidth}
\includegraphics[width=\columnwidth]{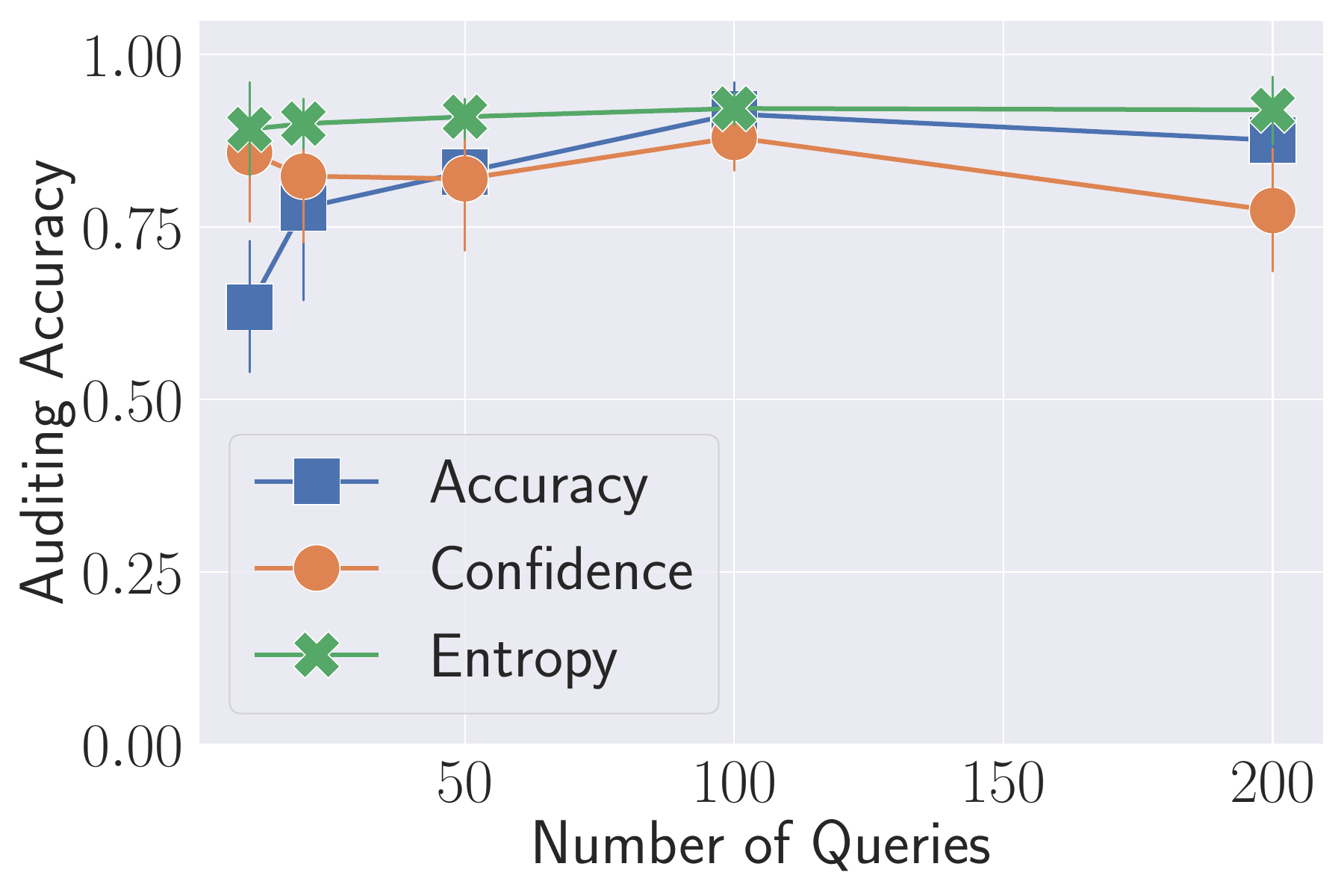}
\caption{\sone}
\label{figure:audit_classifier_x_query_gpt3.5_real_s1}
\end{subfigure}
\begin{subfigure}{0.45\columnwidth}
\includegraphics[width=\columnwidth]{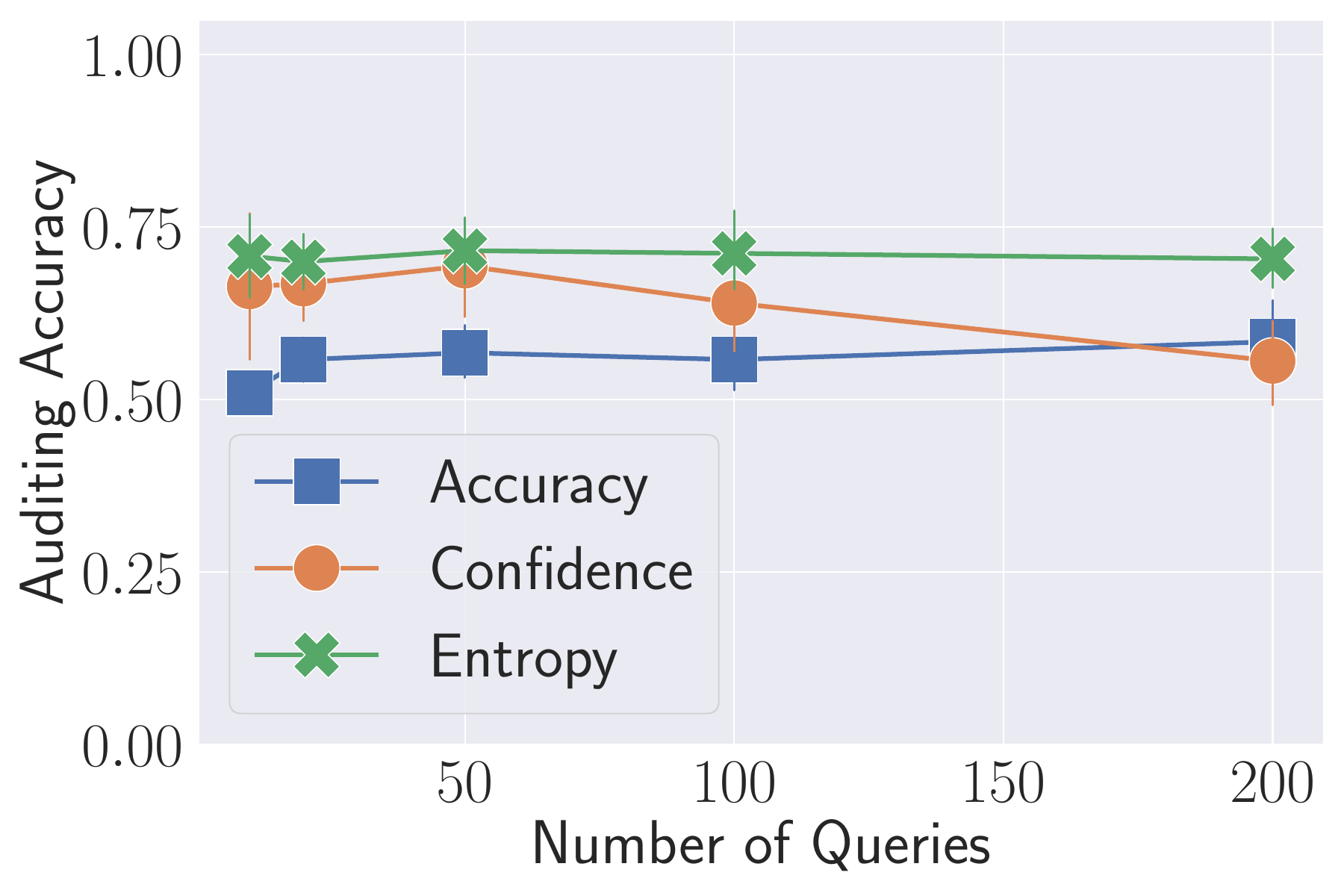}
\caption{\stwo}
\label{figure:audit_classifierr_x_query_gpt3.5_real_s2}
\end{subfigure}
\caption{Metric-based auditing performance for target classifiers fine-tuned on DistilBERT with varying query budgets of \qr $\{10, 20, 50, 100, 200\}$ for \tthree.
The source is GPT-3.5.
The performance of other sources/tasks is in~\refappendix{appendix:metric_based_audit_classifier_qr_sup}.}
\label{figure:audit_classifier_x_query_real}
\end{figure}

\subsubsection{Metric-Based Auditing with \qr}
\label{section:metric_qr_result}

We follow the same evaluation setup and show the auditing performance with varying query budgets in~\autoref{figure:audit_classifier_x_query_real}.
The varying query budgets are described as different numbers of queries, and we randomly select real queries each time.
We find that as the query budget for \qr increases, the auditing performance is relatively stable.
For example, metric-based auditing using Entropy achieves only $0.892 \pm 0.126$ with 10 real queries in S1, and it increases to $0.920 \pm 0.091$, a small margin of only 0.028, with 200 queries.
It might be because real data, carefully handcrafted by humans, has more stable text quality compared to synthetic data, enabling a considerable auditing performance even with $|\mathcal{Q}_{\textit{real}}| = 10$ in some cases.
Meanwhile, we can observe the varying auditing performance using different metrics.
As illustrated in~\autoref{table:different_metrics}, we also average all auditing results conducted with 200 real queries.
Metric-based auditing using Entropy achieves the best auditing performance, i.e., $0.766 \pm 0.094$ on average with 10 queries and $0.796 \pm 0.065$ on average with 200 queries.
In addition, we observe that, with Confidence as the metric, real queries not only underperform compared to using Entropy, but their performance also deteriorates with the growth of the query budget.
We attribute it to the less distinct decision boundary among real input from different classes (\autoref{figure:tsne_plot_word2vec}).
This results in both synthetic and real classifiers displaying less decisive confidence, thereby making predictions more difficult.
Consequently, it is challenging to establish a threshold for synthetic artifact auditing based on Confidence which is a metric focused on a single class and leverages Accuracy.
Alternatively, real queries can rely on Entropy, which measures the uncertainty across the entire probability distribution of all classes.
This approach can lead to superior auditing performance.
Similar to using \qs, we demonstrate that 20 reference classifiers are sufficient in~\refappendix{appendix:metric_based_audit_classifier_qr_sup}.
We present more results on other LLMs and tasks in~\refappendix{appendix:metric_based_audit_classifier_qr_sup} with similar conclusions.

\begin{figure}[!t]
\centering
\begin{subfigure}{0.45\columnwidth}
\includegraphics[width=\columnwidth]{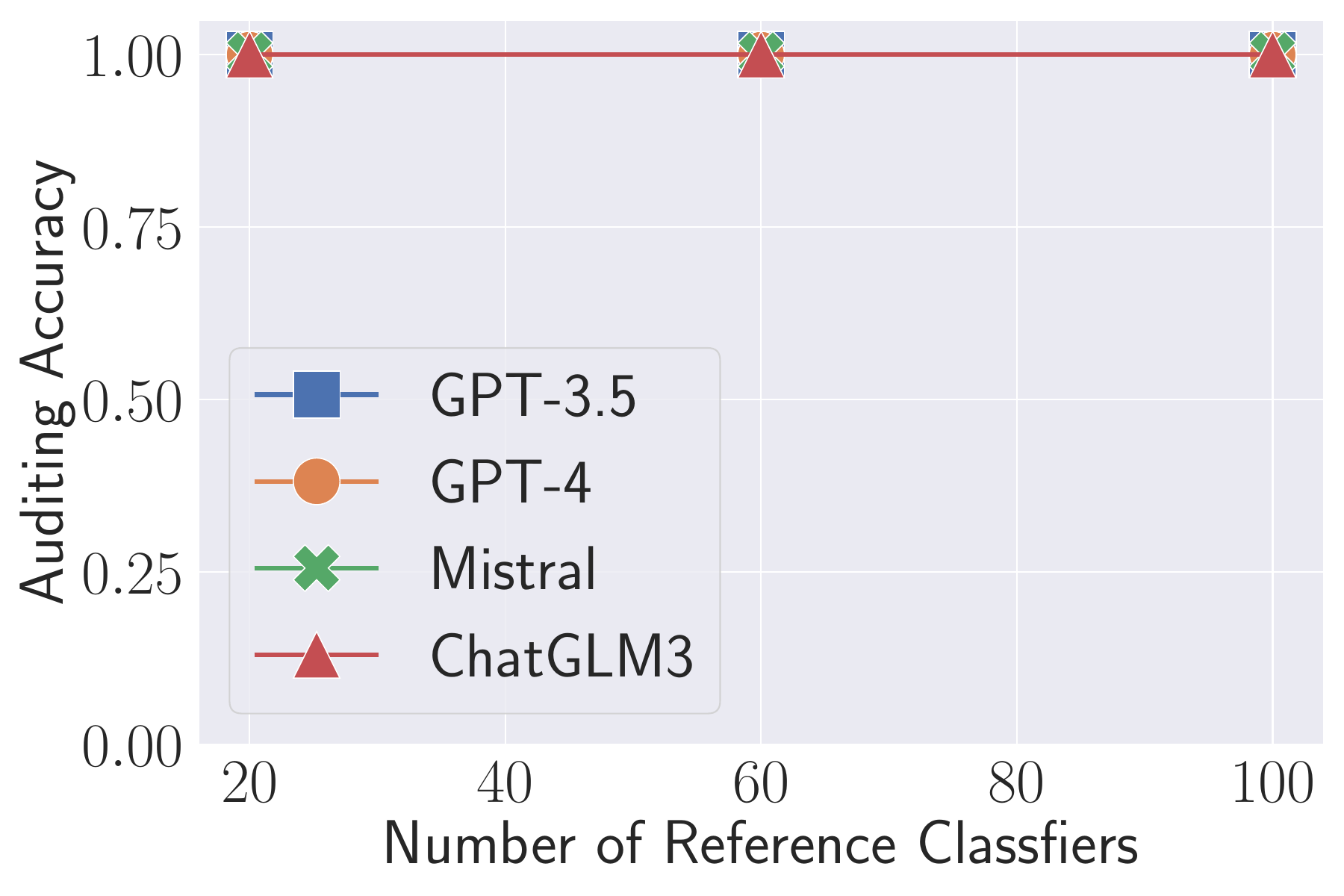}
\caption{\sone}
\label{figure:audit_classifier_x_ns_sd_tuning_t3_s1}
\end{subfigure}
\begin{subfigure}{0.45\columnwidth}
\includegraphics[width=\columnwidth]{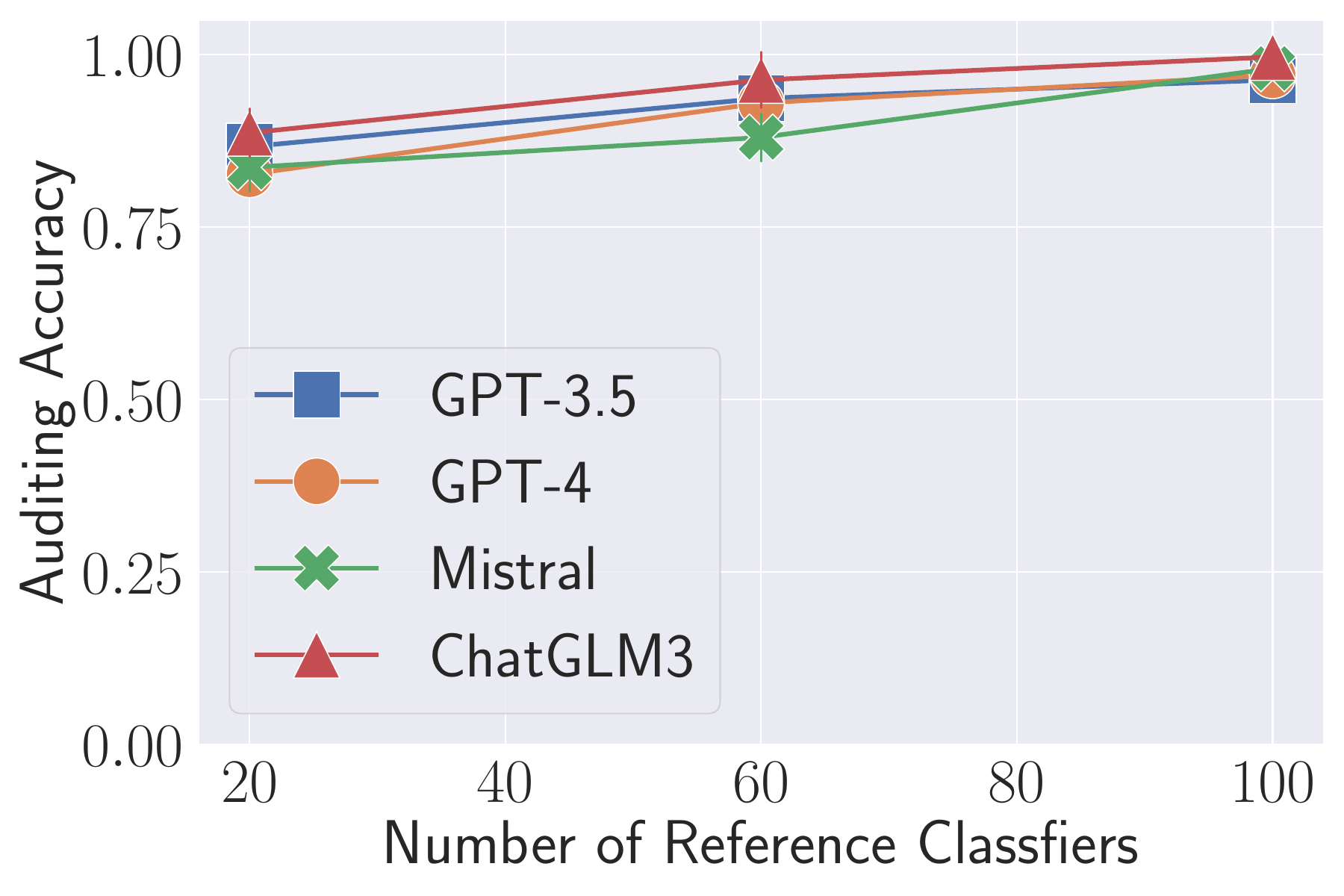}
\caption{\stwo}
\label{figure:audit_classifier_x_ns_sd_tuning_t3_s2}
\end{subfigure}
\caption{Tuning-based auditing performance for target classifiers fine-tuned on DistilBERT with varying numbers of reference classifiers $\{20, 60, 100\}$ for \tthree in (a) \sone and (b) \stwo.
The performance of other tasks is shown in~\refappendix{appendix:tuning_based_audit_classifier_qt_sup}.}
\label{figure:audit_classifier_x_ns_tuning_t3}
\end{figure}

\subsubsection{Tuning-Based Auditing}
\label{section:tuning_result}

\begin{figure*}[!t]
\centering
\begin{subfigure}{0.45\columnwidth}
\includegraphics[width=\columnwidth]{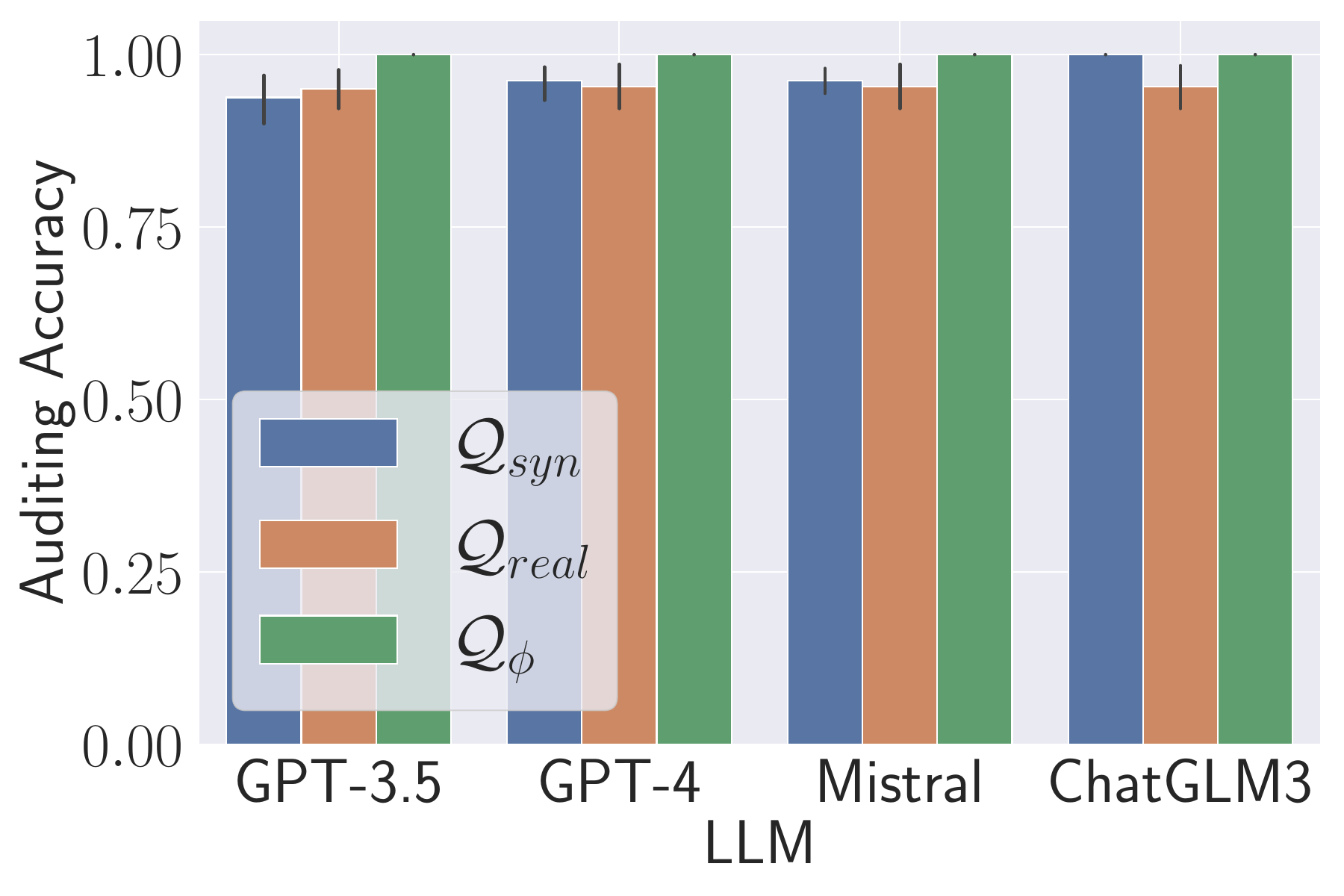}
\caption{\tone}
\label{figure:audit_classifier_barplot_t1_s1}
\end{subfigure}
\begin{subfigure}{0.45\columnwidth}
\includegraphics[width=\columnwidth]{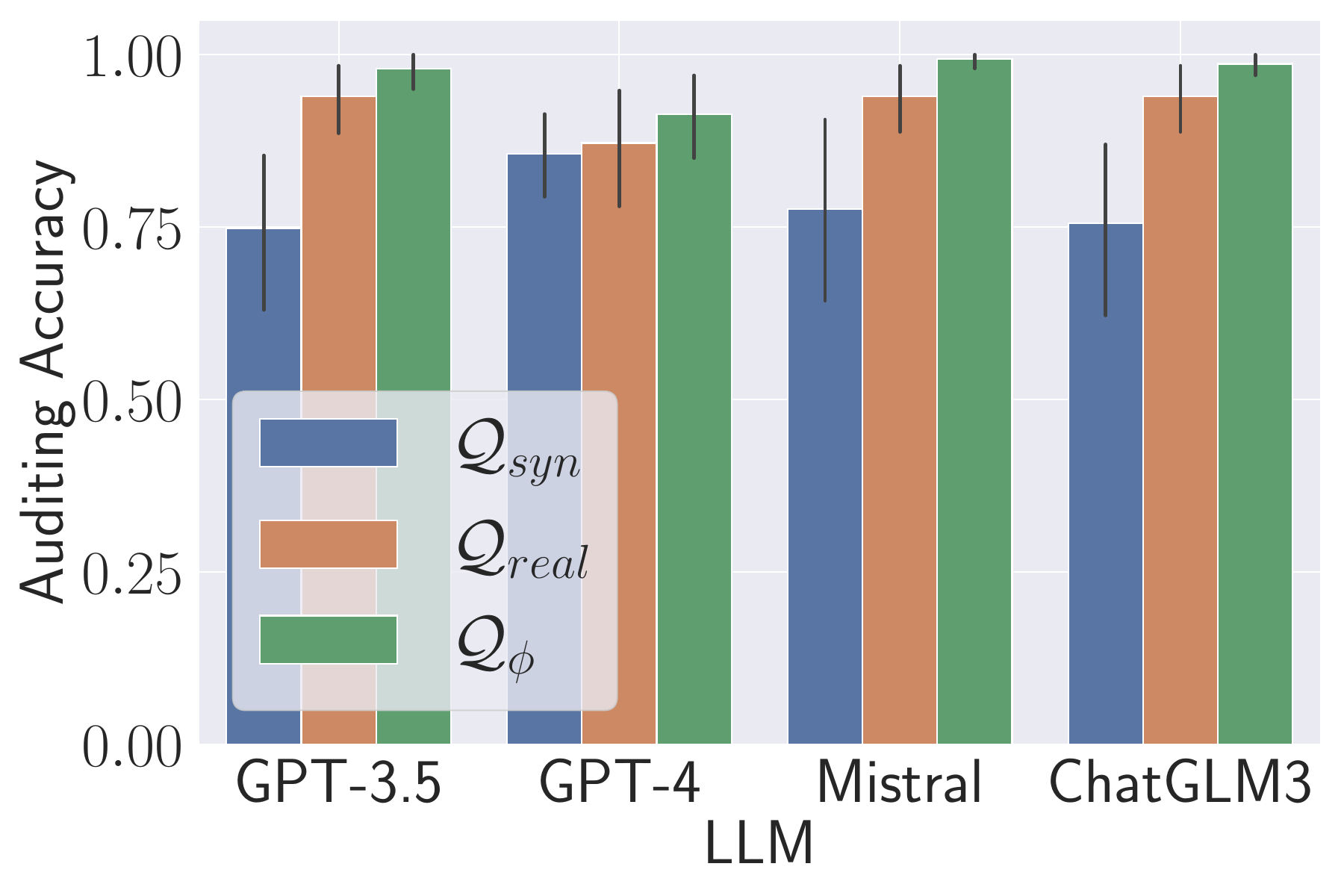}
\caption{\ttwo}
\label{figure:audit_classifier_barplot_t2_s1}
\end{subfigure}
\begin{subfigure}{0.45\columnwidth}
\includegraphics[width=\columnwidth]{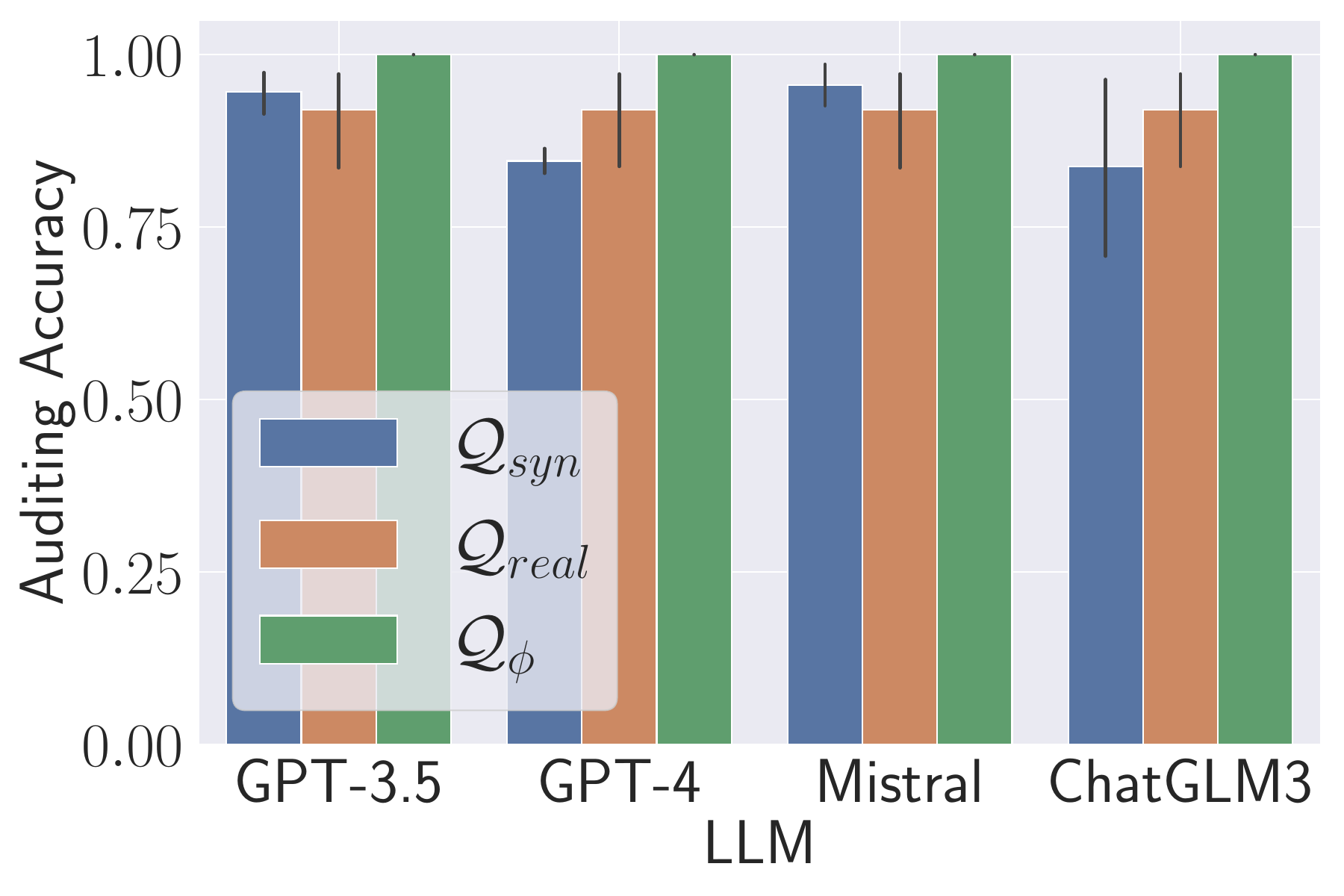}
\caption{\tthree}
\label{figure:audit_classifier_barplot_t3_s1}
\end{subfigure}
\caption{Auditing performance for target classifiers fine-tuned on the pre-trained DistilBERT model using metric-based auditing with \qr and \qs and tuning-based auditing with \qt across three tasks and four LLMs in \sone.
The auditing performance on the pre-trained BERT model is in~\refappendix{appendix:audit_classifier_bert_sup}.}
\label{figure:audit_classifier_barplot_s1}
\end{figure*}

\begin{figure*}[!t]
\centering
\begin{subfigure}{0.45\columnwidth}
\includegraphics[width=\columnwidth]{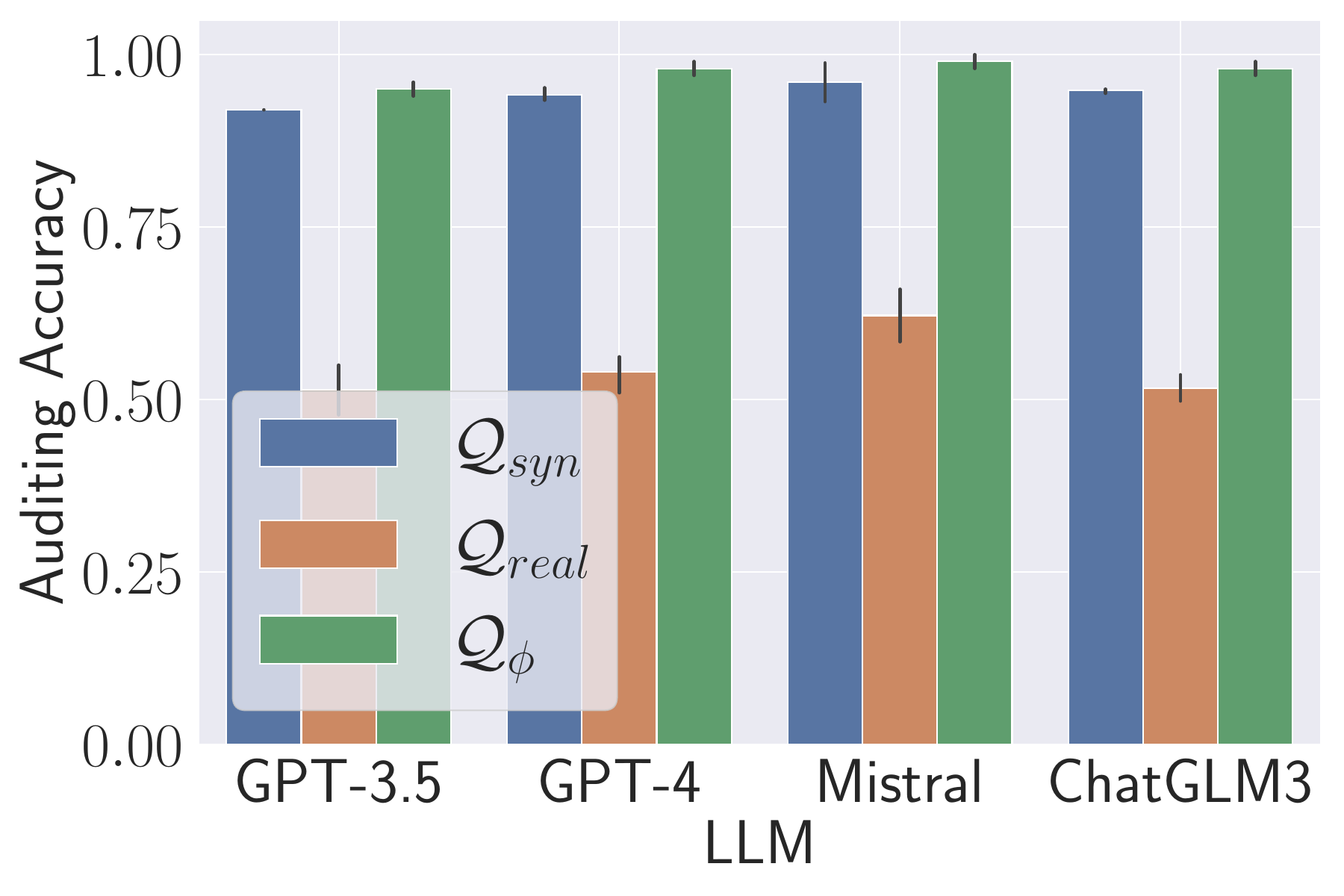}
\caption{\tone}
\label{figure:audit_classifier_barplot_t1_s2}
\end{subfigure}
\begin{subfigure}{0.45\columnwidth}
\includegraphics[width=\columnwidth]{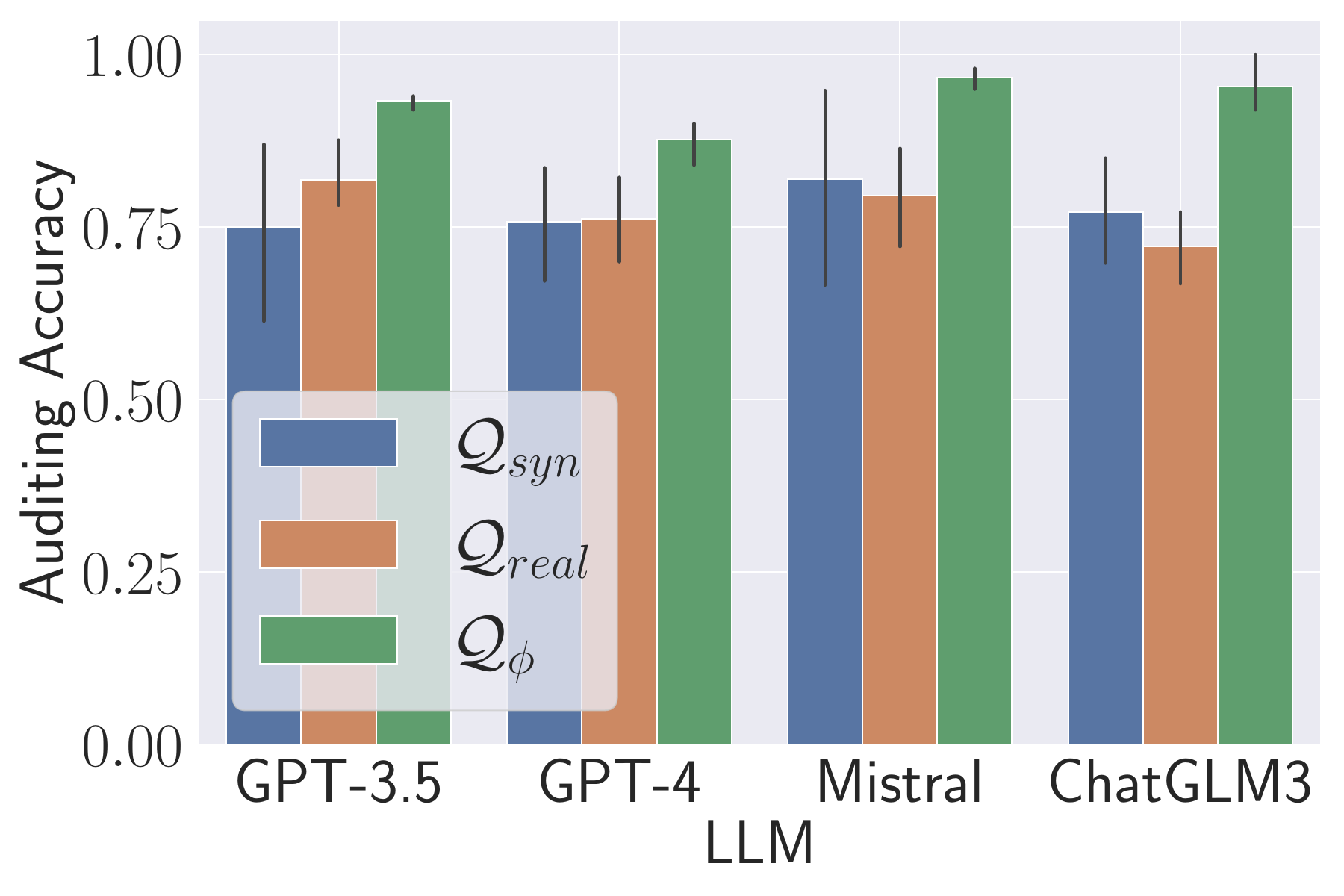}
\caption{\ttwo}
\label{figure:audit_classifier_barplot_t2_s2}
\end{subfigure}
\begin{subfigure}{0.45\columnwidth}
\includegraphics[width=\columnwidth]{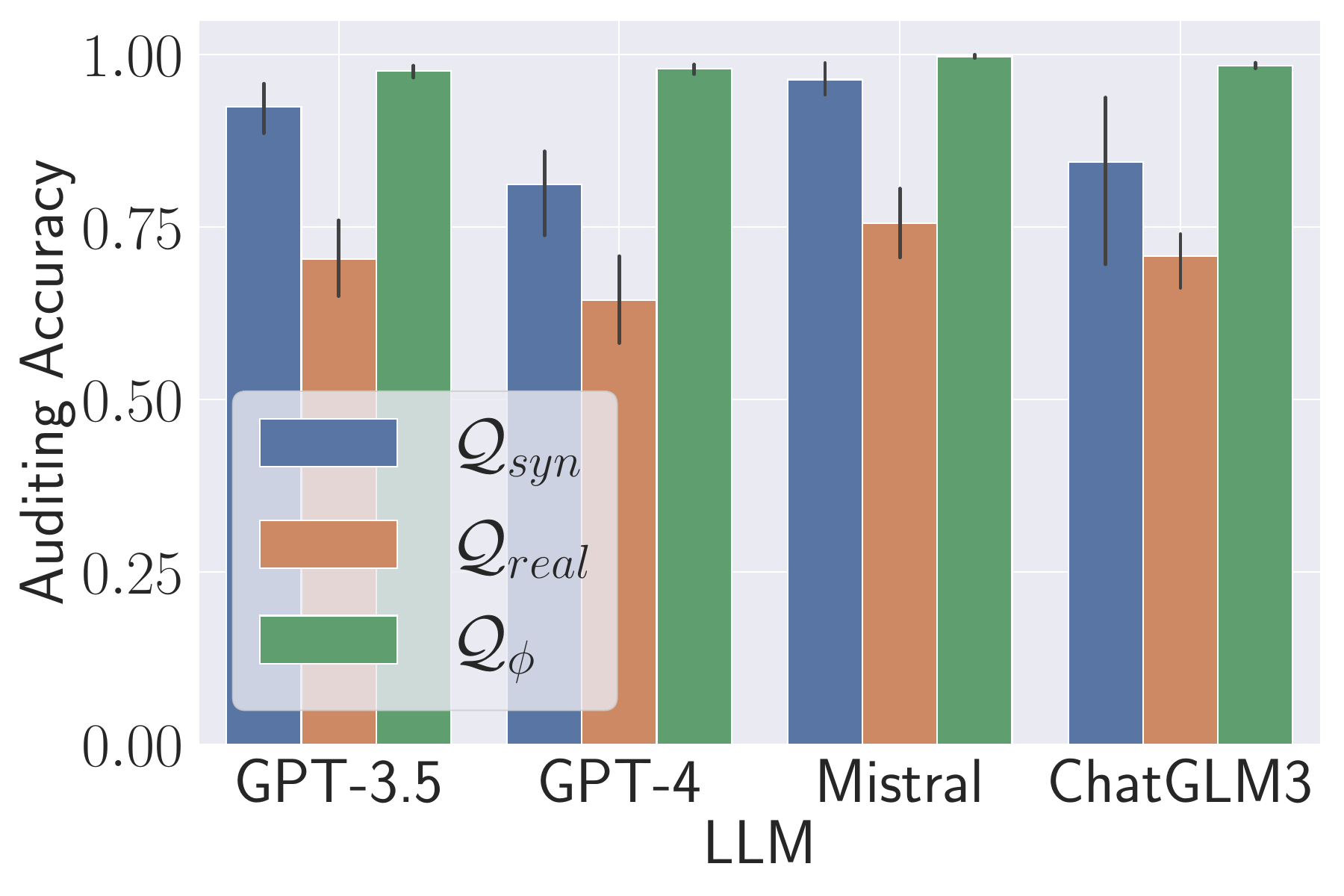}
\caption{\tthree}
\label{figure:audit_classifier_barplot_t3_s2}
\end{subfigure}
\caption{Auditing performance for target classifiers fine-tuned on the pre-trained DistilBERT model using metric-based auditing with \qr and \qs and tuning-based auditing with \qt across three tasks and four LLMs in \stwo.}
\label{figure:audit_classifier_barplot_s2}
\end{figure*}

With white-box access, we adopt a simple gradient-based approach to learn a query set \qt, i.e., embedding vectors, to feed into the target classifier.
In this section, we determine the appropriate number of tuned queries and reference classifiers to enable tuning-based auditing.
We present the auditing performance with varying numbers of reference classifiers in~\autoref{figure:audit_classifier_x_ns_tuning_t3}.
The size of the tuned query set is five, i.e., $|\mathcal{Q}_{\phi}| = 5$, since we demonstrate that this is sufficient to launch the tuning-based auditing in~\refappendix{appendix:tuning_based_audit_classifier_qt_sup}.
We maintain the numbers of reference real and synthetic classifiers the same, and the synthetic proportions of synthetic classifiers are the same as those in~\autoref{section:metric_qs_result}.
We observe that tuning-based auditing achieves strong performance.
For example, in \sone, the tuning-based auditing shows superior auditing performance, i.e., $1.000 \pm 0.000$, even with only 20 reference classifiers.
Meanwhile, in \stwo, we observe that more reference classifiers (e.g., 100 reference classifiers), meaning a larger training dataset lead to better tuning-based auditing performance.
Due to the space limitation, we present more results on other LLMs and tasks in~\refappendix{appendix:tuning_based_audit_classifier_qt_sup}, and similar conclusions can be drawn.

\subsection{Main Evaluation}
\label{section:classifier_main_result}

In this section, we evaluate three tasks, two pre-trained models, and three scenarios.
We set up the auditing method based on the results in previous sections.
For metric-based auditing (\qs and \qr), we set the query budget to 200.
The performance metrics used are Confidence for \qs and Entropy for \qr.
The reference classifiers include 10 reference real classifiers and 10 reference synthetic classifiers for each scenario.
In \sone, the reference synthetic classifiers have a 100\% synthetic proportion.
In \stwo, we select a reference synthetic classifier for each of the ten different proportions.
In \sthree, each reference synthetic classifier has a random proportion.
For tuning-based auditing (\qt), we learn five tuned queries, and the number of reference classifiers is set to 100, including 50 real and 50 synthetic classifiers for each scenario.
The synthetic proportions in \sone and \sthree are the same as those for the metric-based auditing.
In \stwo, we select five reference synthetic classifiers for each of the ten different proportions.

\mypara{Cost}
We mainly consider the cost of training reference classifiers.
Training a reference classifier for \tone, \ttwo, and \tthree costs 88.28 seconds, 65.16 seconds, and 69.45 seconds, respectively, across three training scenarios on average.
The corresponding costs of metric-based auditing (20 reference classifiers) for conducting training on a Google GCP A100 are \$1.82, \$1.34, and \$1.44, respectively, while those of tuning-based auditing (100 reference classifiers) are \$9.10, \$6.70, and \$7.20, respectively.

\mypara{Auditing Performance in \sone}
As shown in~\autoref{figure:audit_classifier_barplot_s1}, we observe that, in general, all proposed methods are effective across three tasks and four LLMs in \sone.
Especially for tuning-based auditing, it can achieve higher accuracy and a smaller standard deviation compared to metric-based auditing.
For example, it achieves an average accuracy of $0.989 \pm 0.009$ in \sone.
Metric-auditing using \qr, with an accuracy of $0.932 \pm 0.069$, follows behind yet outperforms \qs, which achieves an accuracy of $0.882 \pm 0.077$.

\mypara{Auditing Performance in \stwo}
As illustrated in~\autoref{figure:audit_classifier_barplot_s2}, we then report the auditing performance with the same evaluation setting in \stwo.
Tuning-based auditing still achieves the best performance, with an average accuracy of $0.962 \pm 0.017$.
In contrast, metric-based auditing with \qr shows a substantial decline in performance, while \qs demonstrates greater resilience, with its auditing accuracy in \stwo nearly matching that of \sone.
Specifically, \qr achieves an average accuracy of $0.675 \pm 0.060$ in \stwo, reflecting a notable decrease of 0.257 compared to its performance in \sone.
In contrast, \qs achieves an average accuracy of $0.868 \pm 0.077$ in \stwo, only a marginal decrease of 0.014 from \sone.
We speculate that this divergence in performance may be attributed to the inclusion of real data in the training dataset for synthetic classifiers in \stwo.
In \sone, the synthetic classifier (100\% synthetic proportion) and the real classifier (0\% synthetic proportion) exhibit clear behavior disparities with both \qs and \qr, thereby enabling good performance when using both types of queries.
Continuing with our earlier speculation, the decision boundary between different classes in the synthetic data is clearer than in the real data, facilitating easier feature-label correlation during classifier training.
Consequently, in \stwo, even a synthetic classifier with a 10\% synthetic proportion can make predictions on \qs with higher confidence compared to a real classifier, resulting in behavior disparities.
Nevertheless, a synthetic classifier with 10\% synthetic proportion, i.e., 90\% real data, and a real classifier (100\% real data) may have similar confidence levels when making predictions on \qr, resulting in negligible behavior disparities.

\begin{figure}[!t]
\centering
\includegraphics[width=0.75\columnwidth]{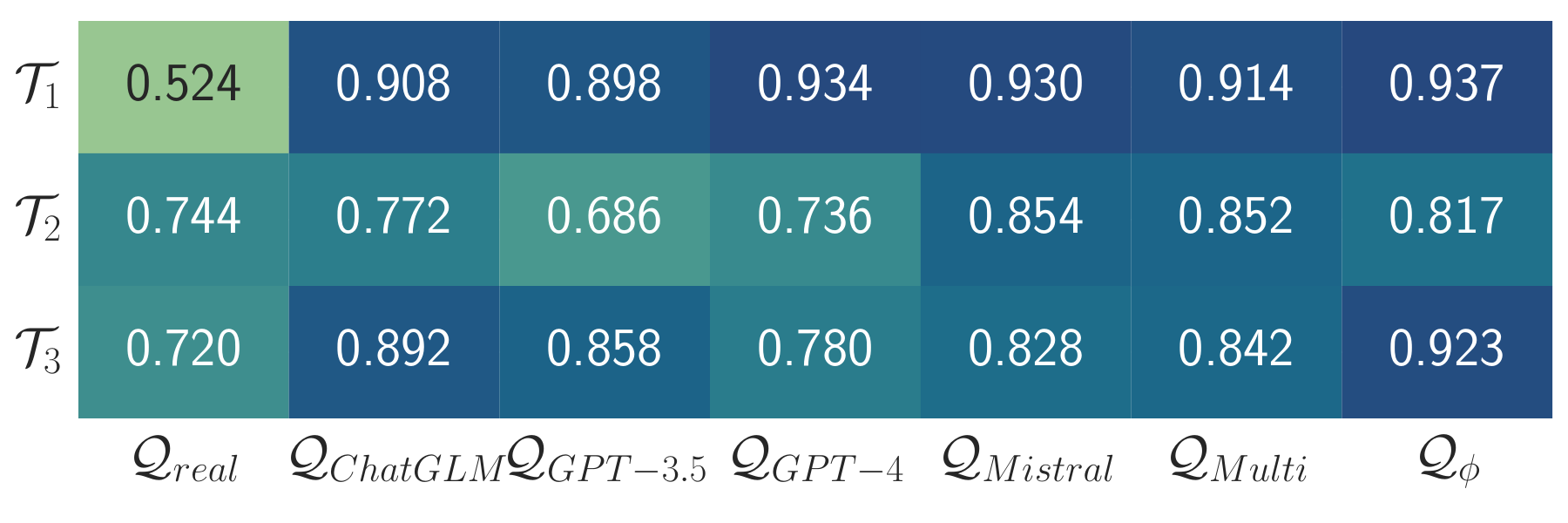}
\caption{Auditing performance for target classifiers fine-tuned on pre-trained DistilBERT across three tasks in \sthree.}
\label{figure:audit_classifier_heatmap_s3}
\end{figure}

\mypara{Auditing Performance in \sthree}
Finally, we report the auditing performance in \sthree.
Synthetic classifiers are trained on a combination of real and synthetic data from multiple sources (all four LLMs in our evaluation).
As such, we consider synthetic data from different source LLMs as the query to enable our metric-based auditing.
Specifically, we consider constructing the query set using \qr, \qt, and synthetic data solely from ChatGLM ($\mathcal{Q}_{\textit{ChatGLM}}$), GPT-3.5 ($\mathcal{Q}_{\textit{GPT-3.5}}$), GPT-4 ($\mathcal{Q}_{\textit{GPT-4}}$), Mistral ($\mathcal{Q}_{\textit{Mistral}}$), and a mix of random data from all LLMs ($\mathcal{Q}_{\textit{Multi}}$).
As shown in~\autoref{figure:audit_classifier_heatmap_s3}, tuning-based auditing outperforms metric-based auditing in most cases, achieving an average accuracy of $0.892 \pm 0.023$ on three tasks.
Metric-based auditing using $\mathcal{Q}_{\textit{Mistral}}$ and $\mathcal{Q}_{\textit{Multi}}$ follow closely behind.
The performance of \qr remains unsatisfactory, with an accuracy of only $0.663 \pm 0.061$.
We leverage $\mathcal{Q}_{\textit{Multi}}$ as the default setting for auditing synthetic classifiers with the synthetic query set, as it can achieve decent performance in various settings.
The results for another pre-trained model BERT are shown in~\refappendix{appendix:audit_classifier_bert_sup}, and similar conclusions can be drawn.

\mypara{Takeaways}
In general, tuning-based auditing achieves the best performance, with an average accuracy of $0.944 \pm 0.018$, in all three scenarios.
However, it requires white-box access to target classifiers, and the auditor needs extra training resources to train more reference classifiers and develop a meta-classifier.
If the auditor only has black-box access to target classifiers or limited training resources, we recommend utilizing metric-based auditing with \qs.
This non-NN-based auditing method requires fewer reference classifiers and achieves decent auditing performance, with an average accuracy of $0.868 \pm 0.071$ using 200 synthetic queries and 20 reference classifiers.

\section{Generator Auditing}
\label{section:generator_audit}

\subsection{Metric-Based Auditing}
\label{section:generator_metric_audit}

\mypara{Intuition}
The black-box auditing approach for generators shares similarities with auditing classifiers as discussed in~\autoref{section:classifier_metric_audit}.
In this study, we focus on text summarization generators.
The input text $x_i$ used for training these models is human-crafted, i.e., real, while the corresponding output summary $y_i$ can be synthetic.
We hypothesize that generators trained using real summaries, referred to as the \textit{real generator}, may outperform the \textit{synthetic generator} when generating summaries for real input text.
Based on this hypothesis, we utilize the real input text as the query and its corresponding real summary as the reference text (ground truth) to form the query set \qr.
We employ standard performance metrics for text summarization tasks to enable metric-based auditing.

\mypara{Methodology}
The metric-based auditing process for the generator is similar to the process for the classifier, as illustrated in \autoref{figure:metric_based_audit}.
Here, an auditor interacts with the target generator by submitting a query set and receiving the generated summary.
Subsequently, they calculate the performance metric using this generated summary alongside a reference text.
A predefined threshold is utilized to classify the target generator either as a \textit{synthetic generator} or a \textit{real generator}.
Formally, we define the metric-auditing for the target generator \gt using \qr as follows:
\begin{equation}
   \mathcal{I}( \mathcal{G}_{\textit{target}}, \mathcal{Q}_{\textit{real}}) =  \mathds{1}\{ f(\mathcal{G}_{\textit{target}}(x_i), y_i) < \tau, \forall (x_i, y_i) \in \mathcal{Q}_{\textit{real}}\},
\end{equation}
where $f(\cdot)$ denotes the performance metric.
The auditor empirically determines the threshold through comparisons with reference generators.
The processes of collecting training data, training the reference generators \gress and \grers, and selecting the final threshold are identical to those used when auditing classifiers in~\autoref{section:classifier_metric_audit}.

\mypara{Note}
Here, we do not consider tuning-based auditing.
The reasons are two-fold: (1) it is difficult to assign a reference text to a tuned query; 
(2) for a reference-free metric, such as perplexity~\cite{RWCLAS19,SCBZ23}, the meta classifier struggles to converge during training, as the generators only output placeholder tokens for this nonsensical query in the initial stage of training.

\subsection{Evaluation Setup}

We conduct two text summarization tasks on representative datasets: CNN/DM~\cite{SLM17} (\tfour) and XSum~\cite{NCL18} (\tfive).
We provide the specific details for these two tasks in~\refappendix{appendix:eval_setup_data_split_sup}.

\subsubsection{Target and Reference Generator Setup}

We primarily follow the setup in~\autoref{figure:overview_eval_setup}, including the same data split, paraphrasing-based synthetic data generation, and identical training scenarios.
We provide specific details in~\autoref{appendix:eval_setup_sup}.
We use the widely adopted pre-trained model BART~\cite{LLGGMLSZ20} as the backbone for \textit{target/reference generators} in both tasks.
We employ cross-entropy as the loss function and use Adam as the optimizer, with a learning rate of 2e-5.
The number of beams is set to 4.
We fine-tune the generators for 3 epochs.
Overall, we train a total of 50 target real generators, 200 target synthetic generators in \sone (50 per LLM), 200 target synthetic generators in \stwo, and 50 target synthetic generators in \sthree for each task.
We also train the same number of reference generators to determine the threshold values.
We ensure that target synthetic generators achieve performance comparable to target real generators (see details in~\refappendix{appendix:target_generator_performance}).

\begin{figure}[!t]
\centering
\begin{subfigure}{0.45\columnwidth}
\includegraphics[width=\columnwidth]{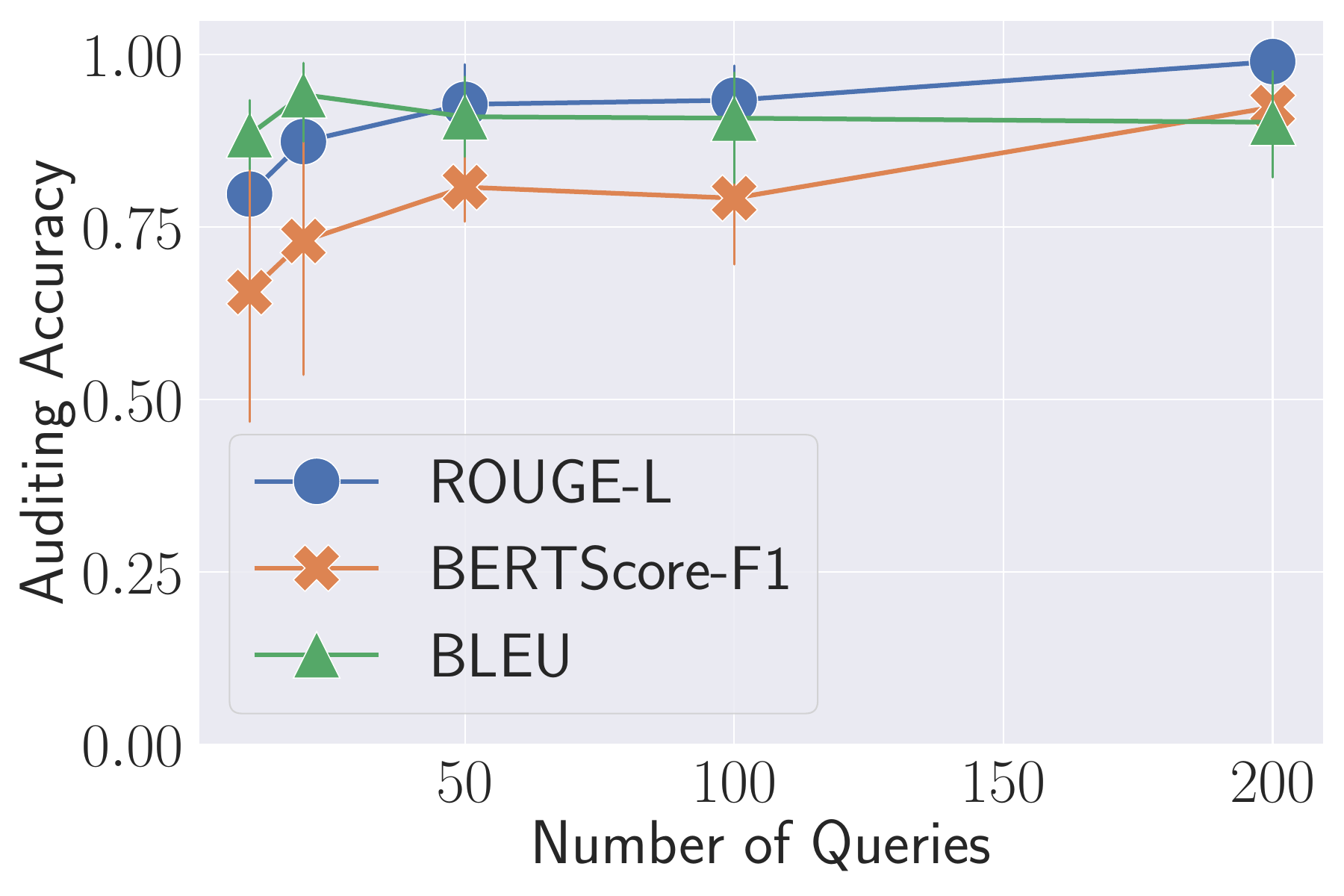}
\caption{\sone}
\label{figure:audit_generator_x_query_gpt3.5_s1}
\end{subfigure}
\begin{subfigure}{0.45\columnwidth}
\includegraphics[width=\columnwidth]{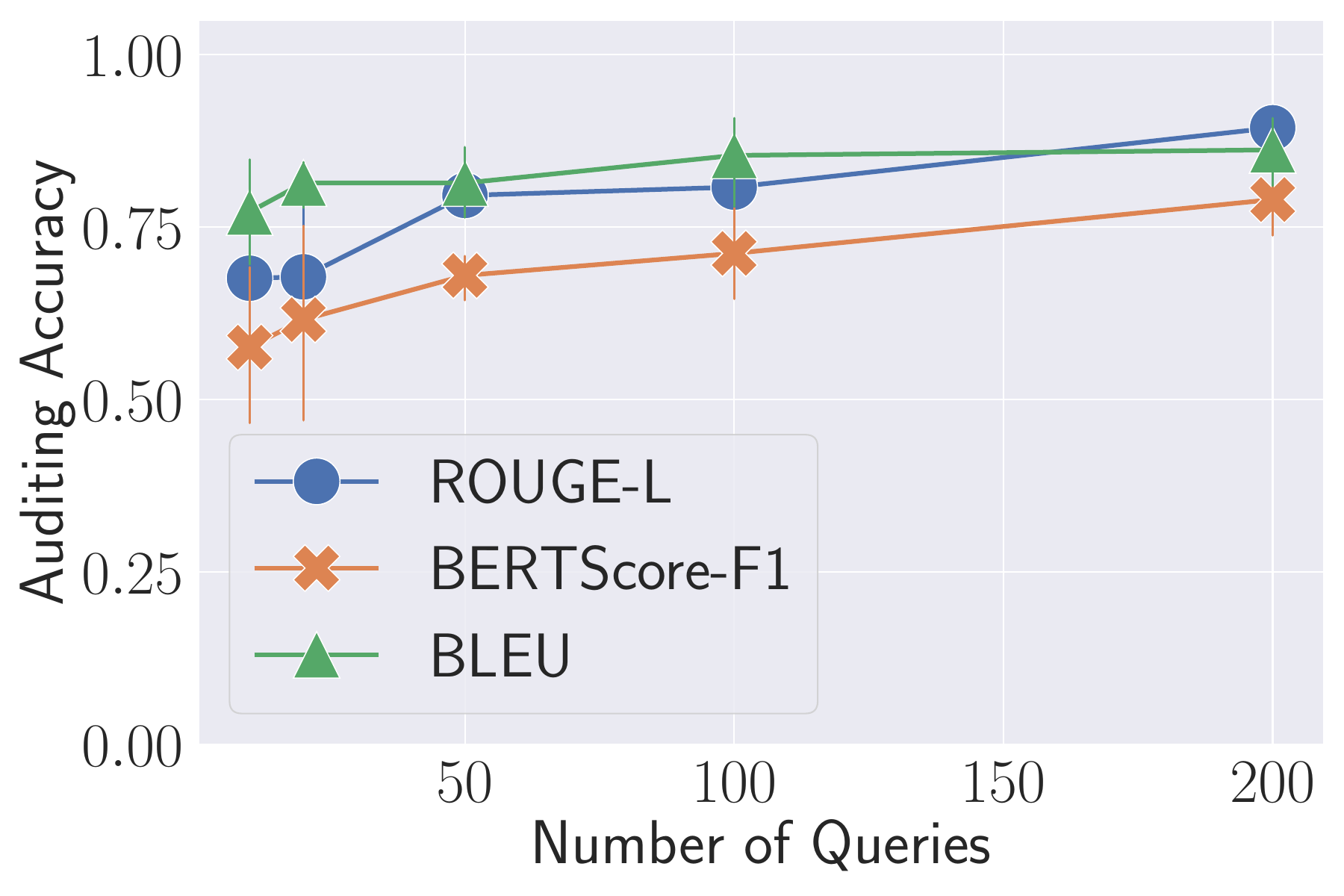}
\caption{\stwo}
\label{figure:audit_generator_x_query_gpt3.5_s2}
\end{subfigure}
\caption{Metric-based auditing performance for target generators fine-tuned on pre-trained BART with varying query budgets of \qr $\{10, 20, 50, 100, 200\}$ for \tfour in (a) \sone and (b) \stwo.
The source LLM of synthetic data is GPT-3.5.}
\label{figure:audit_generator_x_query}
\end{figure}

\subsubsection{Auditing Setup}

\mypara{Auditing Model}
We only consider metric-based auditing for generators.
Since it is a non-NN-based method, we calculate a performance metric value for each generator and use the reference generator set to determine a threshold value for this performance metric for auditing.
We select three widely used performance metrics BERTScore~\cite{ZKWWA20}, ROUGE~\cite{L04}, and BLEU~\cite{PRWZ02} to enable the metric-based auditing.
They all evaluate the quality of synthetic text relative to reference text but differ in focus.
BERTScore measures semantic similarity using contextual embeddings from models like BERT.
BLEU emphasizes precision, reflecting how much relevant information matches the reference, while ROUGE focuses on recall, indicating how much information from the reference is captured in the synthetic text.

\mypara{Summarization/Auditing Evaluation Metrics}
For summarization, we utilize the three standard text summarization evaluation metrics mentioned above.
A higher value of the metric indicates better performance.
For auditing, we balance the class distribution in target generators and reference generators, so we consider auditing accuracy on target generators as the main metric.
The target generators include 50 real generators and 50 synthetic generators for each scenario, and the synthetic proportions for each scenario are the same as those in~\autoref{section:classifier_eval_setup}.
Each experiment is run five times with different seeds and evaluated on all 100 target generators.
We present the average score with the error bar.

\subsection{Preliminary Investigation}

\begin{figure}[!t]
\centering
\begin{subfigure}{0.45\columnwidth}
\includegraphics[width=\columnwidth]{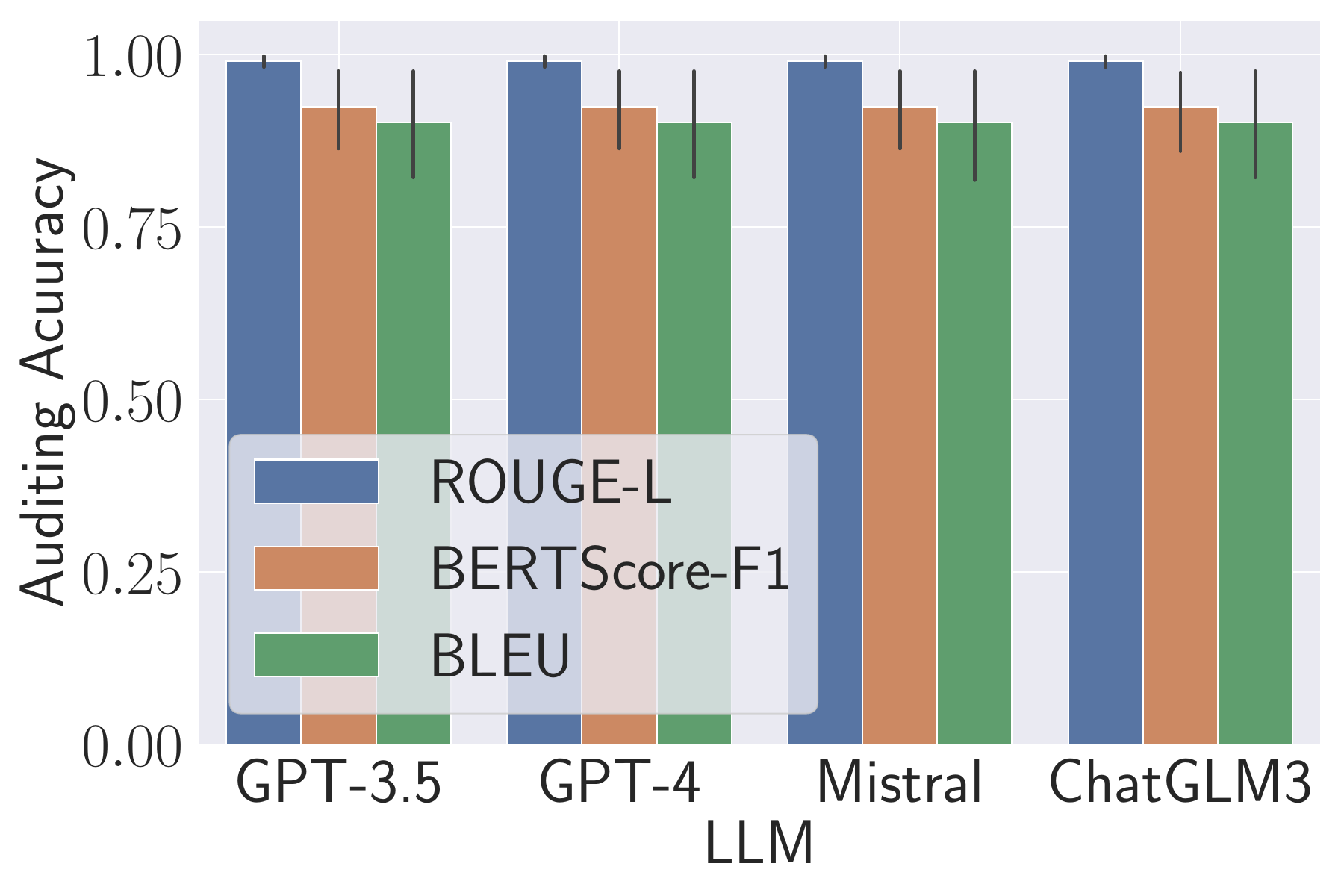}
\caption{\tfour}
\label{figure:audit_generator_barplot_t4_s1}
\end{subfigure}
\begin{subfigure}{0.45\columnwidth}
\includegraphics[width=\columnwidth]{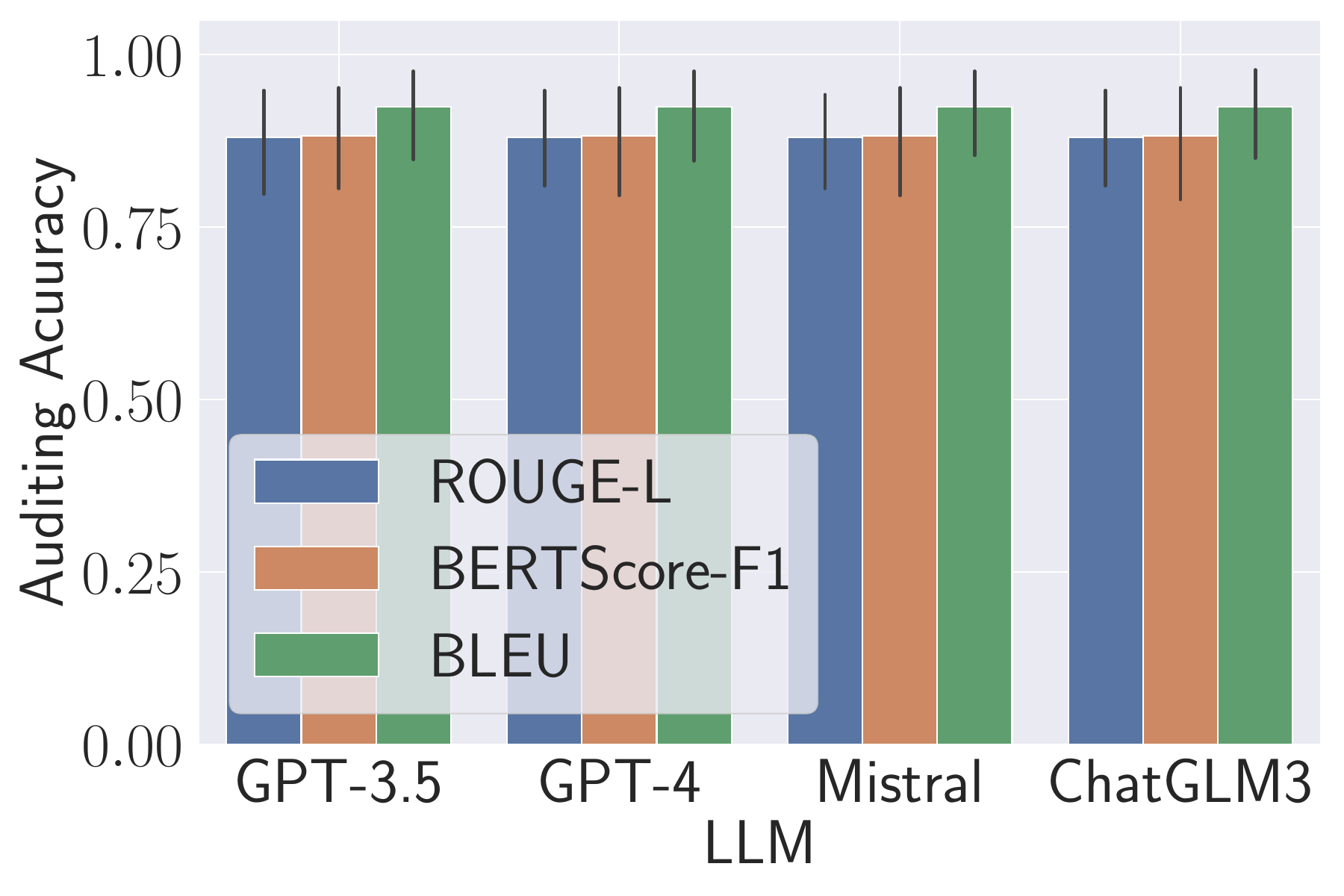}
\caption{\tfive}
\label{figure:audit_generator_barplot_t5_s1}
\end{subfigure}
\caption{Auditing performance for target generators fine-tuned on pre-trained BART using metric-based auditing with three metrics across two tasks and four source LLMs in \sone.}
\label{figure:audit_generator_barplot_s1}
\end{figure}

We investigate the appropriate query budget, the number of reference generators, and performance metrics to enable metric-based auditing with \qr.
For each scenario, we initially leverage 10 reference real generators and 10 reference synthetic generators, and the synthetic proportions are the same as those in~\autoref{section:metric_qs_result}.
We start by investigating auditing performance with varying query budgets.
As shown in~\autoref{figure:audit_generator_x_query}, the metric-based auditing with \qr achieves good performance even with 10 random real queries using all three metrics in \sone.
For example, using BLEU achieves an accuracy of $0.896 \pm 0.082$.
Meanwhile, we find that in both two scenarios, more real queries result in better auditing performance, i.e., higher accuracy and lower standard deviation.
For example, the metric-based auditing using ROUGE-L achieves $0.707 \pm 0.231$ with 10 queries in \sone, but it increases to $0.924 \pm 0.074$ with 200 queries, a large margin of $0.217$.
We demonstrate that 20 reference classifiers are sufficient to find a good threshold to achieve decent auditing performance, and the benefit of more reference generators is minimal in~\autoref{appendix:audit_generator_sup}.
Hence, we default to using all three metrics, 200 random real queries, and 20 reference classifiers to enable metric-based auditing with \qr.

\subsection{Main Evaluation}

We evaluate two tasks and three scenarios.
The query budget is set to 200.
We leverage 10 reference real generators and 10 reference synthetic generators.
The synthetic proportions for each scenario are the same as those in~\autoref{section:classifier_main_result}.
We report the auditing performance for two tasks in \sone in~\autoref{figure:audit_generator_barplot_s1}.

\mypara{Costs}
Training a reference generator for \tfour and \tfive costs 613.81 seconds, and 656.18 seconds across three training scenarios on average, respectively.
The corresponding total costs of metric-based auditing (20 reference generators) for conducting training on a Google GCP A100 are \$12.80 and \$13.60, respectively.

\begin{figure}[!t]
\centering
\begin{subfigure}{0.45\columnwidth}
\includegraphics[width=\columnwidth]{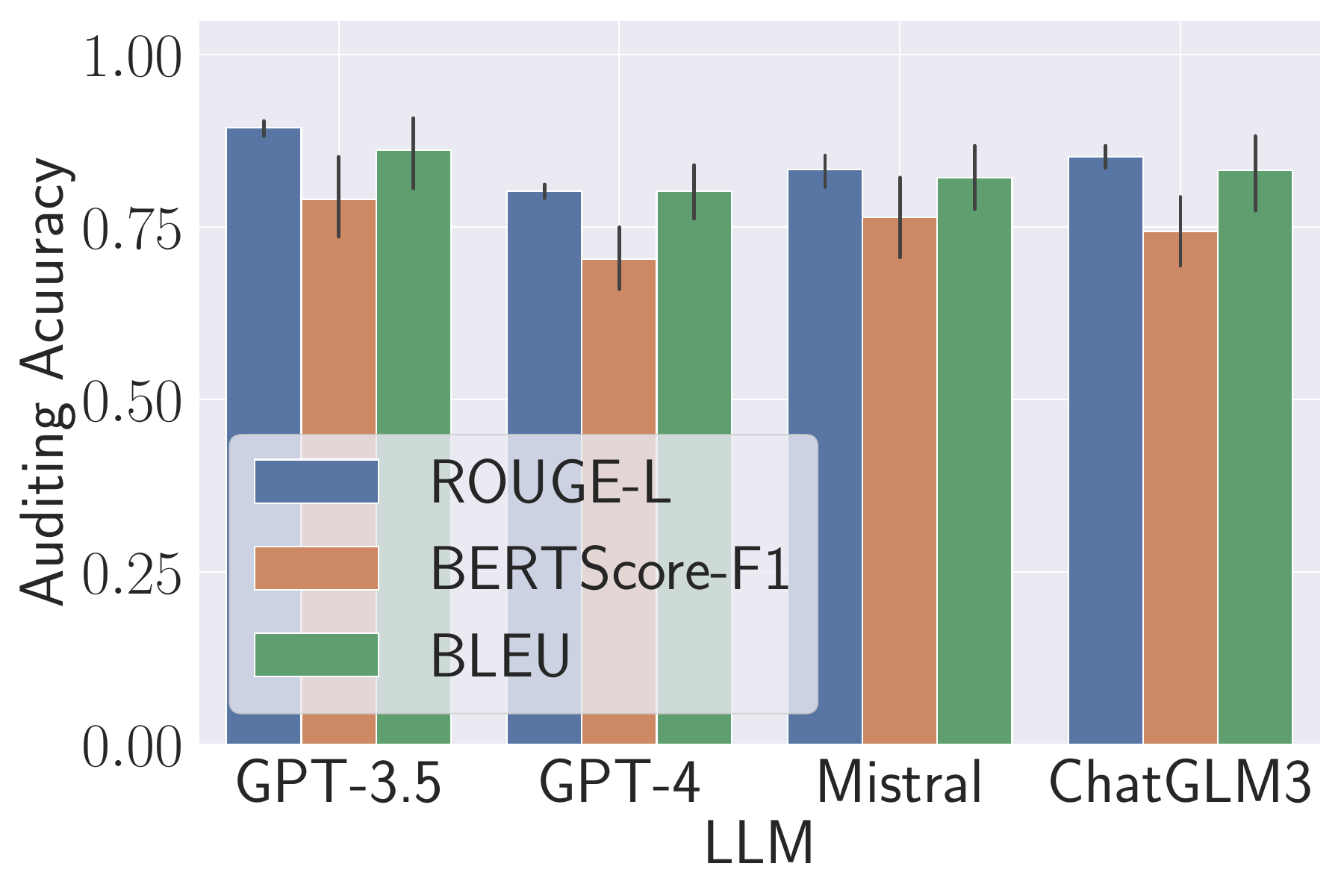}
\caption{\tfour}
\label{figure:audit_generator_barplot_t4_s2}
\end{subfigure}
\begin{subfigure}{0.45\columnwidth}
\includegraphics[width=\columnwidth]{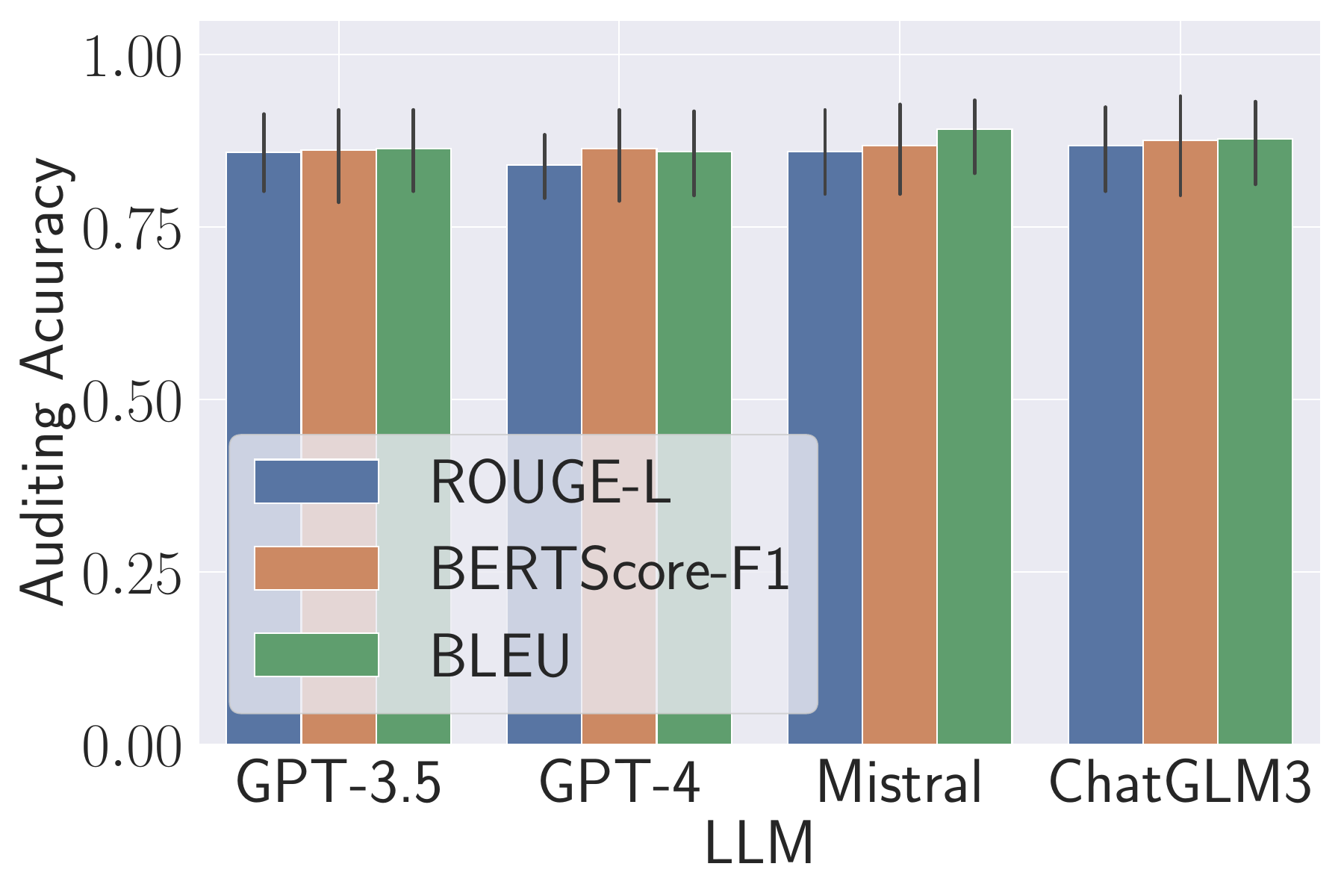}
\caption{\tfive}
\label{figure:audit_generator_barplot_t5_s2}
\end{subfigure}
\caption{Auditing performance for target generators fine-tuned on pre-trained BART using metric-based auditing with three metrics across two tasks and four source LLMs in \stwo.}
\label{figure:audit_generator_barplot_s2}
\end{figure}

\mypara{Results}
We observe that all metrics achieve decent auditing performance, indicating the effectiveness of our auditing methods.
For example, using ROUGE-L can achieve an average accuracy of $0.990 \pm 0.010$ in \tfour, and using BLEU can achieve an average accuracy of $0.902 \pm 0.100$ in \tfive.
Next, we present the auditing performance in \stwo (\autoref{figure:audit_generator_barplot_s2}) and \sthree (\autoref{table:audit_generator_main_s3}).
We observe that our metric-based auditing still achieves good performance, even targeting synthetic generators trained on a mix of synthetic data from multiple sources and real data.
Overall, these metrics consistently achieve good auditing performance across two tasks and three scenarios.
ROUGE-L achieves the best performance with an average accuracy of $0.880 \pm 0.052$.
BLEU follows closely, achieving an average accuracy of $0.877 \pm 0.081$, outperforms BERTScore-F1.

\begin{table}[!t]
\caption{Auditing performance for target generators using metric-based auditing with three metrics on two tasks in \sthree.}
\label{table:audit_generator_main_s3}
\centering
\scalebox{0.65}{
\begin{tabular}{ c |c c c}
\toprule
Task & ROUGE-L & BERTScore-F1 & BLEU \\
\midrule
\tfour & $0.780 \pm	0.025$ &  $0.734 \pm 0.065$ & $0.806 \pm 0.061$ \\
\tfive & $0.818 \pm	0.064$ & $0.840 \pm	0.076	$ & $0.852	 \pm 0.060$ \\
\bottomrule
\end{tabular}}
\end{table}

\mypara{Takeaways}
We demonstrate that the proposed metric-based auditing method using different performance metrics consistently achieves decent performance in all experiment settings.

\section{Statistical Plot Auditing}
\label{section:plot_audit}

\subsection{Classification-Based Auditing}

\mypara{Intuition}
Previous work~\cite{LYLL23} shows that real data and LLM-generated data from the same task represent different patterns in the statistical plots.
Intuitively, we consider developing a binary image classifier \mpc to determine whether the input of the given plot contains synthetic data.

\begin{figure}[!t]
\centering
\includegraphics[width=0.9\columnwidth]{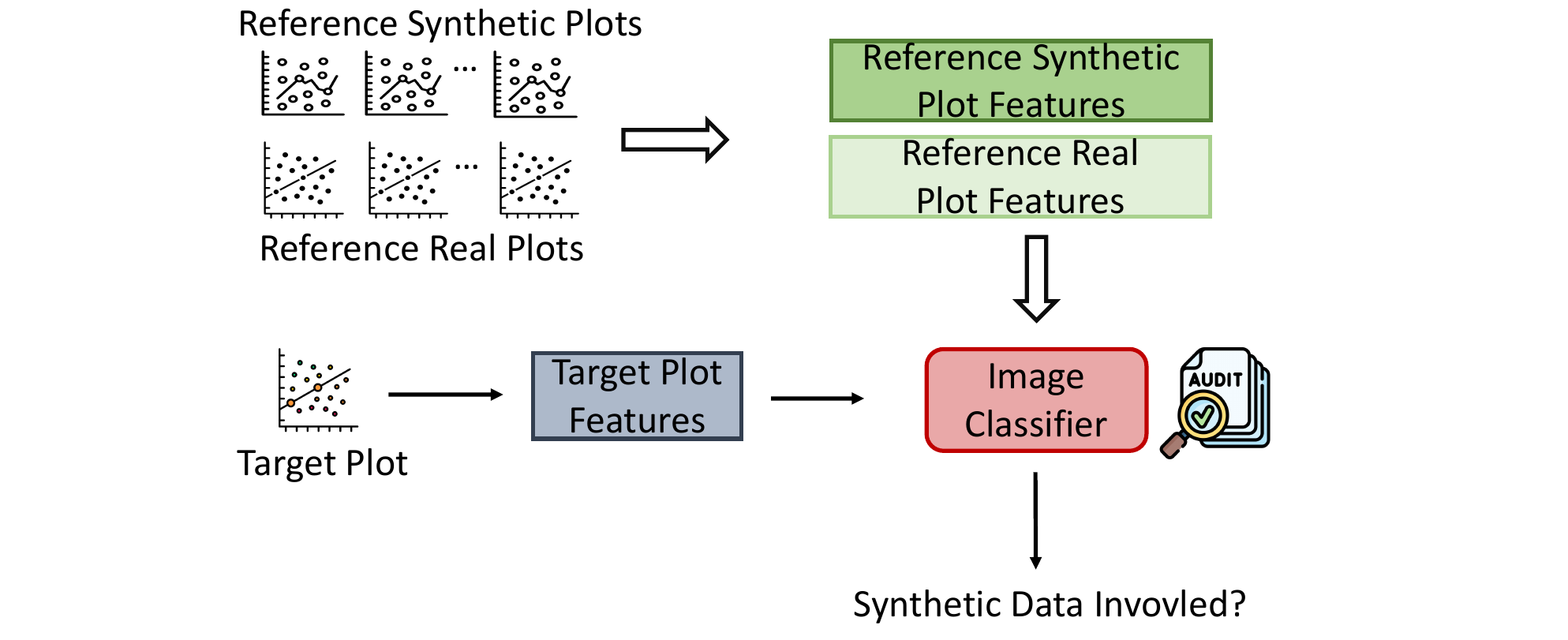}
\caption{Overview of the classification-based auditing.}
\label{figure:classification_based_audit}
\end{figure}

\mypara{Methodology}
The classification-based auditing approach for statistical plots is outlined in Figure \ref{figure:classification_based_audit}.
To construct the training dataset of \mpc, the auditor first generates a set of synthetic reference plots \press that contain synthetic data as input and real reference plots \prers that are derived solely from real data.
Specifically, we focus on data visualization, i.e., t-distributed Stochastic Neighbor Embedding (t-SNE)~\cite{MH08}, on the text classification datasets.
We directly leverage the synthetic data and real data in~\autoref{section:classifier_metric_audit} as the input data to generate these reference plots.
The composition of synthetic versus real data within each \pres can vary.
This allows plots to be generated solely from synthetic data or from a blend of synthetic and real data in random proportions.
We train the image classifier \mpc parameterized by $\omega_2$  via optimizing the following loss function:
\begin{equation} 
    \mathcal{L} = \sum_{i=1}^{k} \mathbf{loss}(1, \mathcal{M}_{\omega_2}(\mathcal{P}^{\textit{syn}}_{\textit{ref}, i})) + \sum_{i=1}^{k} \mathbf{loss}(0, \mathcal{M}_{\omega_2}(\mathcal{P}^{\textit{real}}_{\textit{ref}, i})).
\end{equation}
At inference time, the auditor determines a given plot by querying \mpc with it and obtaining the prediction result.

\subsection{Evaluation Setup}
\label{section:plot_setup}

We conduct two t-distributed Stochastic Neighbor Embedding (t-SNE) visualization~\cite{MH08} tasks.
The task \tsix employs the IMDB dataset from \tone and generates the synthetic data through a zero-shot prompt strategy.
The task \tseven employs the AG dataset from \ttwo and generates the synthetic data using a paraphrasing strategy.

\subsubsection{Target and Reference Plot Setup}

We primarily follow the setup in~\autoref{figure:overview_eval_setup}, including the same data split, the synthetic data generation settings of \tone and \ttwo from~\autoref{section:classifier_eval_setup} to \tsix and \tseven, and the same training scenario.
Standard procedures for generating \textit{target/reference t-SNE plots} are adopted based on established practices \cite{ZHSWZ23}.
Specifically, we first preprocess all input texts by removing stopwords and punctuation.
We then leverage two representative methods -- Word2Vec~\cite{MCCD13} and GloVe~\cite{PSM14} -- to create word embeddings.
Subsequently, we use t-SNE to visualize these embeddings in two-dimensional space.
The resulting plots are saved as 300$\times$300 PNG images, showing only the scattered data points without axes, labels, or titles.
Different colors are employed to indicate the target labels assigned to each data instance in the text classification task.
Overall, we generate a total of 200 target real plots, 800 target synthetic plots in \sone (200 per LLM), 800 target synthetic plots in \stwo, and 200 target synthetic plots in \sthree for each task.
We develop the same number of reference plots to train the auditing model.

\begin{figure}[!t]
\centering
\begin{subfigure}{0.75\columnwidth}
\includegraphics[width=\columnwidth]{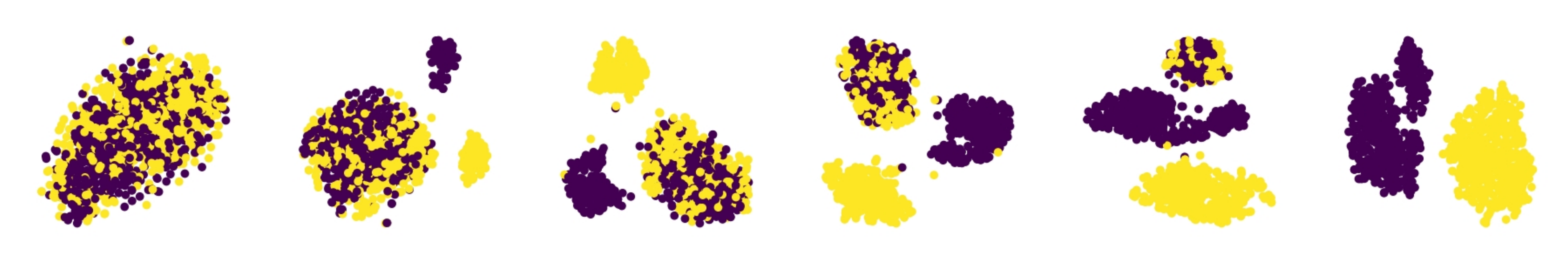}
\caption{\tsix}
\label{figure:tsne_plot_t6}
\end{subfigure}
\begin{subfigure}{0.75\columnwidth}
\includegraphics[width=\columnwidth]{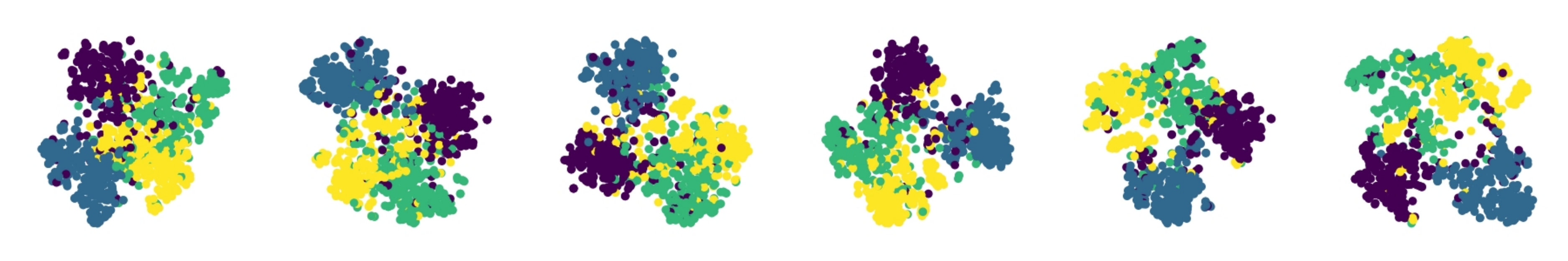}
\caption{\tseven}
\label{figure:tsne_plot_t7}
\end{subfigure}
\caption{
T-SNE plots using Word2Vec with synthetic proportions of input data are set at intervals of 20\%, ranging from 0 to 100\% (left to right) in (a) \tsix and (b) \tseven.
The synthetic data for \tsix and \tseven are generated using zero-shot and paraphrasing prompt strategies, respectively.
Different colors denote the target labels in the text classification task.}
\label{figure:tsne_plot_word2vec}
\end{figure}

\subsubsection{Auditing Setup}

\mypara{Auditing Model}
It is essentially an image classifier.
We leverage the pre-trained RN18~\cite{HZRS16} as the backbone of the image classifier \mpc.
We fit a linear classifier on top of the pre-trained RN18 to conduct synthetic artifact auditing.
We employ cross-entropy as the loss function and Adam as the optimizer with a learning rate of 1e-3.
The model is trained for 50 epochs on the reference plots set.
Note that we convert t-SNE plots into grayscale to eliminate the possibility that the auditing relies on differences in color schemes.

\mypara{Auditing Evaluation Protocol}
We use test accuracy on all target plots as the key metric to assess auditing performance.
The target plots include 200 target real plots and 200 target synthetic plots for each scenario, and the synthetic proportions are the same as those in~\autoref{section:classifier_eval_setup}.
Each experiment is run five times with different seeds, and we present the average score along with the error bar.

\subsection{Main Evaluation}
\label{section:audit_plot_eval}

\autoref{figure:tsne_plot_word2vec} shows examples of t-SNE plots for \tsix and \tseven.
From left to right, the proportion of synthetic data increases from 0 to 100\% at intervals of 10\%.
As shown in~\autoref{figure:tsne_plot_t6}, when we leverage a zero-shot prompt strategy to generate synthetic data, there are clear separations between real data and synthetic data in the reduced-dimension space created by t-SNE.
However, when we leverage a paraphrasing prompt strategy (\tseven), the synthetic data are scattered and intertwined with real data, as shown in~\autoref{figure:tsne_plot_t7}, making the auditing challenging.
This observation is consistent with previous work~\cite{BWBVN24,MLKMF23,LYLL23}.
We further observe that as the proportion of synthetic data increases, the distinction between data samples from different classes becomes more pronounced, leading to a more distinct decision boundary.
We attribute the clearer decision boundary among synthetic data of different classes to its generation based on their labels as conditions, which results in features that highly represent the target class.
Meanwhile, this clearer decision boundary distinguishes the real and synthetic data in the reduced-dimension space and thus motivates us to exploit the distinct patterns to conduct the synthetic artifact auditing.

We report the auditing performance for plots using Word2Vec in~\autoref{table:audit_plot_main}.
The reference plots include 200 reference real plots and 200 reference synthetic plots for each scenario, and the synthetic proportions are the same as those for target plots in~\autoref{section:plot_setup}.
The classification-based auditing achieves a superior performance in both tasks.
 It can achieve an average accuracy of $0.990 \pm 0.001$ on \tsix and $0.942 \pm 0.006$ on \tseven across three scenarios.
This demonstrates that although synthetic data for this task is generated through a paraphrasing prompt strategy and highly overlaps with real data, we can still successfully conduct auditing by distinguishing the differences in the decision boundaries between data of different classes.
We also exhibit some examples of t-SNE plots using GloVe and the auditing performance for these plots in~\refappendix{appendix:audit_plot_sup} with similar conclusions.

\begin{table}[!t]
\caption{Auditing performance for target plots across two tasks and four LLMs in three scenarios.}
\label{table:audit_plot_main}
\centering
\scalebox{0.65}{
\begin{tabular}{ c| c |c c c c}
\toprule
 \multirow{2}{*}{Scenario}  & \multirow{2}{*}{Task} & \multicolumn{4}{c}{LLMs} \\
 \cline{3-6}
  & & GPT-3.5 & GPT-4  & Mistral & ChatGLM3 \\
\midrule
 \multirow{2}{*}{\sone}  &  \tsix & $1.000 \pm 0.000$ &     $1.000 \pm 0.000$ & $1.000 \pm 0.000$ & $1.000 \pm 0.000$ \\
 &   \tseven & $1.000 \pm 0.000$ & $0.899 \pm 0.018$ & $1.000 \pm 0.000$ & $1.000 \pm 0.000$   \\
  \midrule
  \multirow{2}{*}{\stwo}  &  \tsix &    $1.000 \pm 0.000$ &     $1.000 \pm 0.000$ & $0.999 \pm 0.001$ &     $0.931 \pm 0.004$   \\
 &   \tseven &    $0.927 \pm 0.006$ &  $0.866 \pm 0.012$ & $0.956 \pm 0.004$ &     $0.945 \pm 0.003$  \\
 \midrule
  \multirow{2}{*}{\sthree}  &  \tsix &  \multicolumn{4}{c}{$0.976 \pm 0.002$}   \\
 &   \tseven &   \multicolumn{4}{c}{$0.882 \pm 0.009$}  \\
\bottomrule
\end{tabular}}
\end{table}

\mypara{Takeaways}
We demonstrate that synthetic data, whether generated through zero-shot or paraphrasing, exhibit clear differences from real data, i.e., the decision boundary of data samples of different classes, and these differences result in distinct patterns on statistical plots.
This signal can facilitate the synthetic artifact auditing of t-SNE plots.

\section{Related Work}
\label{section:related_work}

\mypara{Synthetic Data Detection}
This task can be formulated as a classification problem that distinguishes texts generated by language models (i.e., synthetic data) from those authored by humans (i.e., real data)~\cite{NTNYE17, GSR19, FZ21, GZWJNDYW23, MLKMF23, LYLL23, WMISSTAMPAAHGN24, BWBVN24}.
These methods address the distinctions between synthetic and real data by exploiting their different characteristics.
Recent studies~\cite{LYLL23,HSCBZ24} show that LLM-generated synthetic data has unique lexical, structural, and semantic features that distinguish it from real data.
There are fine-tuning-based methods~\cite{CKZLSR23, LYLL23, WMISSTAMPAAHGN24, BWBVN24} that analyze texts' latent features and train classifiers to identify synthetic data.
The differences between synthetic and real data, along with the success of classifiers in identifying synthetic data, inspire us to propose a hypothesis.
Classifiers trained on synthetic data tend to be more confident with synthetic inputs but less confident with real inputs, as they memorize the latent patterns of synthetic data.
We then propose a metric-based auditing method grounded in this hypothesis (\autoref{section:classifier_metric_audit}) and demonstrate its effectiveness through evaluation (\autoref{section:classifier_main_result}).

\mypara{Membership Inference Attacks (MIAs)~\cite{SSSS17,SZHBFB19,NSH18,LZ21,CCNSTT22}}
Although both MIAs and synthetic artifact auditing employ binary classification and leverage classifier outputs (confidence scores, entropy, posteriors), they differ fundamentally in their attack targets and goals.
MIAs target a given sample's membership, aiming to detect whether a specific data sample was used during training.
In contrast, we target trained classifiers, generators, or plots, aiming to distinguish between artifacts trained on or derived from real versus synthetic data.
Additionally, our attack processes diverge: MIAs train shadow models solely to mimic the target model's behavior, while tuning-based auditing trains reference artifacts to optimize queries.
MIAs use specific queries to infer their membership, whereas we employ optimized (tuning-based) or random (metric-based) queries to audit target artifacts.

\section{Discussion}
\label{section:discussion}

\mypara{Data Contamination}
The PLMs used in our evaluation, i.e., BERT, BART, and DistilBERT, were trained on a corpus consisting of Wikipedia (2,500 million words) and Google's BookCorpus (800 million words) and released in 2019.
Their pretraining occurred before the release of ChatGPT in 2022, which facilitated the generation of synthetic data at a massive scale.
Given this timeline, we argue that it is improbable that they incorporated synthetic data in the pre-training process.

\begin{figure}[!t]
\centering
\begin{subfigure}{0.45\columnwidth}
\includegraphics[width=\columnwidth]{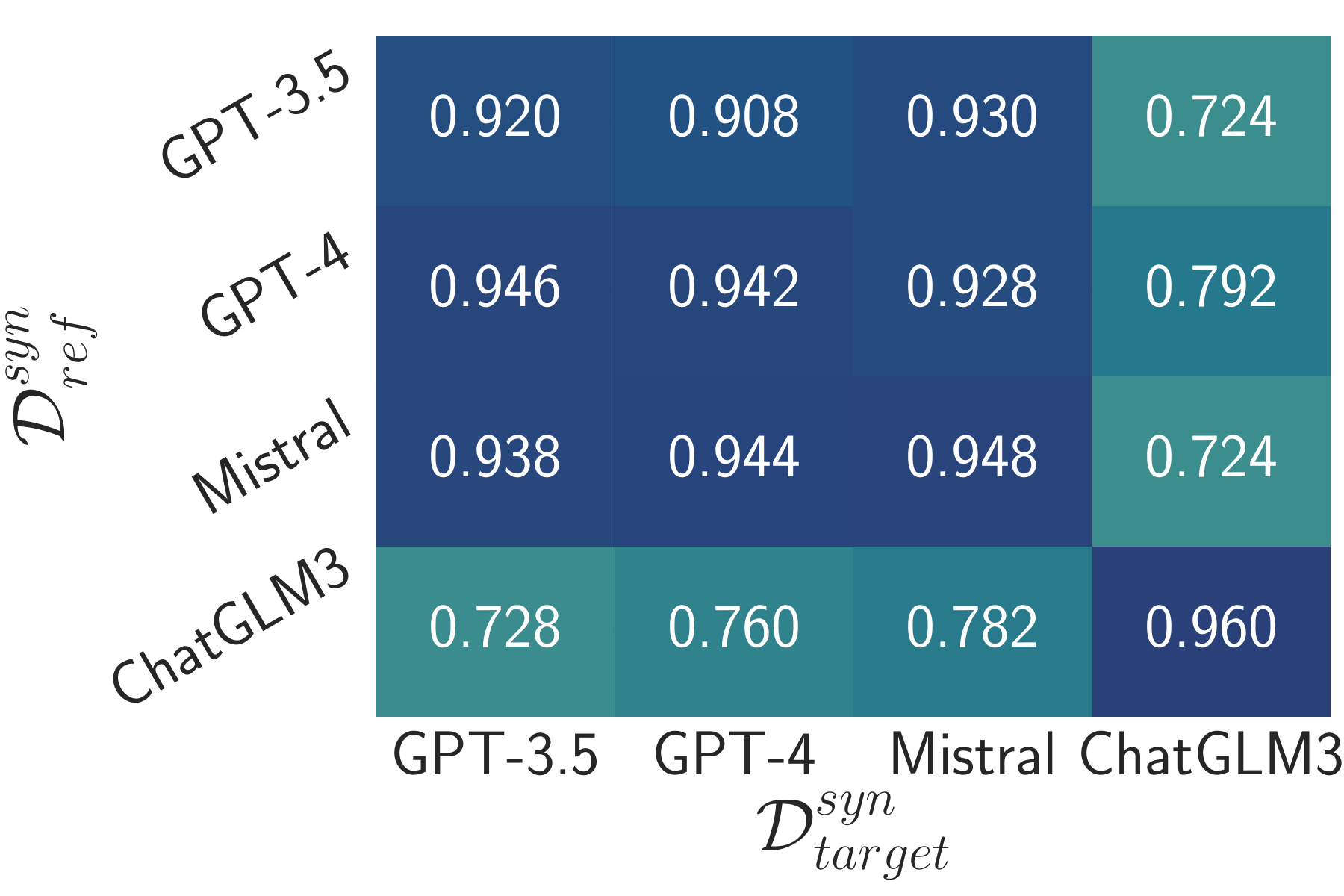}
\caption{IMDB}
\label{figure:sdimdb_tdimdb_t3}
\end{subfigure}
\begin{subfigure}{0.45\columnwidth}
\includegraphics[width=\columnwidth]{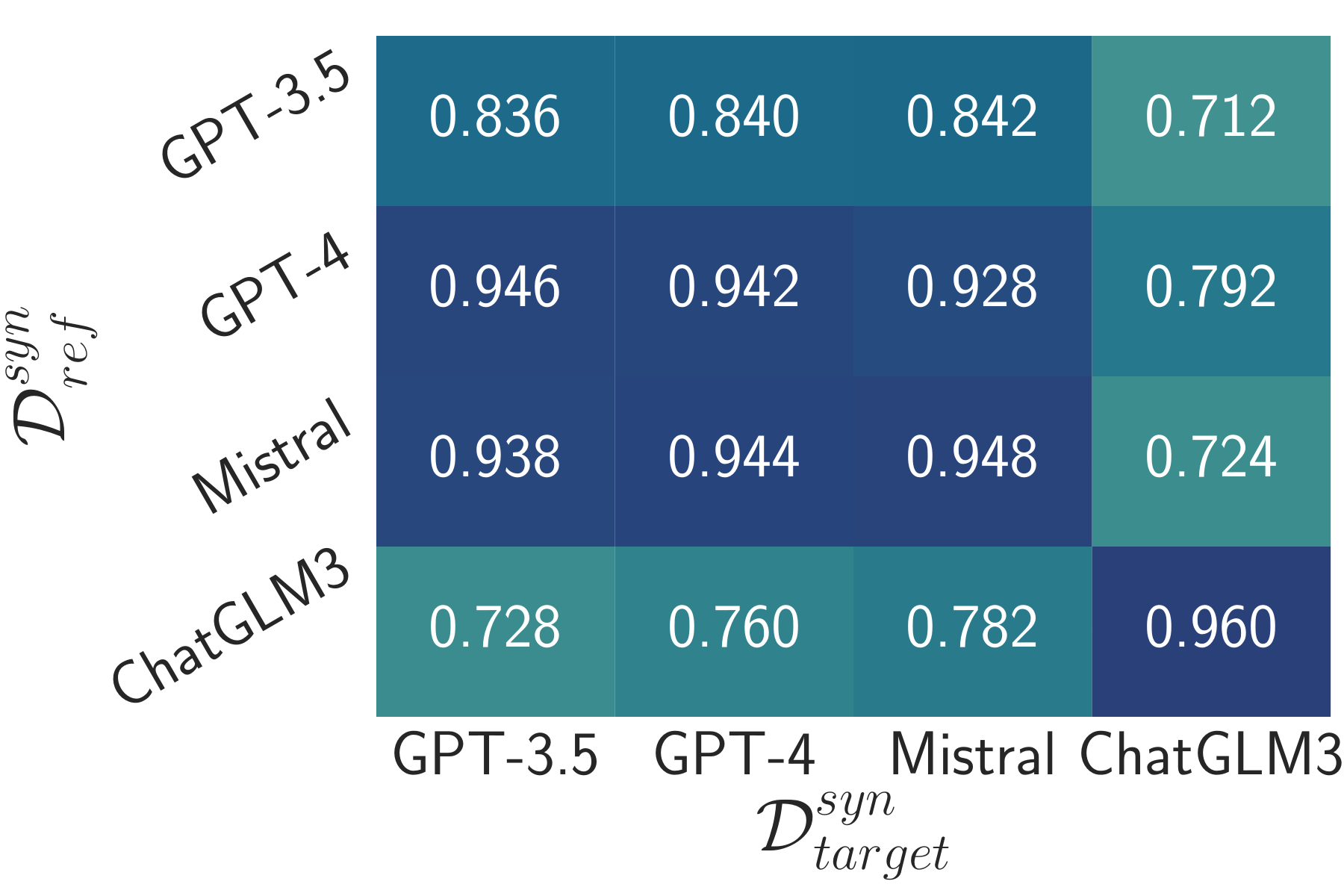}
\caption{Rotten Tomatoes}
\label{figure:sdtomato_tdimdb_t3}
\end{subfigure}
\caption{Auditing performance on \tone with different source LLMs in \stwo.
\dtr is derived from IMDB.
\drr is (a) disjoint data from IMDB, and (b) data from Rotten Tomatoes.}
\label{figure:audit_classifier_diff_dist_s2}
\end{figure}

\mypara{Practicability}
We further explore more practical scenarios, and the experimental settings are consistent with those detailed in~\autoref{section:classifier_main_result}.
First, we explore the scenarios where \drs and \dts are generated from different source LLMs.
As shown in~\autoref{figure:audit_classifier_diff_dist_s2},  our auditing framework maintains comparable performance when the source LLMs differ in most cases.
We hypothesize that this outcome arises due to the inherent similarities in the synthetic data produced by different LLMs.
Therefore, we recommend that the auditors incorporate multiple source LLMs when constructing \drs.
Additionally, we observe that even when the distributions of 
\drr and \dtr differ (Rotten Tomatoes~\cite{PL05} for \drr and IMDB for \dtr), our approach still achieves performance comparable to when the distributions are the same (see more details in~\refappendix{appendix:ablation_study}).
We then evaluate the impact of the size of the reference dataset on training each classifier.
Specifically, we conduct an experiment where each reference classifier is trained using one-third of the original dataset size.
In this setting, the auditing accuracy is 0.870, compared to 0.938 in the original setting with \tone and \sone, suggesting that the dataset size could potentially be reduced further.
Furthermore, we validate that our methods can distinguish artifacts from different LLMs (GPT-3.5, GPT-4, Mistral, and ChatGLM).
For example, in \tone and \sone, tuning-based auditing achieves 0.945 accuracy on this four-class classification task.
These results indicate that our methods could be potentially used to infer the unauthorized use of LLM-generated data to develop competitive artifacts, which is often prohibited by tech giants~\cite{OPENAI_LICENSE,LLAMA_LICENSE}.

\section{Limitations}
\label{section:limitation}

\mypara{Design Choices}
We take the initial step to introduce synthetic artifact auditing and propose an auditing framework with three methods.
For simplicity, we aim to present the intuition behind our methods in straightforward terms and develop auditing processes using simple yet effective design choices, such as using random synthetic queries.
Our method, especially metric-based auditing, is flexible, with many different design possibilities yet to be explored.
For instance, the methodology may incorporate mixing synthetic and real queries, and leveraging other performance metrics, e.g., METEOR~\cite{BL05} for generators.
We plan to explore additional design options and further extend our auditing framework in future research.

\mypara{Evaluation on LLMs And Additional Tasks}
Evaluating LLMs requires training reference LLMs from scratch, but this is impractical due to infrastructure constraints.
Our work remains valuable, as small models are still widely used for their efficiency, cost-effectiveness, and flexibility.
Furthermore, we empirically demonstrate that synthetic artifacts capture the unique patterns of synthetic data, distinguishing them from real artifacts.
As a result, our methods are expected to be feasible for LLMs and generalizable to other tasks.

\section{Conclusion}

In this paper, we introduce the concept of synthetic artifact auditing.
We propose an auditing framework with three methods that require no disclosure of proprietary training specifics: metric-based auditing, tuning-based auditing, and classification-based auditing.
This framework is extendable, currently supporting auditing for classifiers, generators, and statistical plots.
We evaluate it on three text classification tasks, two text summarization tasks, and two data visualization tasks across three scenarios.
The evaluation demonstrates the effectiveness of all proposed auditing methods across all these tasks.
We hope our research will promote the ethical and responsible use of synthetic data.

\section*{Acknowledgments}

We thank all anonymous reviewers for their constructive comments.
This work is partially funded by the European Health and Digital Executive Agency (HADEA) within the project ``Understanding the individual host response against Hepatitis D Virus to develop a personalized approach for the management of hepatitis D'' (DSolve, grant agreement number 101057917) and the BMBF with the project ``Repräsentative, synthetische Gesundheitsdaten mit starken Privatsphärengarantien'' (PriSyn, 16KISAO29K).

\section*{Ethics Considerations}

The datasets used in our evaluation are either publicly available or generated by LLMs, so there is a
minimal or non-existent presence of personally identifiable information (PII).
Hence, there is no risk of user de-anonymization, and our work does not fall under the category of human subjects research according to our Institutional Review Boards (IRB).
The proposed framework is to audit digital artifacts, identify those that have been trained on or derived from synthetic data, and enhance user awareness.
This effort aims to mitigate unforeseen consequences and risks in downstream applications, such as data hallucinations and inherent biases.
Through our research, we aspire to improve model transparency and support regulatory compliance, promoting the ethical and responsible use of synthetic data while building trust among users and stakeholders.

\section*{Open Science}

We hope our research will enhance model transparency and regulatory compliance, ensuring ethical and responsible use of synthetic data and fostering trust among users and stakeholders.
Hence, we open-source our datasets and code to facilitate further research.

\begin{small}
\bibliographystyle{plain}
\bibliography{normal_generated_py3}
\end{small}

\appendix
\label{section:appendix}
\renewcommand{\thesubsection}{\thesection.\arabic{subsection}}

\section{Metric-Based Auditing Using Entropy}
\label{appendix:equation_for_entropy}

Formally, we define the metric-based auditing using Entropy as the performance metric as follows:
\begin{equation}
\begin{aligned}
   &\mathcal{I}_{entr}( \mathcal{C}_{\textit{target}}, \mathcal{Q}_{\textit{syn}}) =  &\\ 
    &\mathds{1}\{-\frac{1}{n}\sum_{i=1}^{n}\sum_{j=0}^{c} \mathcal{C}_{\textit{target}}(x_i)_{j}log(\mathcal{C}_{\textit{target}}(x_i)_{j}) < \tau \},
\end{aligned}
\end{equation}
where $c$ is the number of the target class.

\begin{equation}
\begin{aligned}
   &\mathcal{I}_{entr}( \mathcal{C}_{\textit{target}}, \mathcal{Q}_{\textit{real}}) =  &\\ 
    &\mathds{1}\{-\frac{1}{n}\sum_{i=1}^{n}\sum_{j=0}^{c} \mathcal{C}_{\textit{target}}(x_i)_{j}log(\mathcal{C}_{\textit{target}}(x_i)_{j}) > \tau \},
\end{aligned}
\end{equation}

\section{Details of Evaluation Setup}
\label{appendix:eval_setup_sup}

\subsection{Details of Tasks}
\label{appendix:eval_setup_task_sup}

We consider three text classification tasks, two text summarization tasks, and two data visualization tasks on these five representative datasets.
The details of text classification tasks are as follows:
\begin{itemize}
\item \tone.
We consider a classifier that classifies the sentiment of movie reviews from the IMDB website~\cite{IMDB} into \textit{positive} or \textit{negative}.
The IMDB dataset~\cite{MDPHNP11} contains 25,000 (review, sentiment label) pairs for training and 25,000 for testing.
\item \ttwo.
We consider a classifier that categories the AG's news dataset (abbreviated as AG)~\cite{ZZL15} into four topics: \textit{World}, \textit{Sports}, \textit{Business}, and \textit{Sci/Tech}.
AG is collected from over 2,000 news sources, containing 120,000 training samples and 7,600 testing samples.
\item \tthree.
We consider a classifier that identifies emails from the Enron-Spam dataset (abbreviated as ES)~\cite{MAP06} as \textit{ham} (legitimate) or \textit{spam}.
ES contains around 31,000 training samples and 2,000 testing samples.
\end{itemize}

\noindent The details of text summarization tasks are as follows:
\begin{itemize}
\item \tfour.
The CNN/DM dataset contains over 312,000 unique news articles, including 287,113 training instances, 13,368 validation instances, and 11,490 testing instances, as written by journalists at CNN and the Daily Mail.
Each sample includes an article with its corresponding highlights written by the article's author.
\item \tfive.
The XSum dataset, collected from online articles from the British Broadcasting Corporation (BBC), includes 204,045 training instances, 11,132 validation instances, and 11,334 testing instances.
Each instance includes a news article with its corresponding one-sentence summary.
\end{itemize}

\noindent The details of data visualization tasks are as follows:
\begin{itemize}
\item \tsix.
This task uses the same dataset, i.e., IMDB, along with its data split and synthetic data generation settings, as \tone.
The synthetic data is generated through a zero-shot prompt method.
We provide movie titles, outlines, and target labels to an LLM to generate reviews.
\item \tseven.
This task uses the same dataset, i.e., AG, along with its data split and synthetic data generation settings, as \ttwo.
The synthetic data is generated using a paraphrasing prompt strategy.
We provide the original article and the target label to the LLM and instruct it to rewrite a new article.
\end{itemize}

\subsection{Details of Data Split}
\label{appendix:eval_setup_data_split_sup}

Below are details of data splitting for each task.
\begin{itemize}
\item \tone (zero-shot).
We use the IMDB training set as the target real dataset \dtr.
We leave out 1,000 samples in the IMDB testing set as \dtest and use the rest as the reference real dataset \drr.
We randomly sample instances from \dtest to construct \qr.
We also leave out 1,000 samples as \qa from the retrieved movies as additional information to constructing \qs in a zero-shot prompt strategy.
The rest of the retrieved movies are evenly split into \dta and \dta.
\item \ttwo, \tthree, \tfour, and \tfive (paraphrasing).
We randomly split the AG/Enron-Spam/CNNDM/XSum training set into two evenly disjoint subsets \dr and \dt.
\dr and \dt are further divided evenly into \drr, \dra, \dtr, and \dta.
We randomly sample 1000 samples from the AG/Enron-Spam/CNNDM/XSum testing set to serve as \dtest.
We then randomly sample instances from \dtest to construct \qr and use them as reference samples \qa for constructing \qs in a paraphrasing prompt strategy.
\end{itemize}

\subsection{Details of Synthetic Data Generation}
\label{appendix:eval_setup_syn_data_sup}

Below are details of synthetic data generation for each task, and all prompts used in our evaluation are shown in~\refappendix{appendix:prompt}.
\begin{itemize}
\item \tone (zero-shot).
These auxiliary sets \dta, \dra, and \qa consist of (movie title, outline, sentiment label) pairs.
We instruct the LLM to use the reference prompt to generate a positive review and a negative review for each pair in \dra and \qa and use the target prompt to generate a positive review and a negative review for each pair in \dta.
The final synthetic sets \drs, \qs, and \dts consist of (generated review, label) pairs.
We set the temperature to 0.5 to balance the diversity and usability.
\item \ttwo and \tthree (paraphrasing).
These auxiliary sets \dta, \dra, and \qa consist of (input text, label) pairs.
For each pair, we include each original input with its target label in the prompt and ask LLMs to paraphrase it into a new synthetic sample.
We use the target prompt for constructing \dts and the reference prompt for constructing \drs and \qs.
The final synthetic sets \drs, \qs, and \dts consist of (synthetic input, label) pairs.
To ensure that the synthetic examples generated by the paraphrasing strategy exhibit substantial differences from the original examples, we set the temperature to 1.
\item \tfour and \tfive (paraphrasing).
These auxiliary sets \dta, \dra, and \qa consist of (article, summary) pairs.
For each original pair, we include both the article and summary in the prompt and ask LLMs to paraphrase the original summary into a new synthetic sample.
We use the target prompt for constructing \dts and the reference prompt for constructing \drs and \qs.
The final synthetic sets \drs, \qs, and \dts consist of (article, synthetic summary) pairs.
To ensure that the synthetic examples generated by the paraphrasing strategy exhibit substantial differences from the original examples, we set the temperature to 1.
\end{itemize}
Note that, before constructing the final synthetic dataset for each task, we perform a filtering process that filters out the refusal outputs.

\subsection{Prompts for Synthetic Data Generation}
\label{appendix:prompt}

We show the prompts of synthetic data generation for all tasks in~\autoref{table:prompt_generation_sentiment_analysis},~\autoref{table:prompt_generation_topic_classification},~\autoref{table:prompt_generation_spam_detection},~\autoref{table:prompt_generation_text_summary_cnn}, and~\autoref{table:prompt_generation_text_summary_xsum}.

\subsection{Details of Training Classifiers}
\label{appendix:eval_setup_training_classifier_sup}

\mypara{Training Dataset Size For Each Classifier}
We list the size of the training dataset for each classifier as follows:
\begin{itemize}
\item \tone: We randomly sample 1,500 reviews per class (3,000 reviews in total).
\item \ttwo: We randomly sample 2,000 news articles per class (8,000 articles in total).
\item \tthree: We randomly sample 2,000 emails per class (4,000 emails in total).
\end{itemize}

\mypara{Target/Reference Classifier Set}
We construct the target classifier set, which includes 50 target real classifiers \deltactrs and 50 target synthetic classifiers \deltactss, to evaluate the proposed auditing methods.
Meanwhile, we construct the reference classifier set, comprising 50 reference real classifiers \deltacrers and 50 reference synthetic classifiers \deltacress, to train the meta-classifier and empirically obtain the threshold values 
We leverage the following setup for each task with each LLM:
\begin{itemize}
\item \textit{Real classifiers:} We run the training procedure 100 times, each with a different seed, to train  \deltactrs on \dtr and \deltacrers on \drr.
Each set contains 50 classifiers.
\item \sone: We run the training procedure 100 times, each with a different seed, to train \deltactss on \dts and \deltacress on \drs.
Each set contains 50 classifiers.
\item \stwo: We run the training procedure 100 times, each with a different seed, to train \deltactss on \dts and \dtr, and \deltacress on \drs and \drr.
We train five classifiers for each of the ten synthetic proportions.
\item \sthree: We run the training procedure 100 times, each with a different seed, to train \deltactss on \dts and \dtr, and \deltacress on \drs and \drr, with random synthetic proportions generated by different seeds.
Each set contains 50 classifiers.
\end{itemize}
Note that we include all four LLMs in \sone and \stwo, so there are a total of 400 synthetic classifiers in \sone, 400 synthetic classifiers in \stwo, and 100 synthetic classifiers in \sthree for each task.

\subsection{Details of Training Generators}
\label{appendix:eval_setup_training_generator_sup}

\mypara{Training Dataset Size For Each Generator}
For \tfour and \tfive, we randomly sample 5,000 articles as the training dataset for each generator.

\mypara{Target/Reference Generator Set}
We construct the target generator set, which includes 50 target real generators \deltagtrs and 50 target synthetic generators \deltagtss, to evaluate the proposed auditing methods.
Meanwhile, we construct the reference generator set, comprising 50 reference real generators \deltagrers and 50 reference synthetic generators \deltagress, to train the meta-classifier and empirically obtain the threshold values.
We leverage the following setup for each task with each LLM:
\begin{itemize}
\item \textit{Real generators:} We run the training procedure 100 times, each with a different seed, to train  \deltagtrs on \dtr and \deltagrers on \drr.
Each set contains 50 generators.
\item \sone: We run the training procedure 100 times, each with a different seed, to train \deltagtss on \dts and \deltagress on \drs.
Each set contains 50 generators.
\item \stwo: We run the training procedure 100 times, each with a different seed, to train \deltagtss on \dts and \dtr, and \deltagress on \drs and \drr.
We train five generators for each of the ten synthetic proportions.
\item \sthree: We run the training procedure 100 times, each with a different seed, to train \deltagtss on \dts and \dtr, and \deltagress on \drs and \drr, with random synthetic proportions generated by different seeds.
Each set contains 50 generators.
\end{itemize}
Note that we include all four LLMs in \sone and \stwo, so there are a total of 400 synthetic generators in \sone, 400 synthetic generators in \stwo, and 100 synthetic generators in \sthree for each task.

\subsection{Details of Generating Plots}
\label{appendix:eval_setup_training_plot_sup}

\mypara{Input Dataset Size For Each Plot}
We randomly sample 1,000 samples from the dataset to generate t-SNE plots, and we follow the same three scenarios in~\autoref{section:classifier_eval_setup} to control the synthetic proportion in the input data.

\mypara{Target/Reference Plot Set}
We construct the target plot set, which includes 200 target real plots \deltaptrs and 200 target synthetic plots \deltaptss, to evaluate the proposed auditing methods.
Meanwhile, we construct the reference plot set, comprising 200 reference real plots \deltaprers and 200 reference synthetic plots \deltapress, to train the meta-classifier and empirically obtain the threshold values.
We leverage the following setup for each task with each LLM:
\begin{itemize}
\item \textit{Real plots:} We run the plotting procedure 400 times, each with a different seed, to plot  \deltaptrs on \dtr and \deltaprers on \drr.
Each set contains 200 plots.
\item \sone: We run the plotting procedure 400 times, each with a different seed, to plot \deltaptss on \dts and \deltapress on \drs.
Each set contains 200 plots.
\item \stwo: We run the plotting procedure 400 times, each with a different seed, to plot \deltaptss on \dts and \dtr, and \deltapress on \drs and \drr.
We generate 20 plots for each of the ten synthetic proportions.
\item \sthree: We run the plotting procedure 400 times, each with a different seed, to plot \deltaptss on \dts and \dtr, and \deltapress on \drs and \drr, with random synthetic proportions generated by different seeds.
Each set contains 200 plots.
\end{itemize}
Overall, for each task, we include all four LLMs in \sone and \stwo, resulting in a total of 400 real plots, 1600 synthetic plots in \sone, 1600 synthetic plots in \stwo, and 400 synthetic plots in \sthree.

\section{Additional Results of Auditing Classifiers}
\label{appendix:audit_classifier_sup}

\subsection{Target Performance of Classifiers}
\label{appendix:target_classifier_performance}

\begin{figure*}[!t]
\centering
\begin{subfigure}{0.65\columnwidth}
\includegraphics[width=\columnwidth]{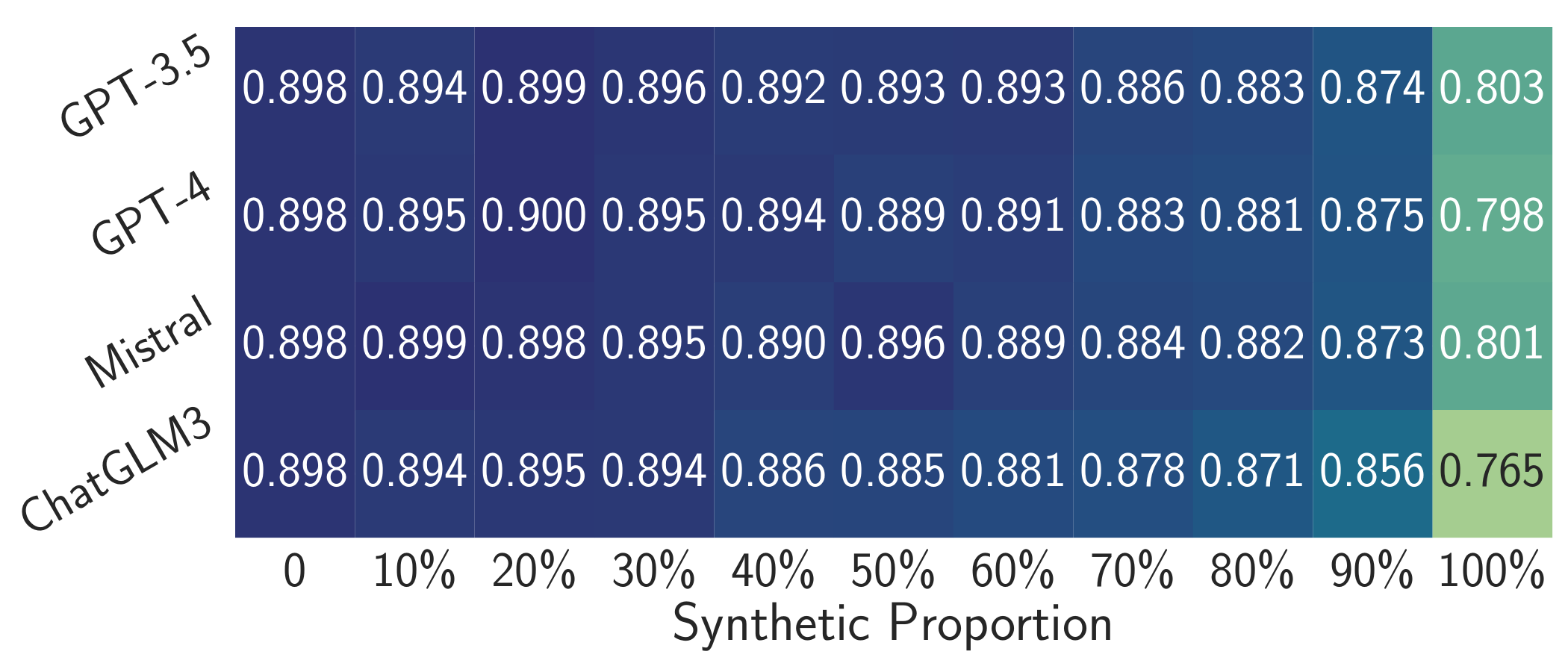}
\caption{\tone}
\end{subfigure}
\begin{subfigure}{0.65\columnwidth}
\includegraphics[width=\columnwidth]{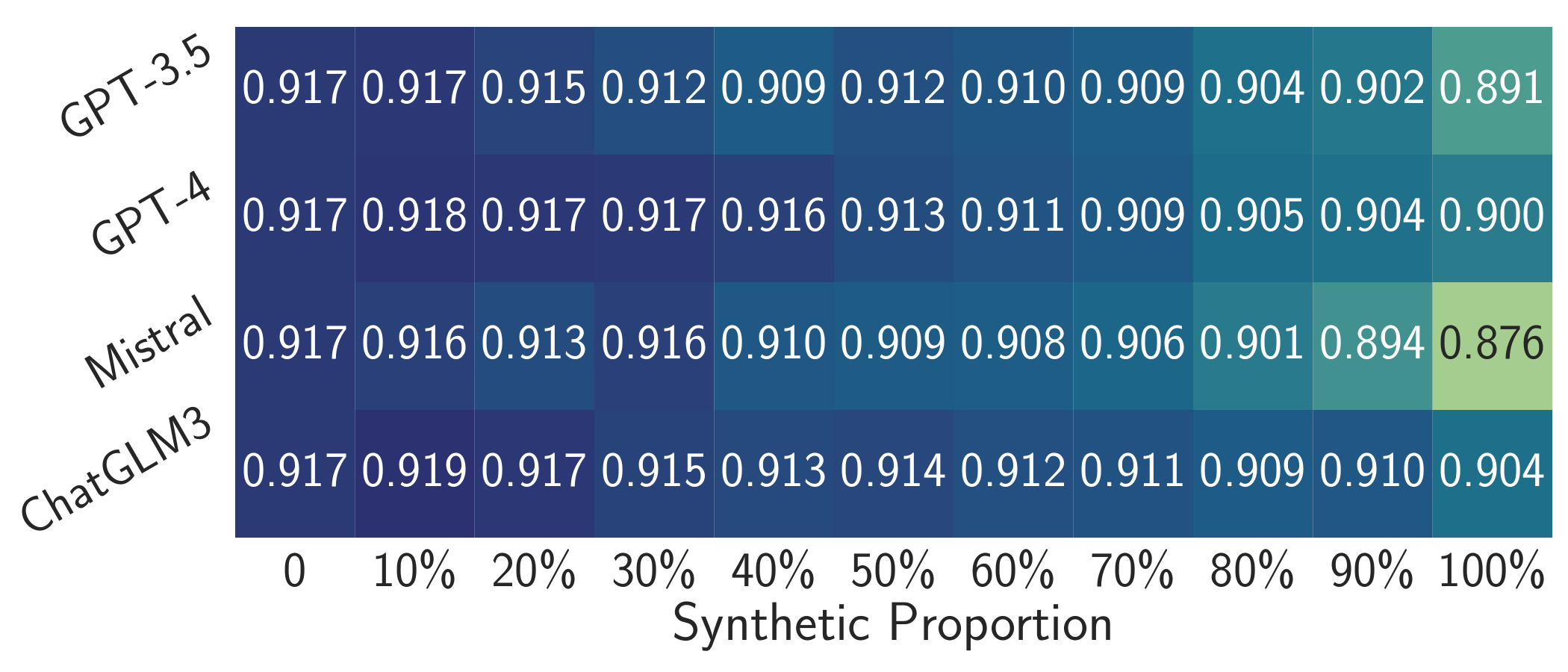}
\caption{\ttwo}
\end{subfigure}
\begin{subfigure}{0.65\columnwidth}
\includegraphics[width=\columnwidth]{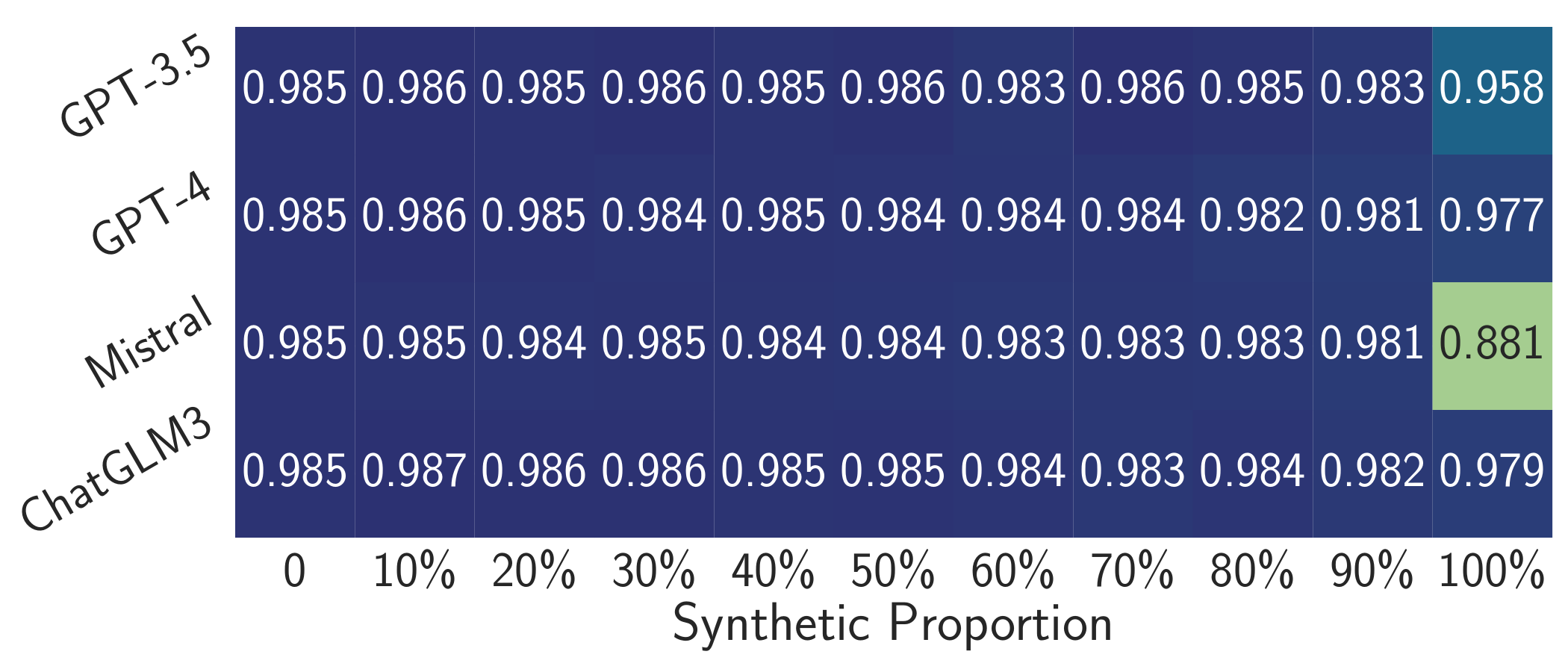}
\caption{\tthree}
\end{subfigure}
\caption{Average target performance of real and synthetic classifiers (\deltactrs and \deltactss) fine-tuned on pre-trained DistilBERT for (a) \tone, (b) \ttwo, and (c) \tthree evaluated on \dtest.}
\label{figure:target_performance_distillbert_sup}
\end{figure*}

\begin{figure*}[!t]
\centering
\begin{subfigure}{0.65\columnwidth}
\includegraphics[width=\columnwidth]{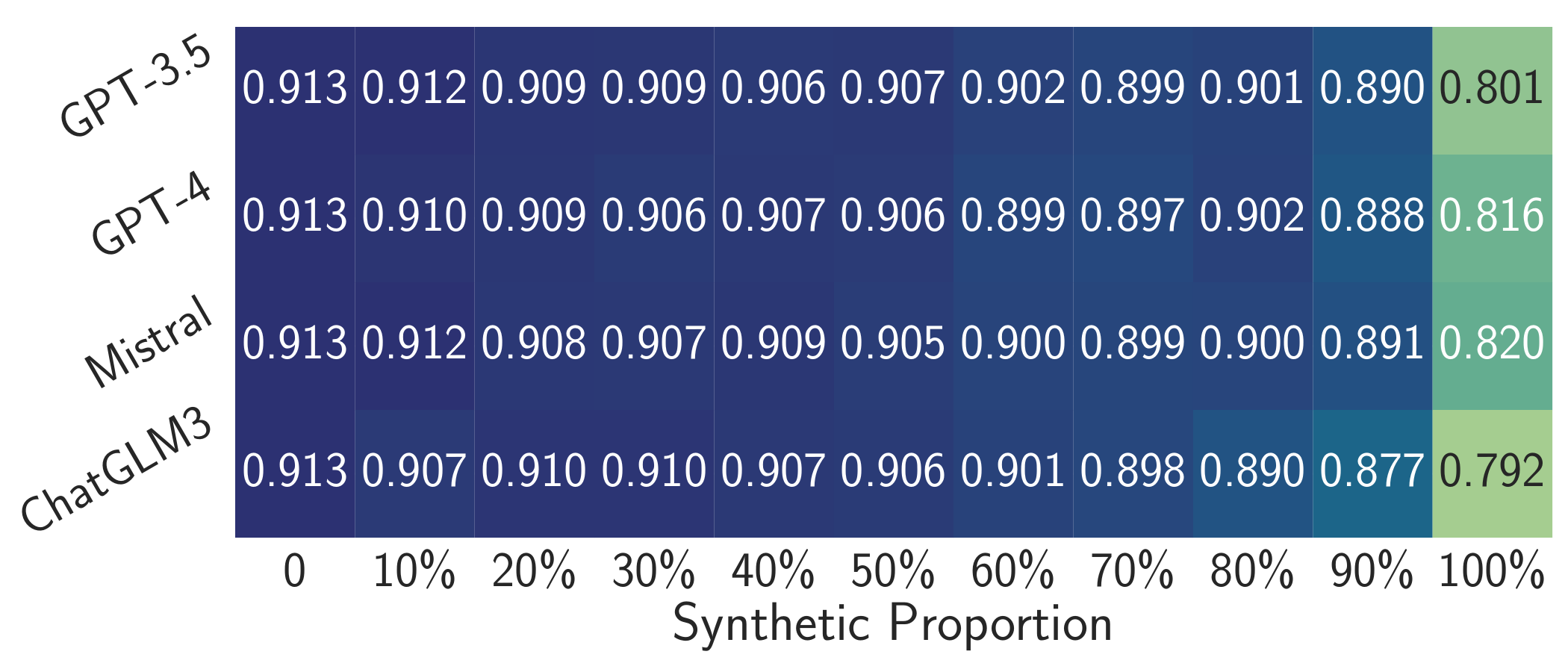}
\caption{\tone}
\end{subfigure}
\begin{subfigure}{0.65\columnwidth}
\includegraphics[width=\columnwidth]{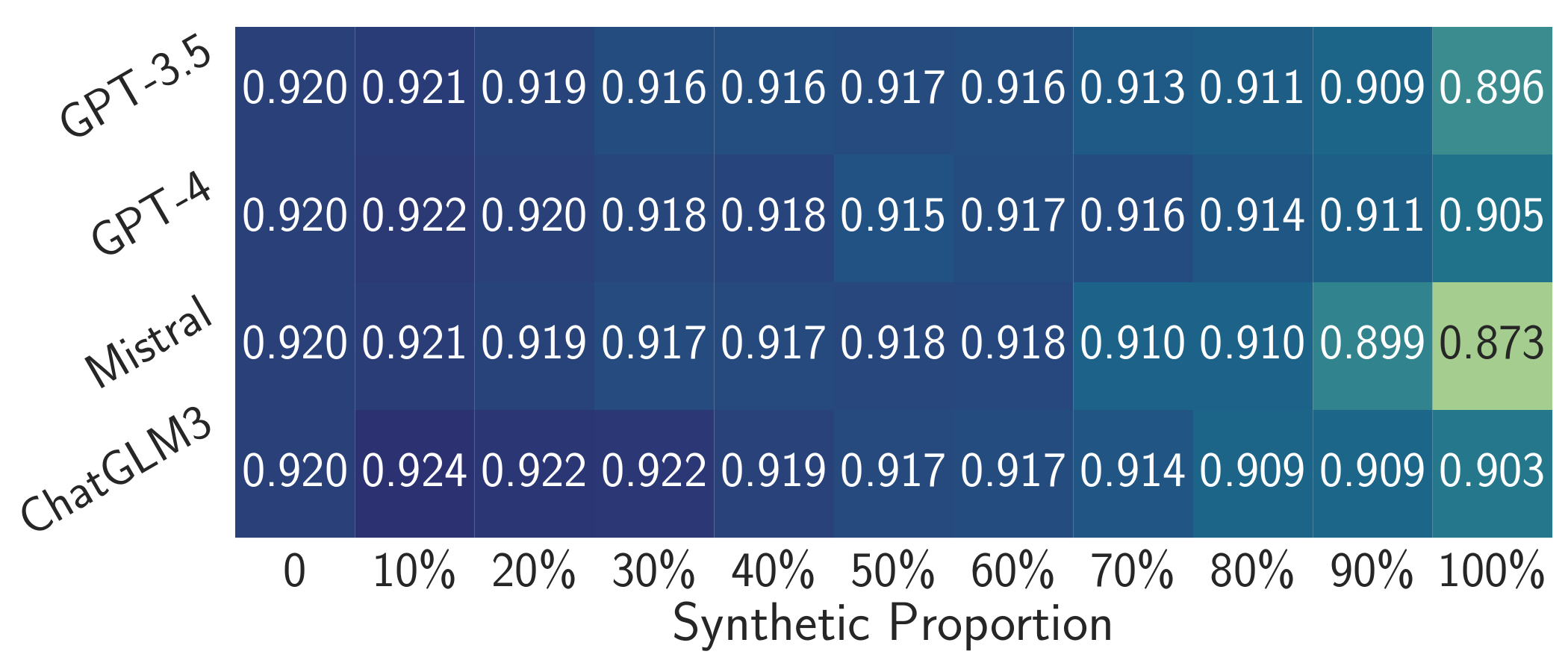}
\caption{\ttwo}
\end{subfigure}
\begin{subfigure}{0.65\columnwidth}
\includegraphics[width=\columnwidth]{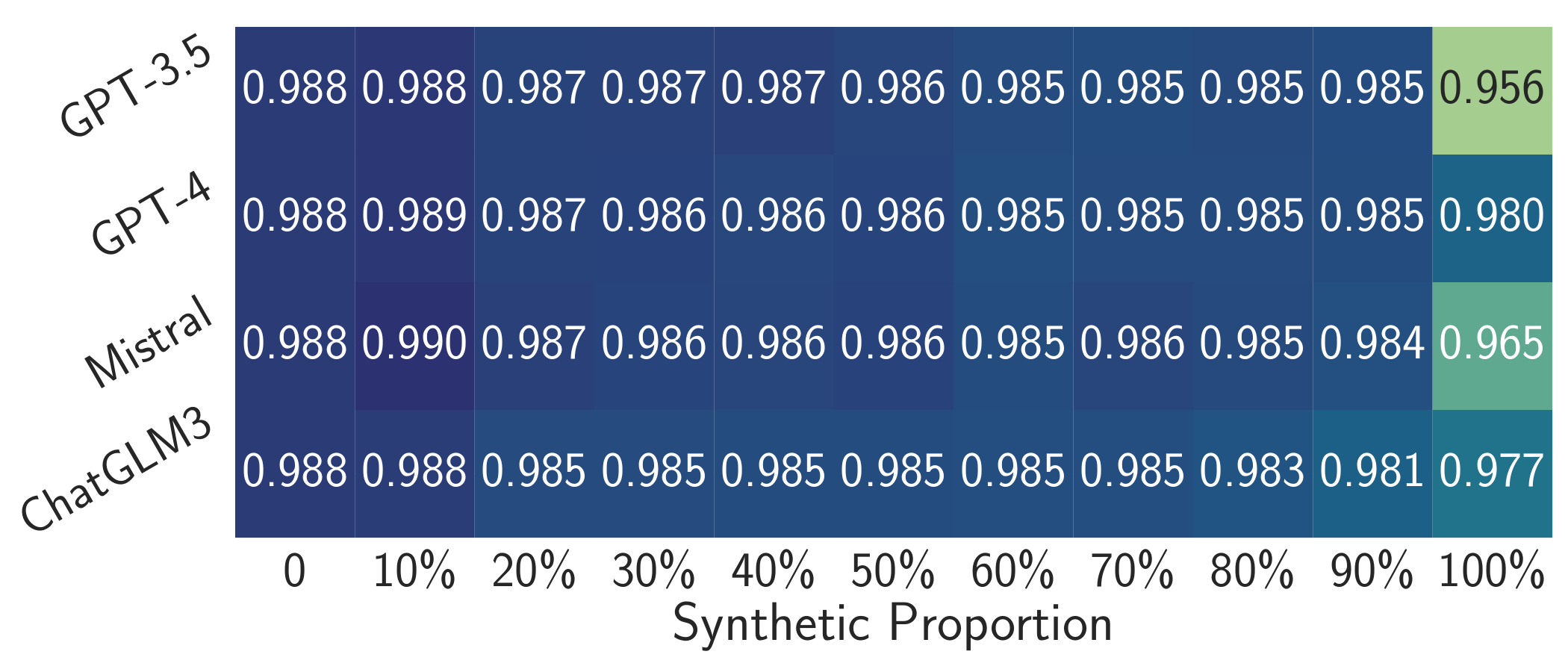}
\caption{\tthree}
\end{subfigure}
\caption{Average target performance of real and synthetic classifiers (\deltactrs and \deltactss) fine-tuned on pre-trained BERT model for (a) \tone, (b) \ttwo, and (c) \tthree evaluated on \dtest.}
\label{figure:target_performance_bert}
\end{figure*}

We present the average target performance of all real and synthetic classifiers fine-tuned on pre-trained DistilBERT in~\autoref{figure:target_performance_distillbert_sup} and~\autoref{figure:target_performance_bert}.
The target performance is evaluated on \dtest, which consists of only real data and is measured by classification accuracy.
We observe that synthetic classifiers, particularly those leveraging a portion of real data (\stwo), achieve performance comparable to real classifiers.
The strong performance, combined with the benefits of convenience and lower cost in generating synthetic data, exemplifies that synthetic classifiers may gain popularity and prominence in the market.
Additionally, the superior target performance of the target synthetic classifiers substantiates the robustness and reliability of our evaluation.

\subsection{Additional Results of Metric-Based Auditing Classifiers with \qs}
\label{appendix:metric_based_audit_classifier_qs_sup}

\begin{figure}[ht]
\centering
\begin{subfigure}{0.45\columnwidth}
\includegraphics[width=\columnwidth]{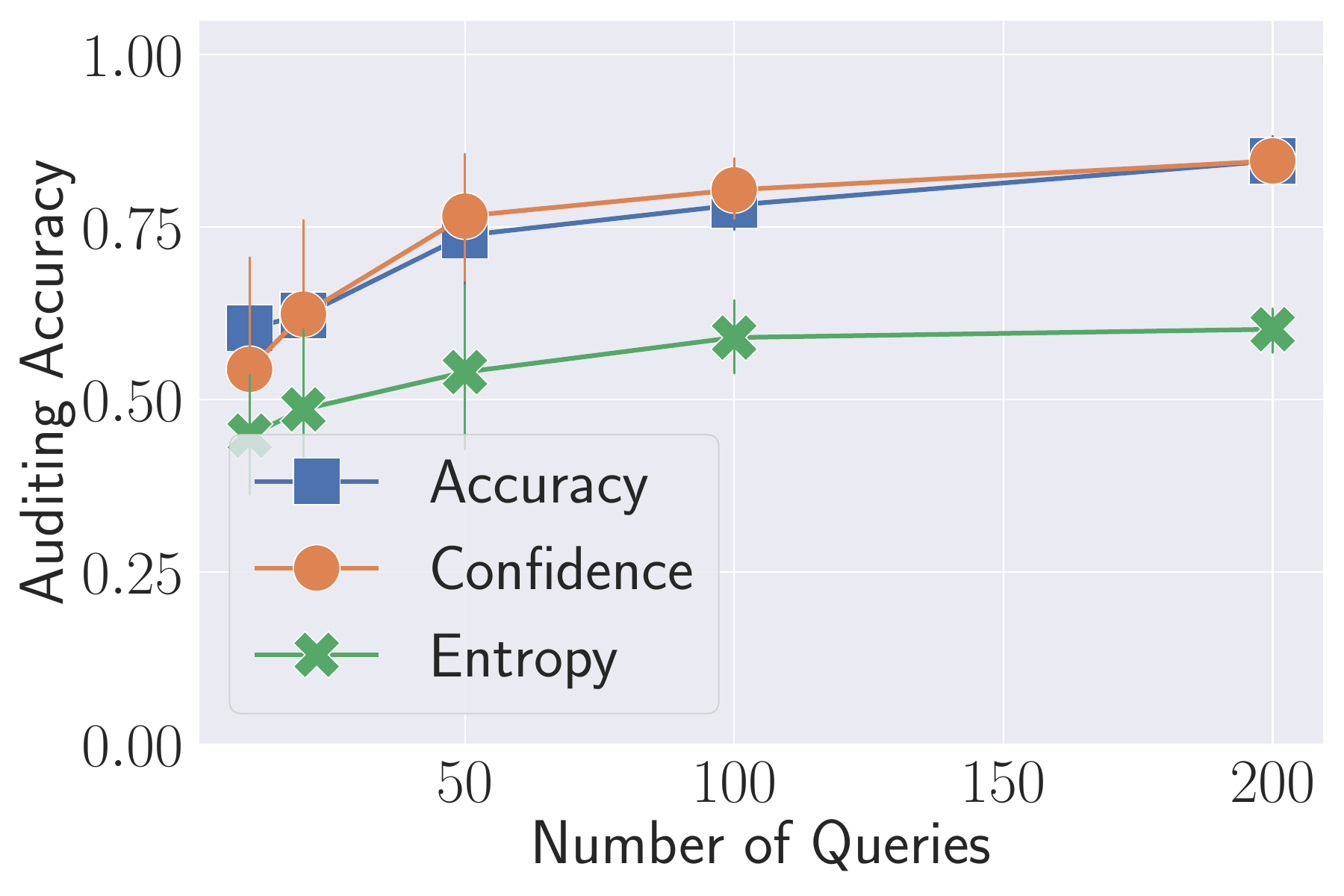}
\caption{\sone}
\end{subfigure}
\begin{subfigure}{0.45\columnwidth}
\includegraphics[width=\columnwidth]{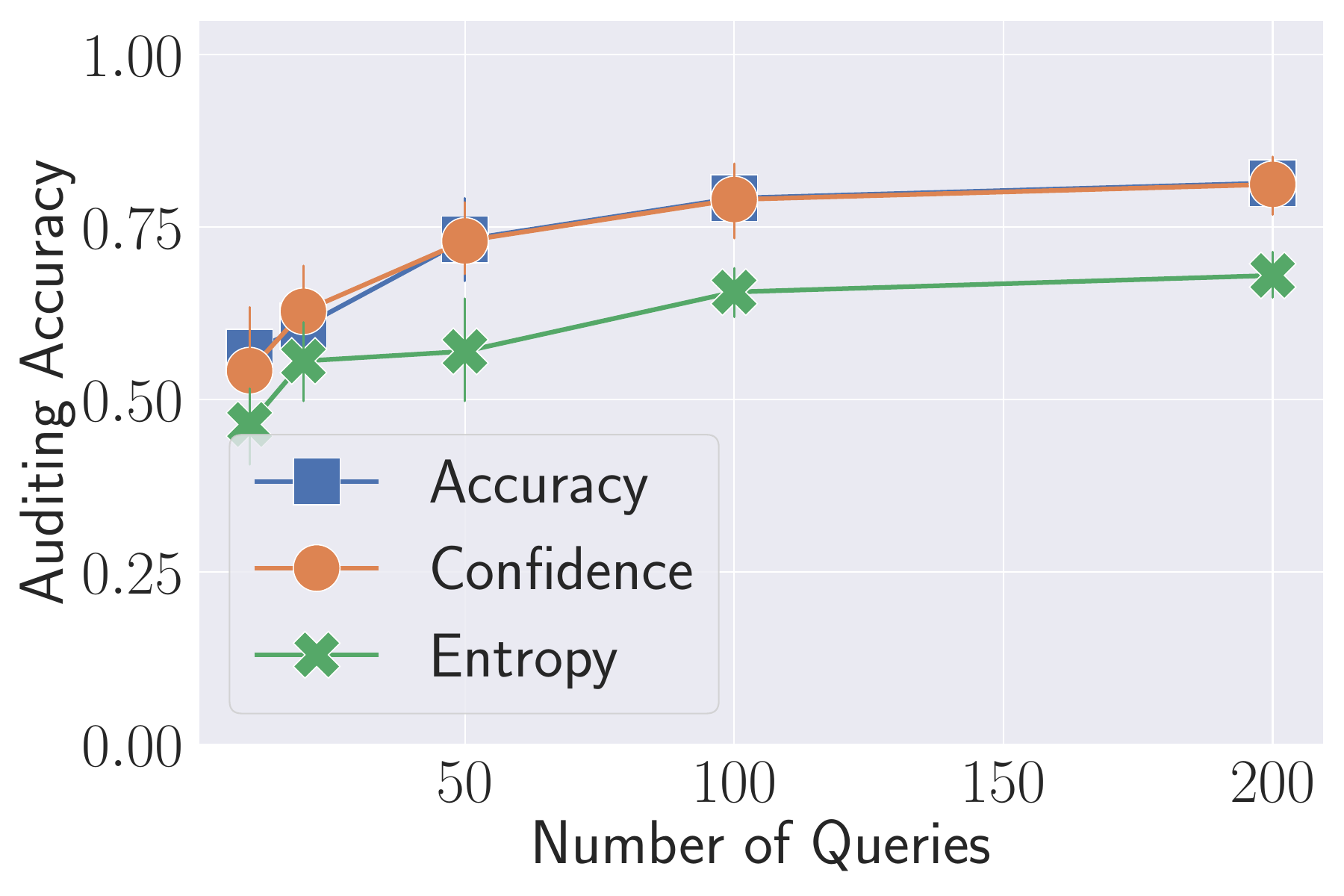}
\caption{\stwo}
\end{subfigure}
\caption{Metric-based auditing performance for target classifiers fine-tuned on DistilBERT with varying query budgets of \qs $\{10, 20, 50, 100, 200\}$ for \tthree in (a) \sone and (b) \stwo.
The source LLM is GPT-4.}
\label{figure:audit_classifier_x_query_syn_gpt4_sup}
\end{figure}

\begin{figure}[ht]
\centering
\begin{subfigure}{0.45\columnwidth}
\includegraphics[width=\columnwidth]{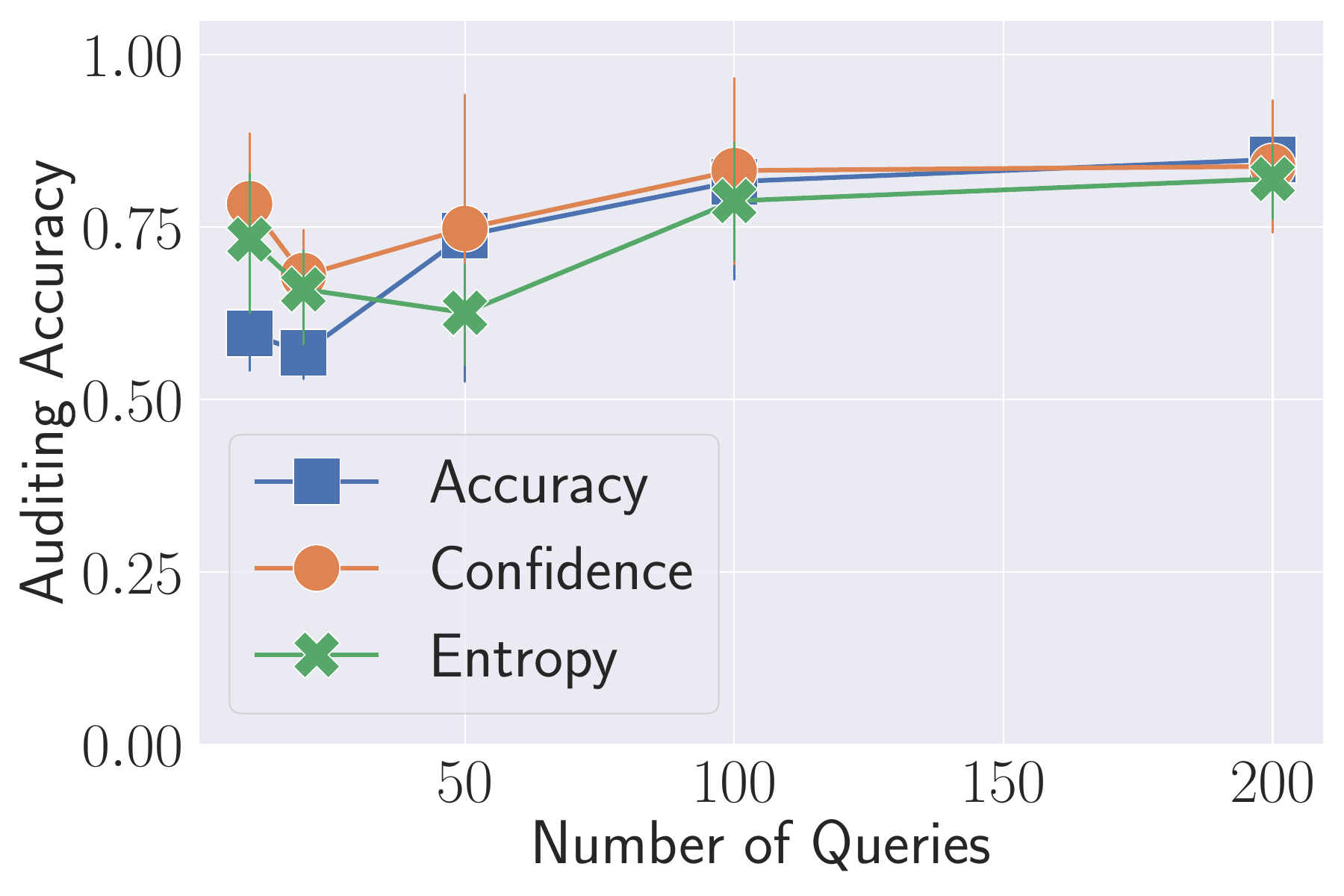}
\caption{\sone}
\end{subfigure}
\begin{subfigure}{0.45\columnwidth}
\includegraphics[width=\columnwidth]{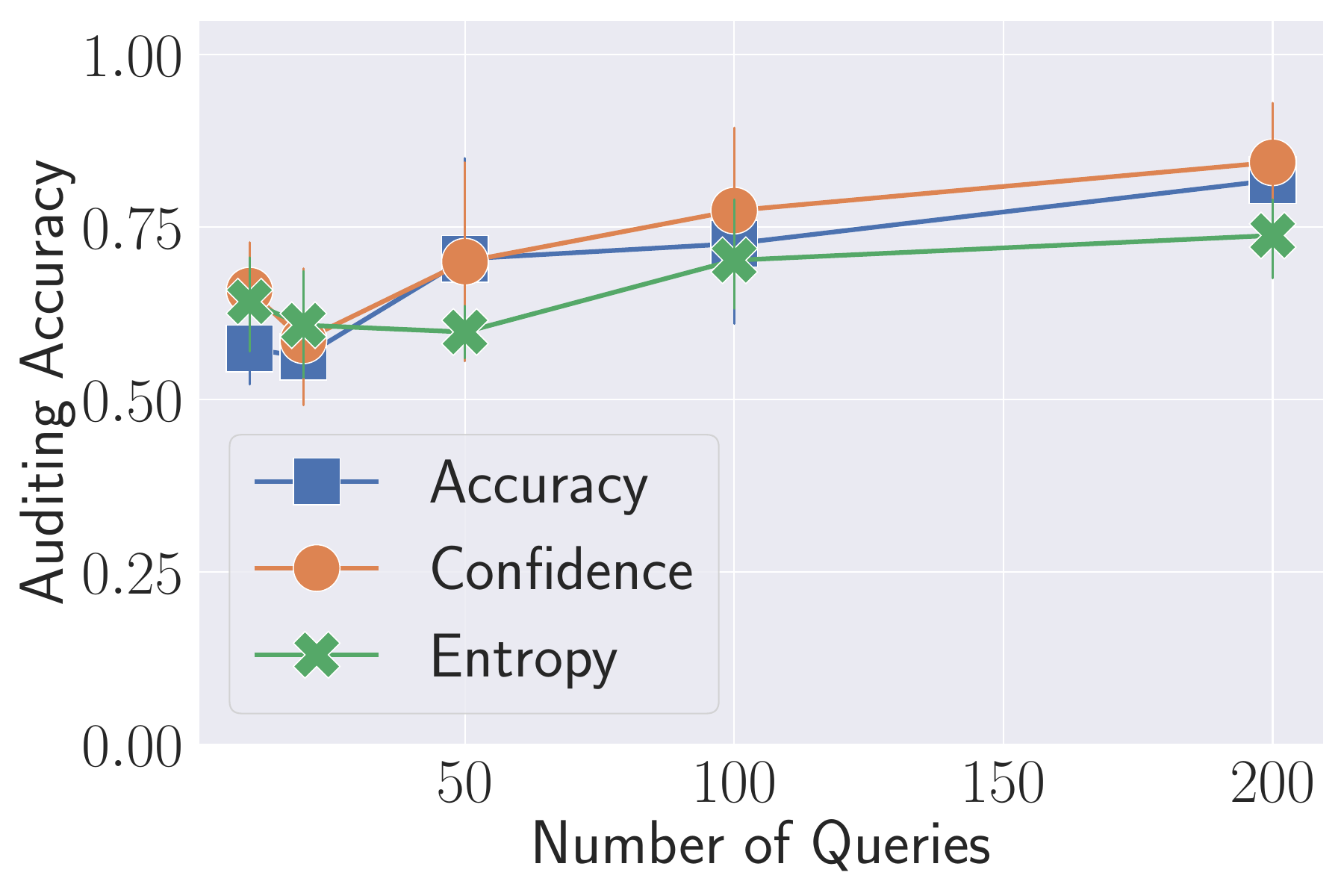}
\caption{\stwo}
\end{subfigure}
\caption{Metric-based auditing performance for target classifiers fine-tuned on DistilBERT with varying query budgets of \qs $\{10, 20, 50, 100, 200\}$ for \tthree in (a) \sone and (b) \stwo.
The source LLM is Mistral.}
\label{figure:audit_classifier_x_query_syn_mistral_sup}
\end{figure}

\begin{figure}[ht]
\centering
\begin{subfigure}{0.45\columnwidth}
\includegraphics[width=\columnwidth]{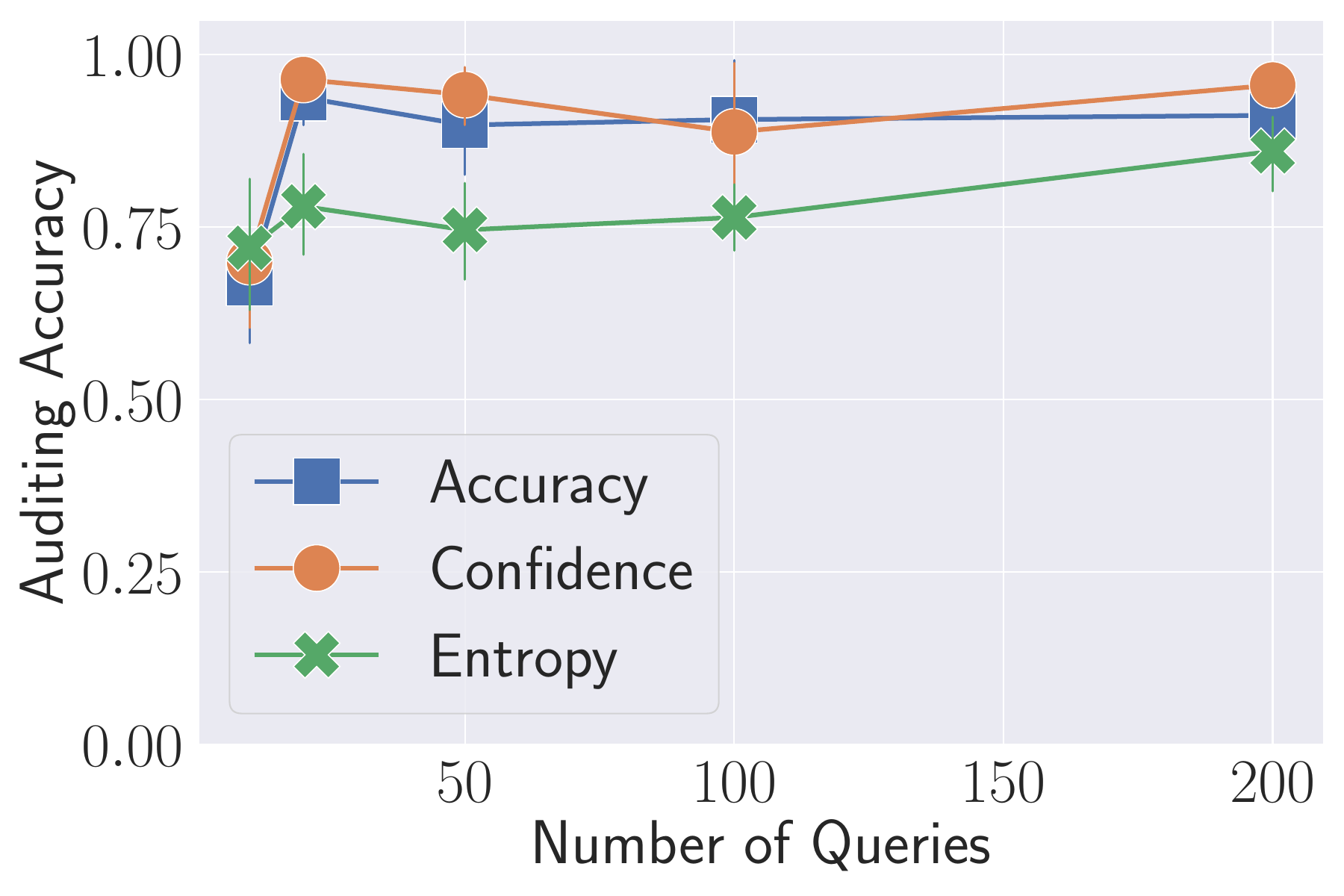}
\caption{\sone}
\end{subfigure}
\begin{subfigure}{0.45\columnwidth}
\includegraphics[width=\columnwidth]{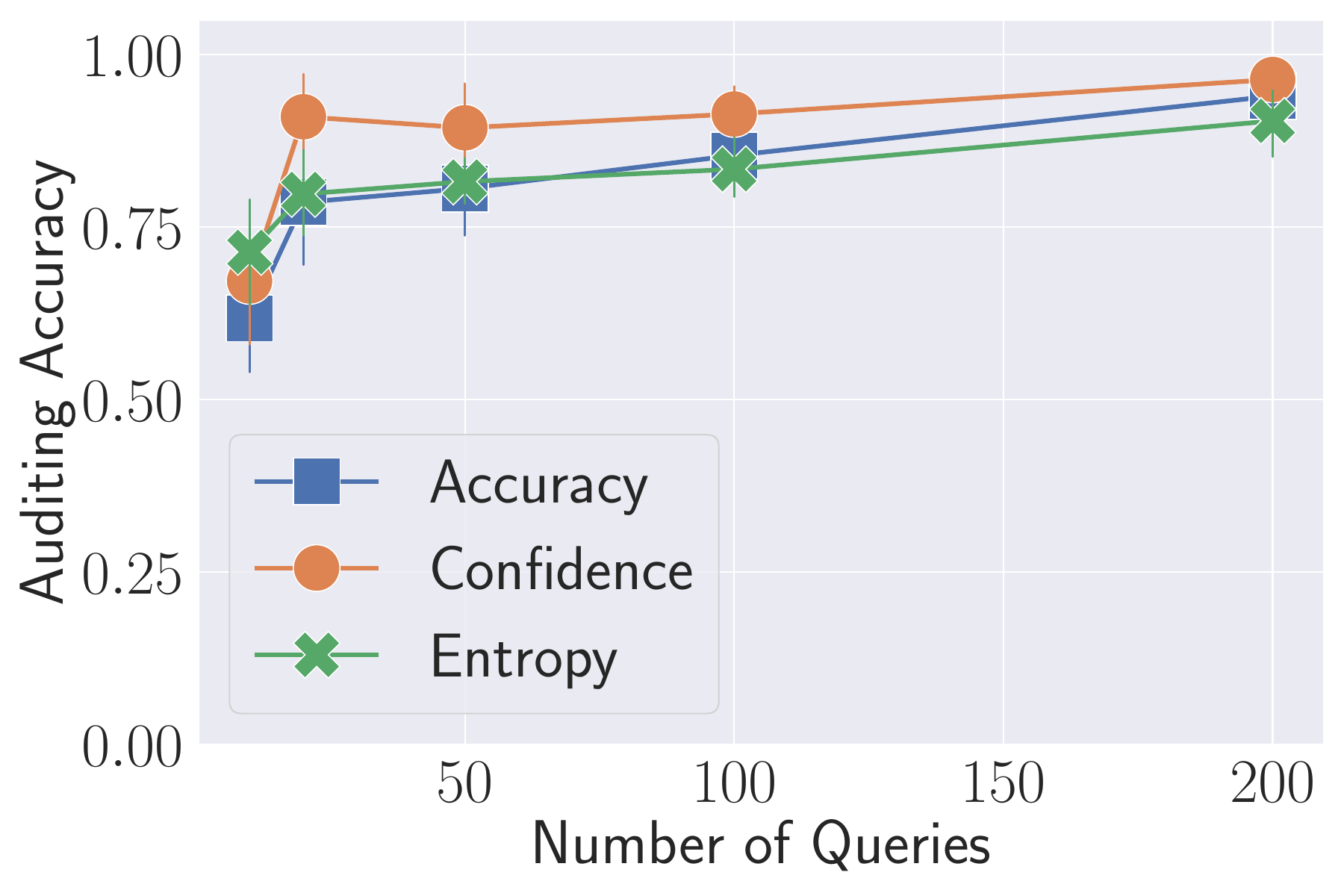}
\caption{\stwo}
\end{subfigure}
\caption{Metric-based auditing performance for target classifiers fine-tuned on DistilBERT with varying query budgets of \qs $\{10, 20, 50, 100, 200\}$ for \tthree in (a) \sone and (b) \stwo.
The source LLM is ChatGLM3.}
\label{figure:audit_classifier_x_query_syn_chatglm_sup}
\end{figure}

In~\autoref{figure:audit_classifier_x_query_syn_gpt4_sup},~\autoref{figure:audit_classifier_x_query_syn_mistral_sup}, and~\autoref{figure:audit_classifier_x_query_syn_chatglm_sup} we present the performance of the metric-based auditing with varying query budgets of \qs on other source LLMs.
We observe the same conclusion as in~\autoref{figure:audit_classifier_x_query_syn}: in both scenarios, more synthetic queries help determine a better threshold, resulting in improved auditing performance.

\begin{figure}[ht]
\centering
\begin{subfigure}{0.45\columnwidth}
\includegraphics[width=\columnwidth]{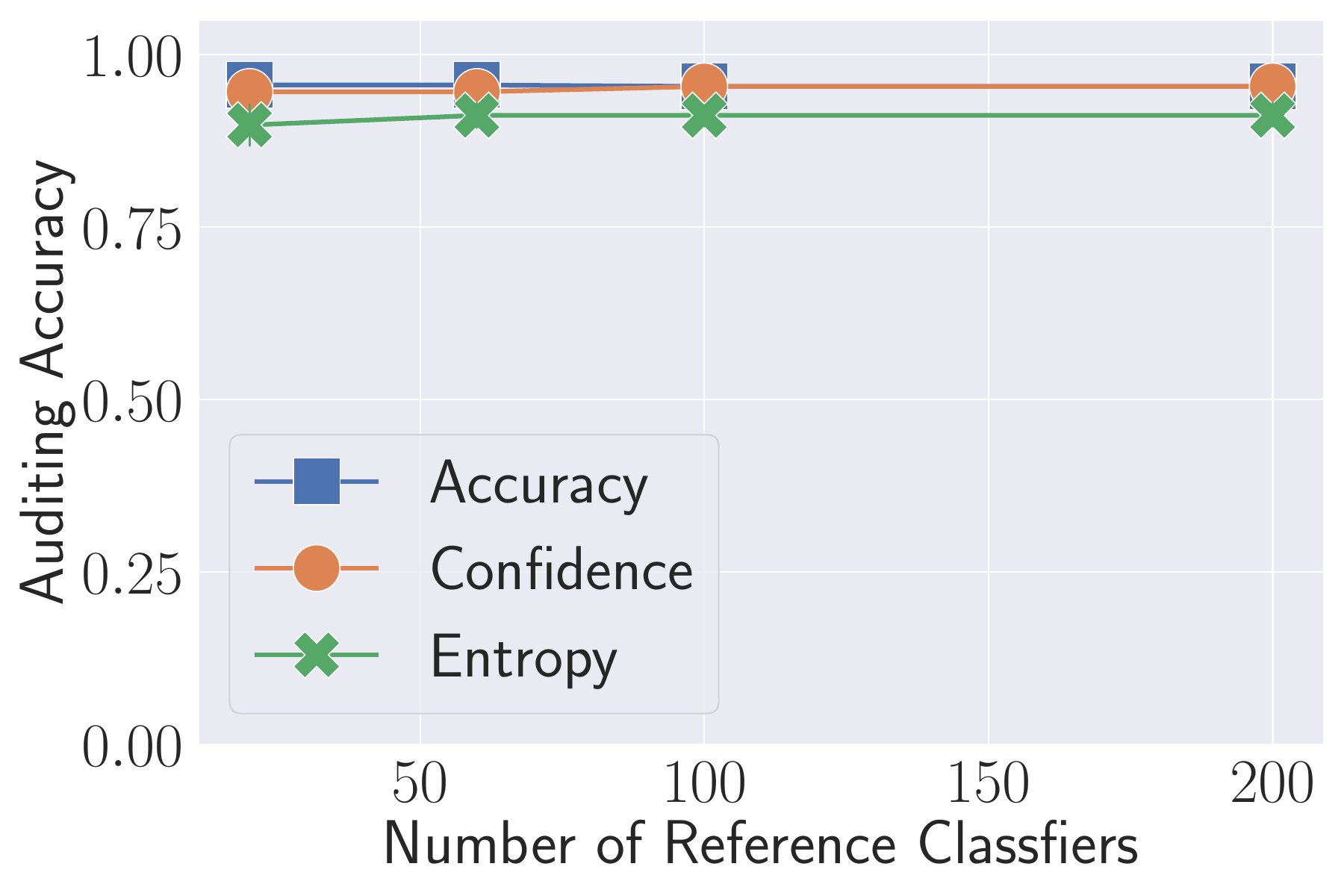}
\caption{\sone}
\end{subfigure}
\begin{subfigure}{0.45\columnwidth}
\includegraphics[width=\columnwidth]{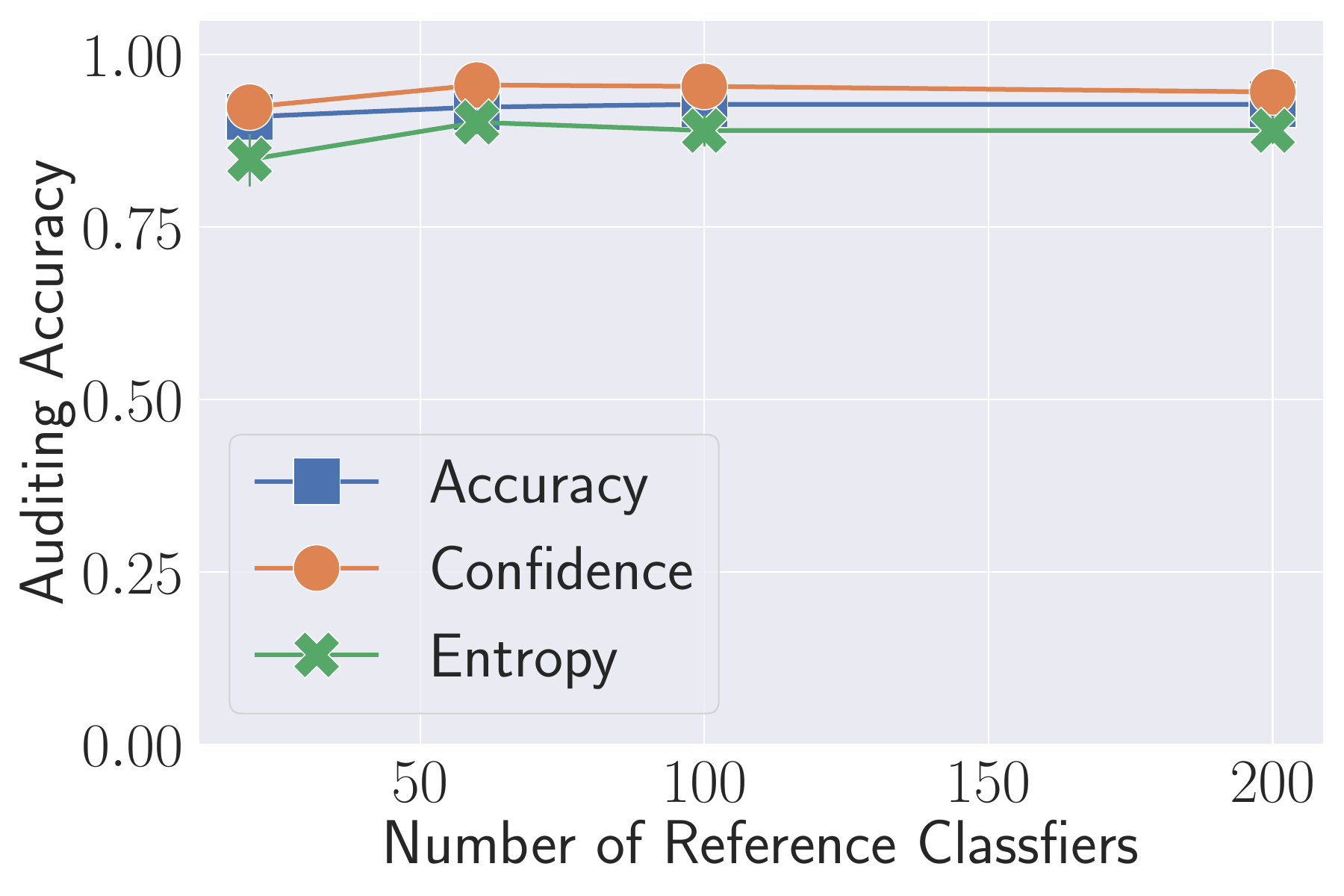}
\caption{\stwo}
\end{subfigure}
\caption{Metric-based auditing performance for target classifiers fine-tuned on DistilBERT with varying number of reference classifiers $\{20, 60, 100, 200\}$ in (a) \sone and (b) \stwo for \tthree.
The query budget of \qs is set to 200.
The source LLM is GPT-3.5.}
\label{figure:audit_classifier_x_ns_syn}
\end{figure}

Meanwhile, as illustrated in~\autoref{figure:audit_classifier_x_ns_syn}, we demonstrate that 20 reference classifiers are sufficient to find a good threshold to achieve decent auditing performance, and the benefit of more reference classifiers for metric-based auditing with \qs is minimal.

\subsection{Additional Results of Metric-Based Auditing Classifiers with \qr}
\label{appendix:metric_based_audit_classifier_qr_sup}

\begin{figure}[ht]
\centering
\begin{subfigure}{0.45\columnwidth}
\includegraphics[width=\columnwidth]{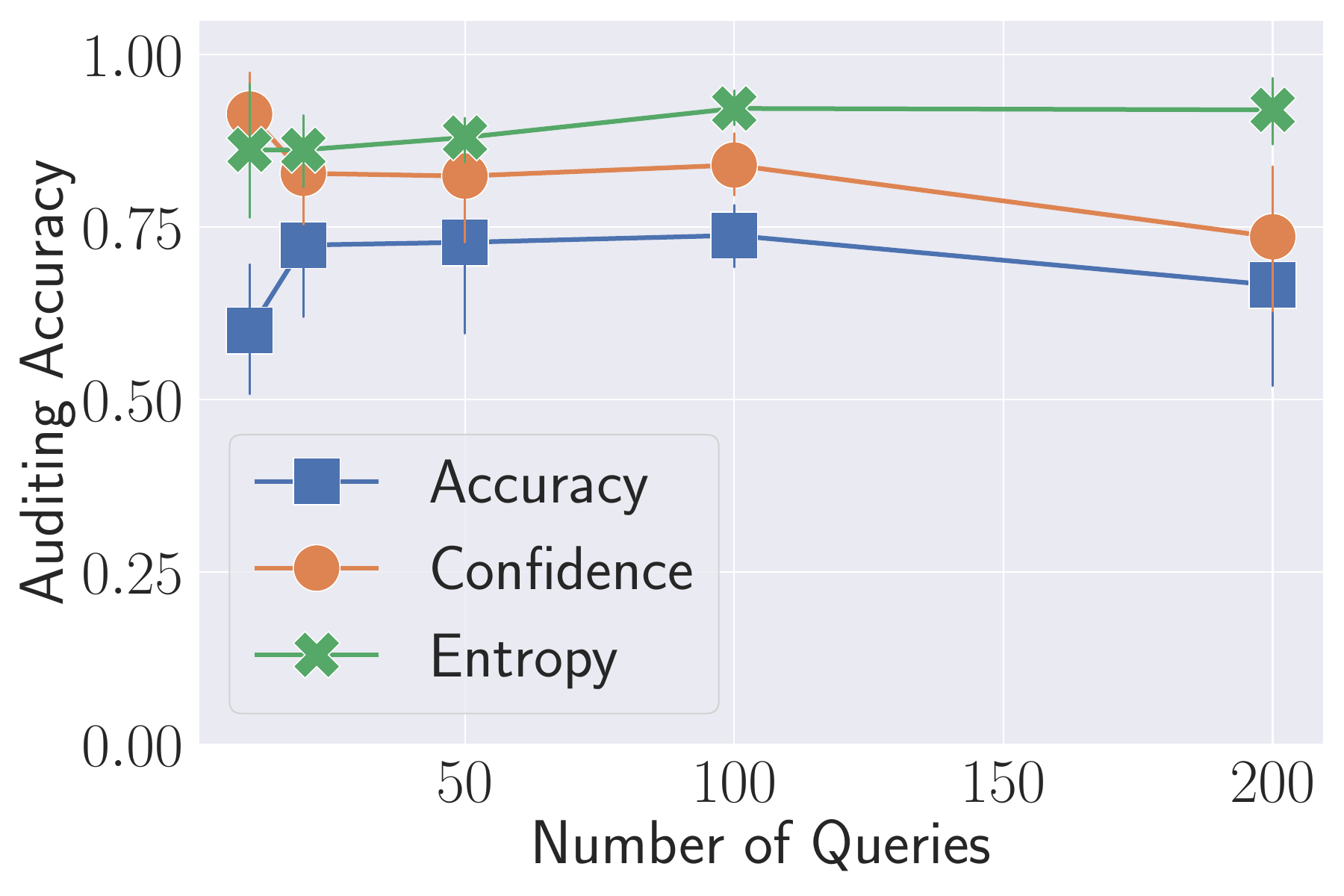}
\caption{\sone}
\end{subfigure}
\begin{subfigure}{0.45\columnwidth}
\includegraphics[width=\columnwidth]{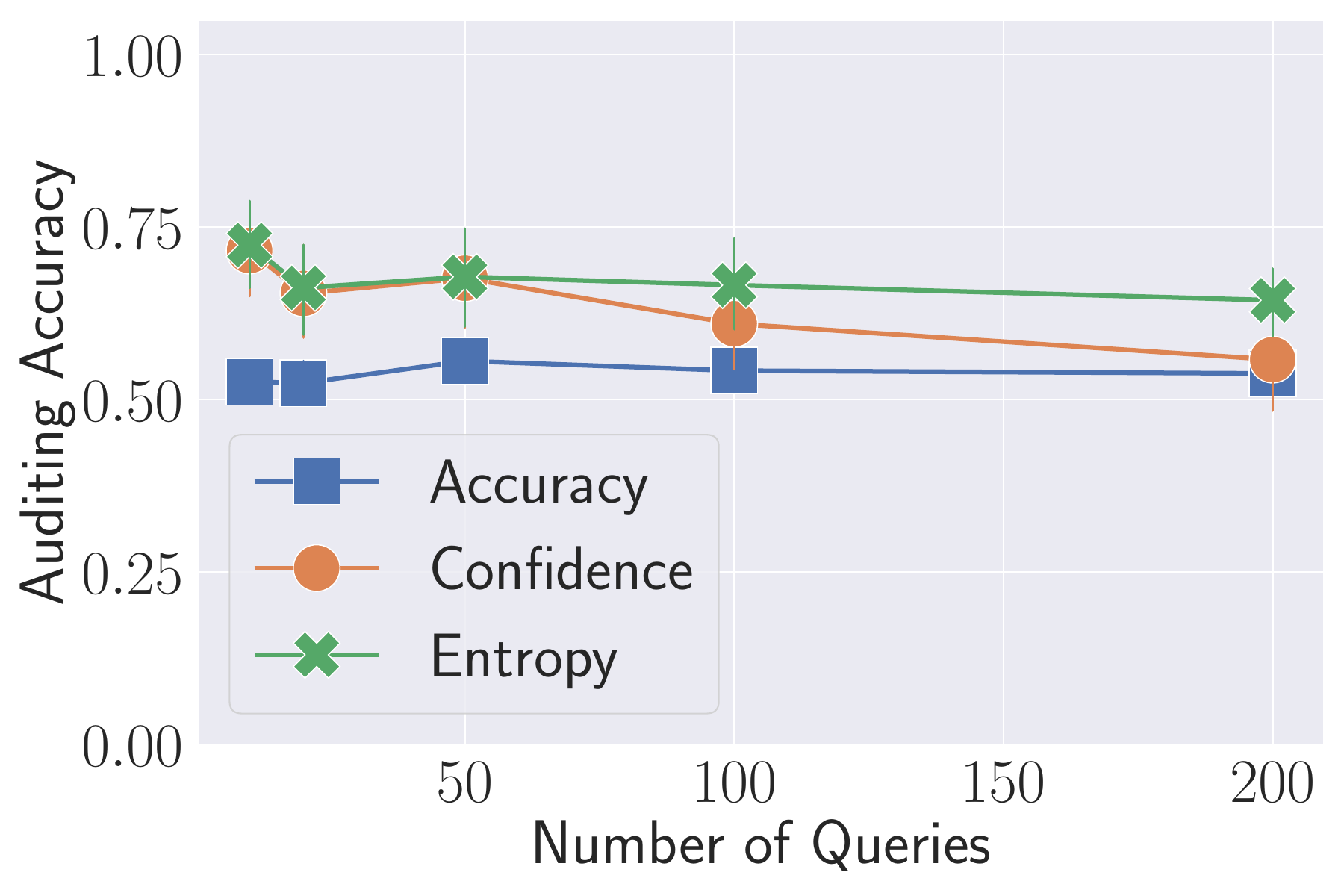}
\caption{\stwo}
\end{subfigure}
\caption{Metric-based auditing performance for target classifiers fine-tuned on DistilBERT with varying query budgets of \qr $\{10, 20, 50, 100, 200\}$ for \tthree in (a) \sone and (b) \stwo.
The source LLM is GPT-4.}
\label{figure:audit_classifier_x_query_real_gpt4_sup}
\end{figure}

\begin{figure}[ht]
\centering
\begin{subfigure}{0.45\columnwidth}
\includegraphics[width=\columnwidth]{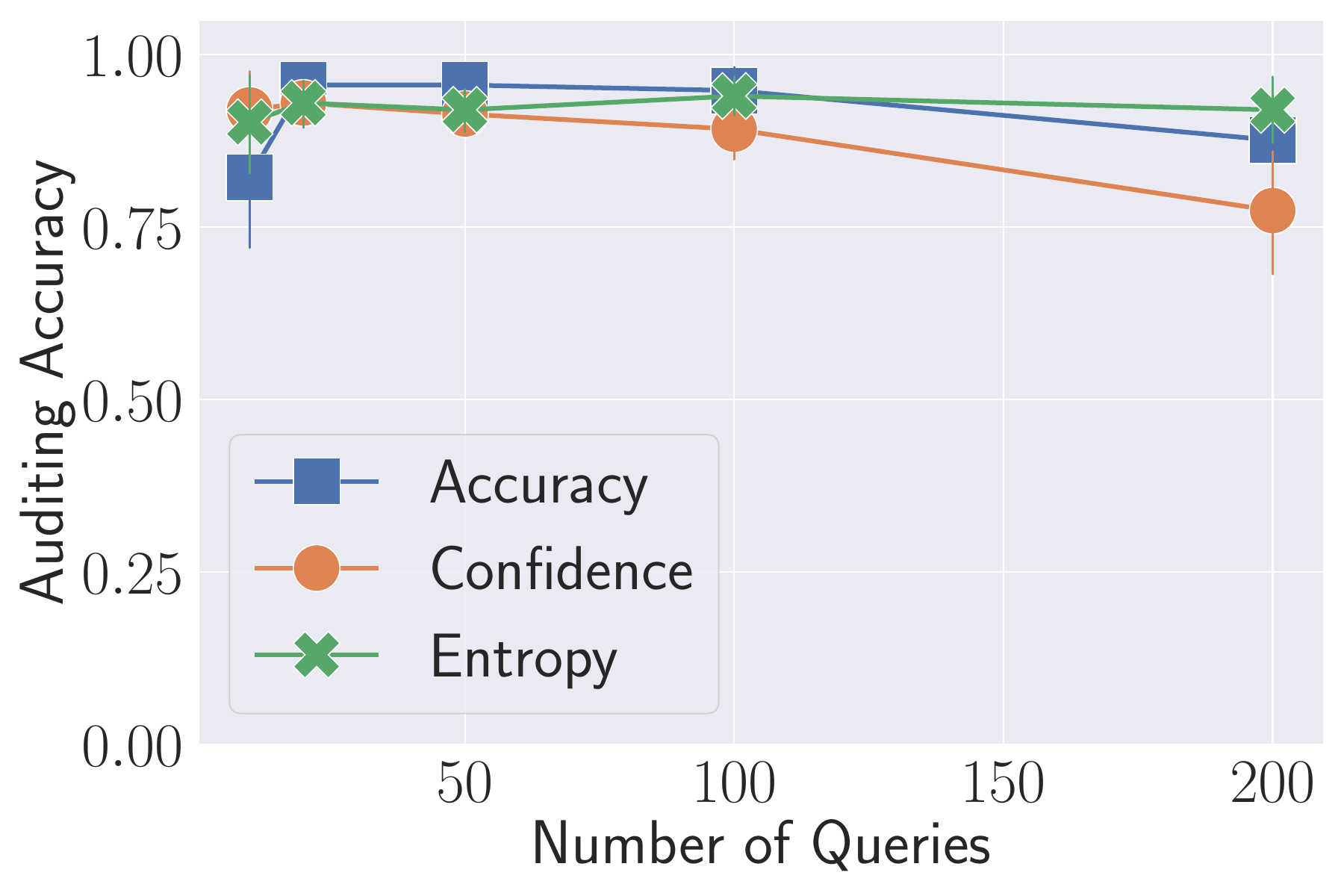}
\caption{\sone}
\end{subfigure}
\begin{subfigure}{0.45\columnwidth}
\includegraphics[width=\columnwidth]{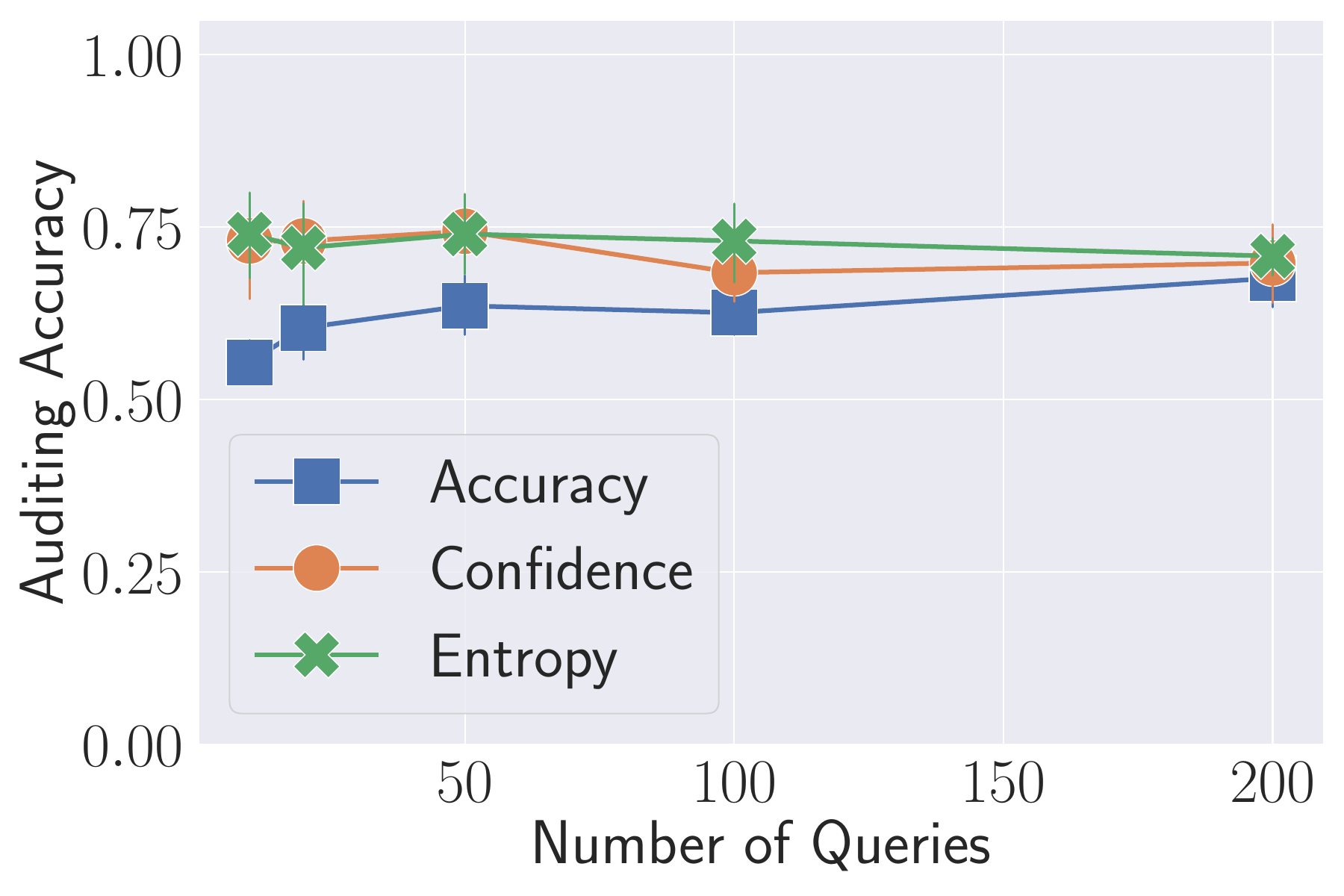}
\caption{\stwo}
\end{subfigure}
\caption{Metric-based auditing performance for target classifiers fine-tuned on DistilBERT with varying query budgets of \qr $\{10, 20, 50, 100, 200\}$ for \tthree in (a) \sone and (b) \stwo.
The source LLM is Mistral.}
\label{figure:audit_classifier_x_query_real_mistral_sup}
\end{figure}

\begin{figure}[ht]
\centering
\begin{subfigure}{0.45\columnwidth}
\includegraphics[width=\columnwidth]{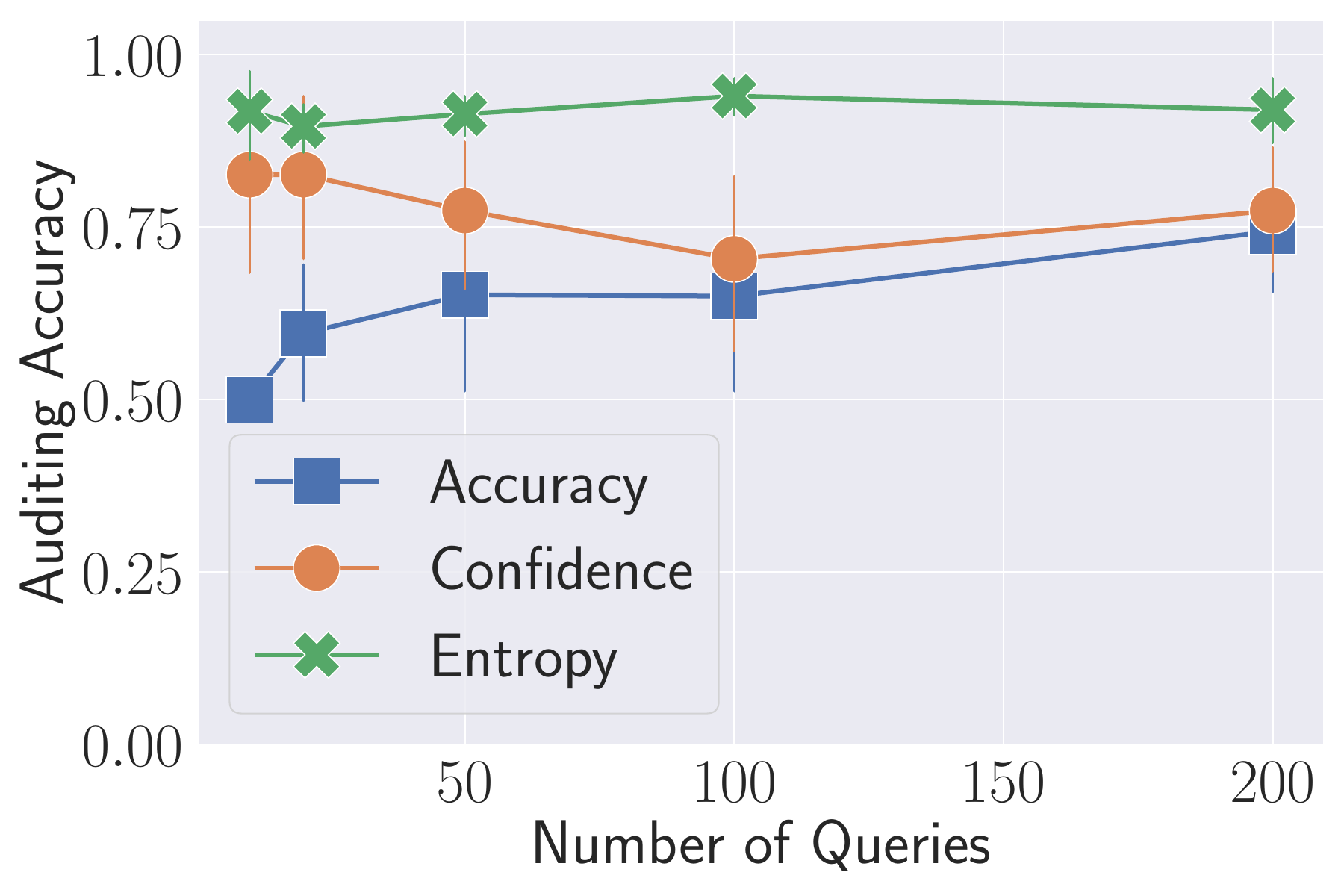}
\caption{\sone}
\end{subfigure}
\begin{subfigure}{0.45\columnwidth}
\includegraphics[width=\columnwidth]{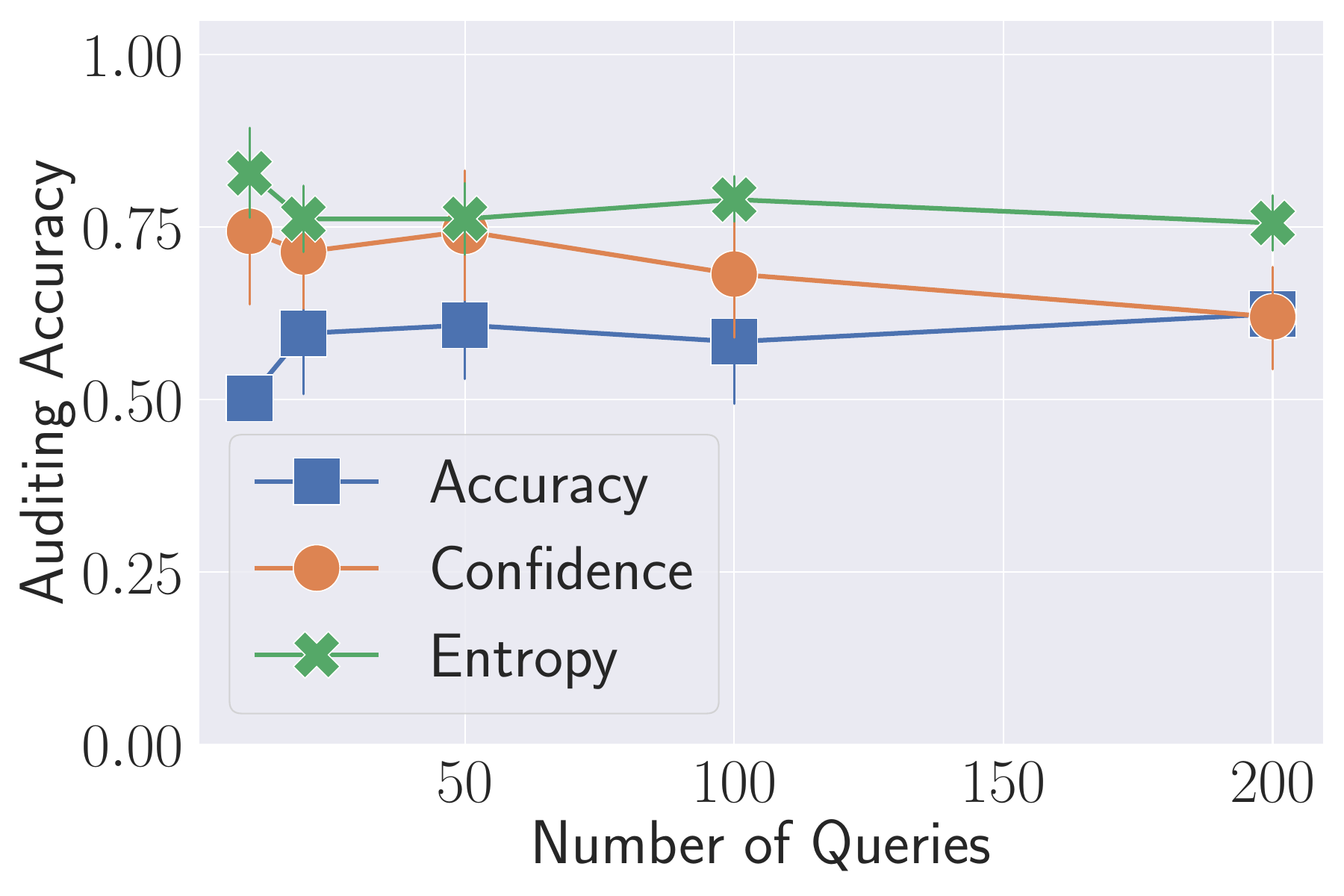}
\caption{\stwo}
\end{subfigure}
\caption{Metric-based auditing performance for target classifiers fine-tuned on DistilBERT with varying query budgets of \qr $\{10, 20, 50, 100, 200\}$ for \tthree in (a) \sone and (b) \stwo.
The source LLM is ChatGLM3.}
\label{figure:audit_classifier_x_query_real_chatglm_sup}
\end{figure}

In~\autoref{figure:audit_classifier_x_query_real_gpt4_sup},~\autoref{figure:audit_classifier_x_query_real_mistral_sup} and~\autoref{figure:audit_classifier_x_query_real_chatglm_sup}, we present the performance of the metric-based auditing with varying query budgets of \qr on other source LLMs.
We observe the same conclusion as in~\autoref{figure:audit_classifier_x_query_real}: in both scenarios, as the query budget for real data increases, the auditing performance is relatively stable.

\begin{figure}[ht]
\centering
\begin{subfigure}{0.45\columnwidth}
\includegraphics[width=\columnwidth]{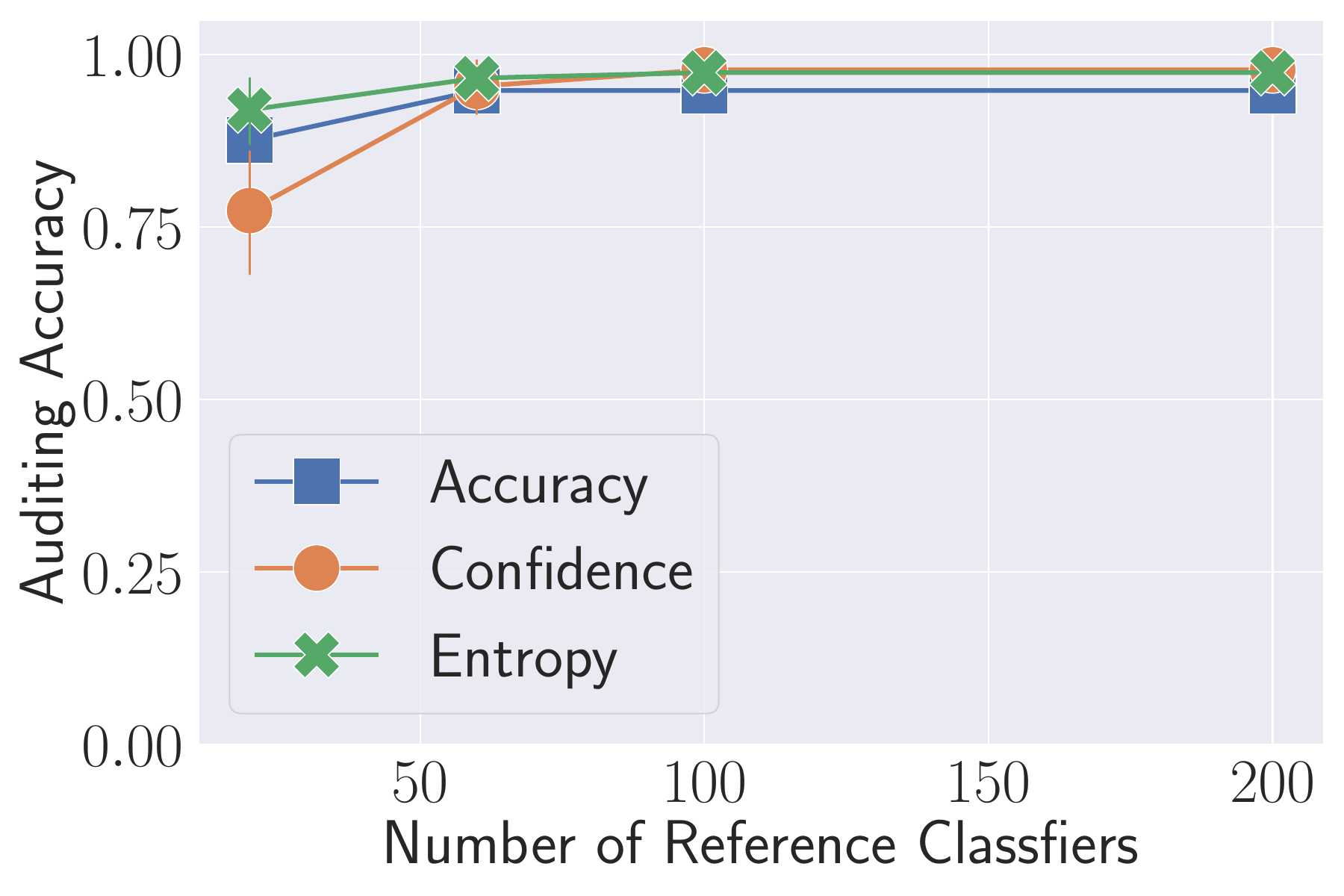}
\caption{\sone}
\end{subfigure}
\begin{subfigure}{0.45\columnwidth}
\includegraphics[width=\columnwidth]{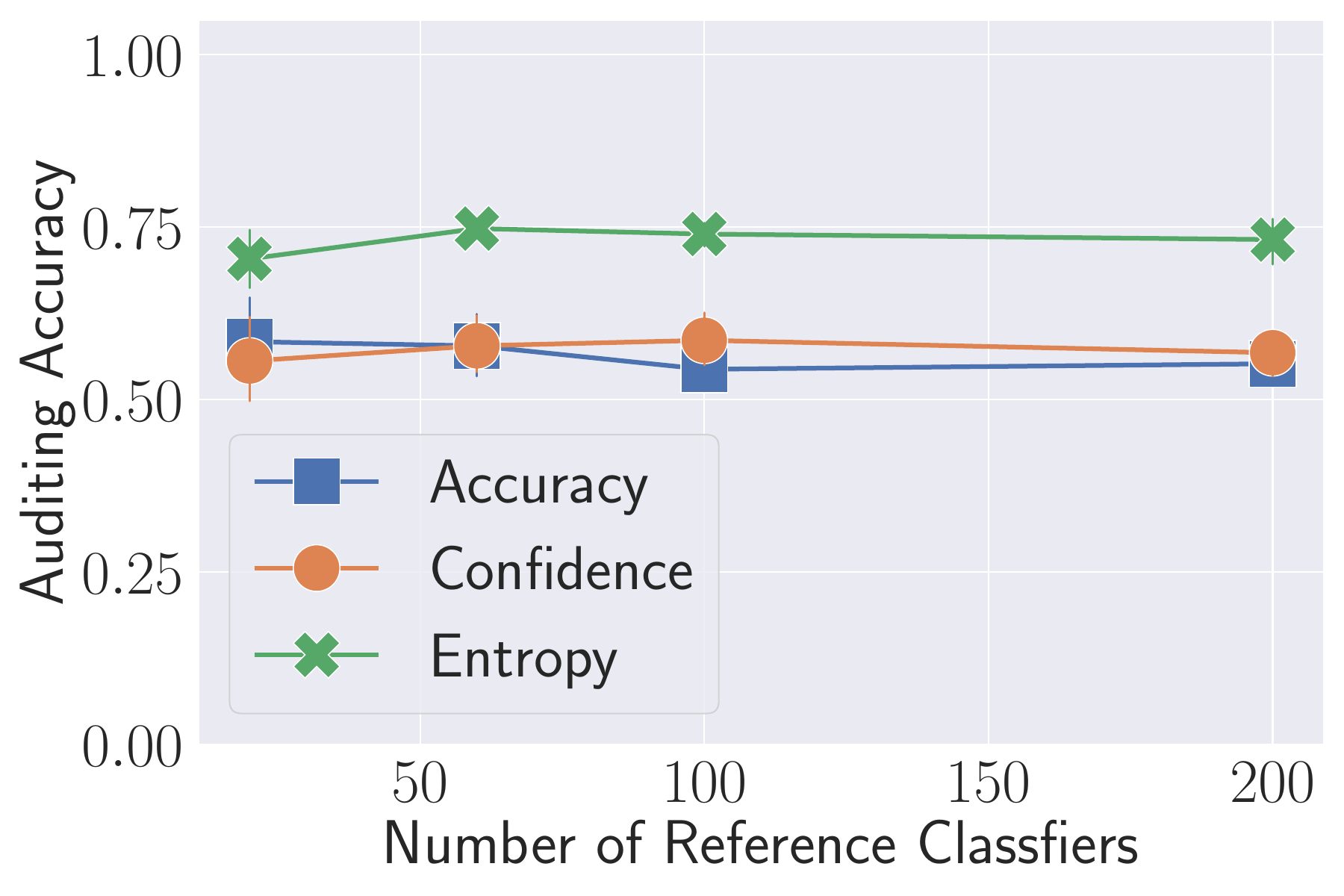}
\caption{\stwo}
\end{subfigure}
\caption{Metric-based auditing performance for target classifiers fine-tuned on DistilBERT with varying number of reference classifiers $\{20, 60,100, 200\}$ in (a) \sone and (b) \stwo for \tthree.
The query budget of \qr is set to 200.
The source LLM is GPT-3.5.}
\label{figure:audit_classifier_x_ns_real}
\end{figure}

Meanwhile, as illustrated in~\autoref{figure:audit_classifier_x_ns_real}, we demonstrate that 20 reference classifiers are sufficient to find a good threshold to achieve decent auditing performance, and the benefit of more reference classifiers for metric-based auditing with \qr is minimal.

\subsection{Additional Results of Tuning-Based Auditing Classifiers with \qt}
\label{appendix:tuning_based_audit_classifier_qt_sup}

\begin{figure}[ht]
\centering
\begin{subfigure}{0.45\columnwidth}
\includegraphics[width=\columnwidth]{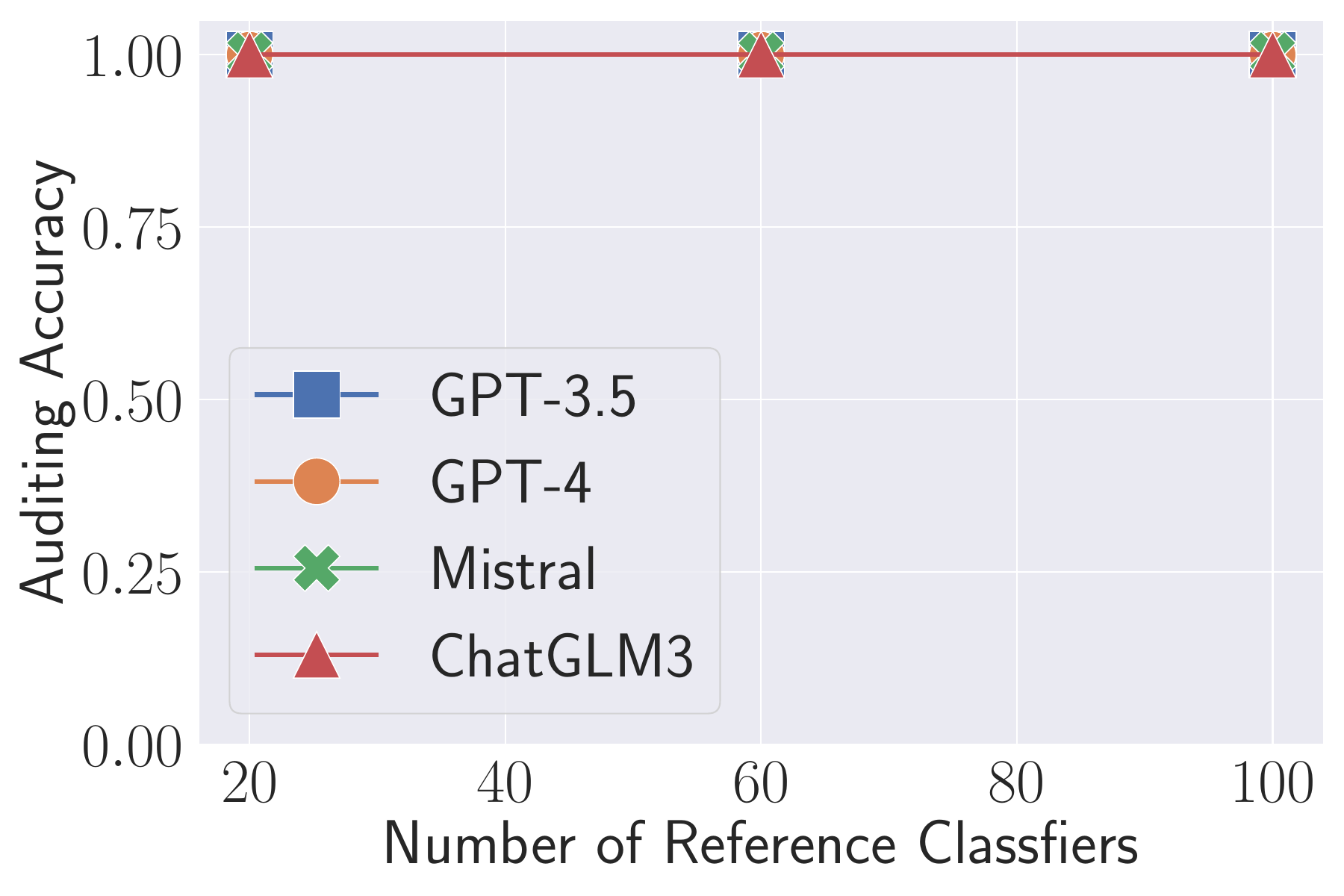}
\caption{\sone}
\end{subfigure}
\begin{subfigure}{0.45\columnwidth}
\includegraphics[width=\columnwidth]{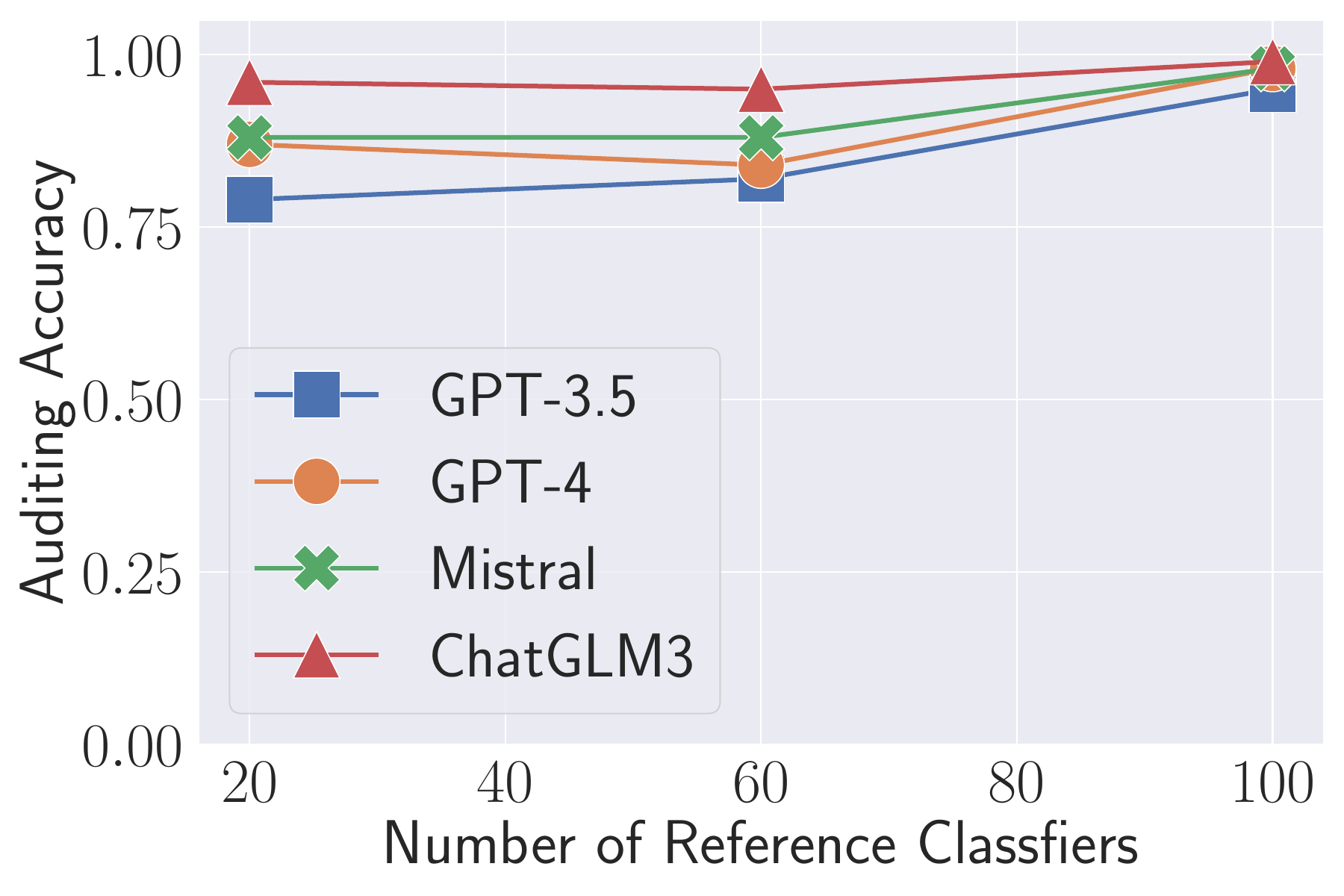}
\caption{\stwo}
\end{subfigure}
\caption{Tuning-based auditing performance for target classifiers fine-tuned on DistilBERT with varying numbers of reference classifiers $\{20, 60, 100\}$ for \tone in (a) \sone and (b) \stwo.}
\label{figure:audit_classifier_x_ns_tuning_t1}
\end{figure}

\begin{figure}[ht]
\centering
\begin{subfigure}{0.45\columnwidth}
\includegraphics[width=\columnwidth]{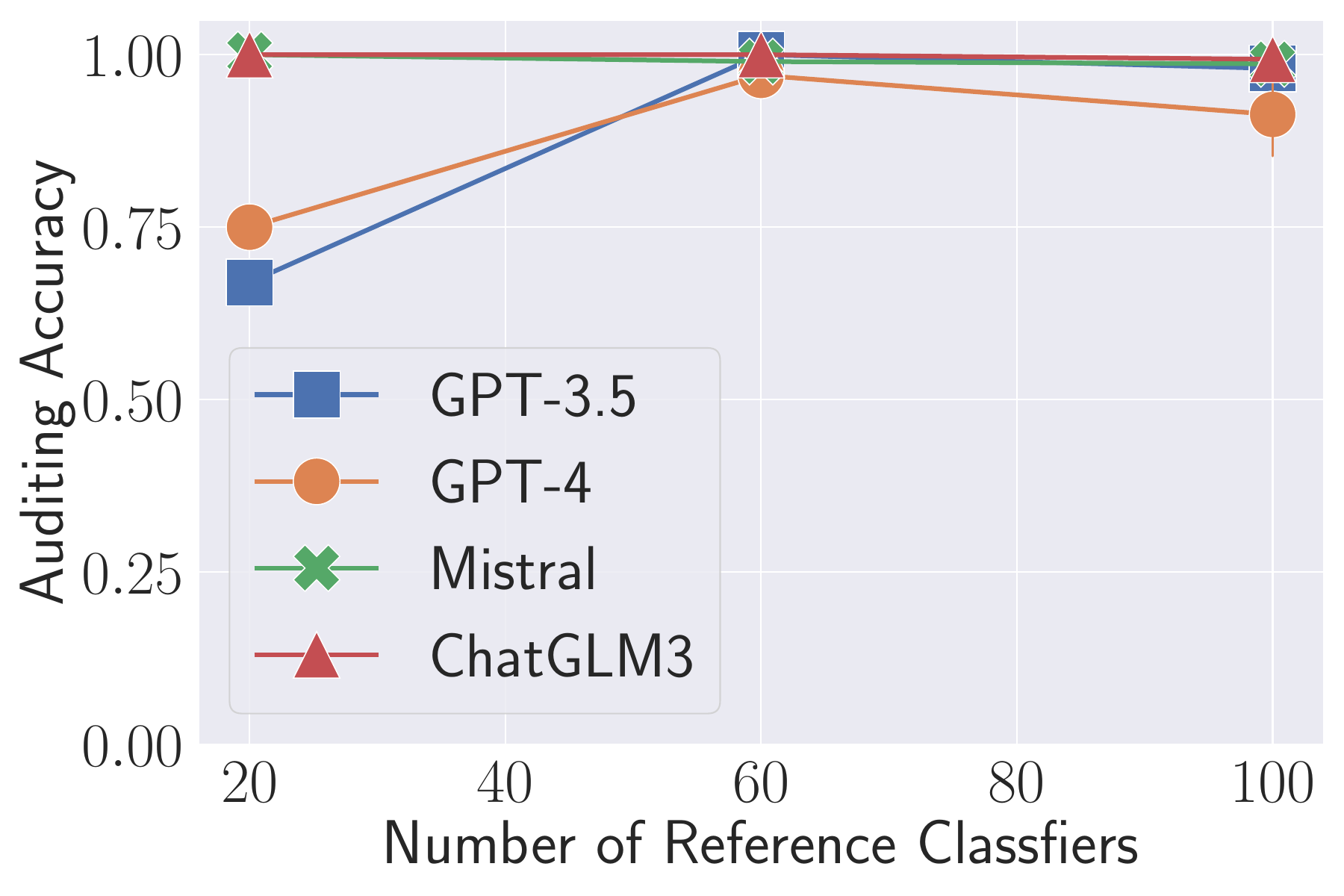}
\caption{\sone}
\end{subfigure}
\begin{subfigure}{0.45\columnwidth}
\includegraphics[width=\columnwidth]{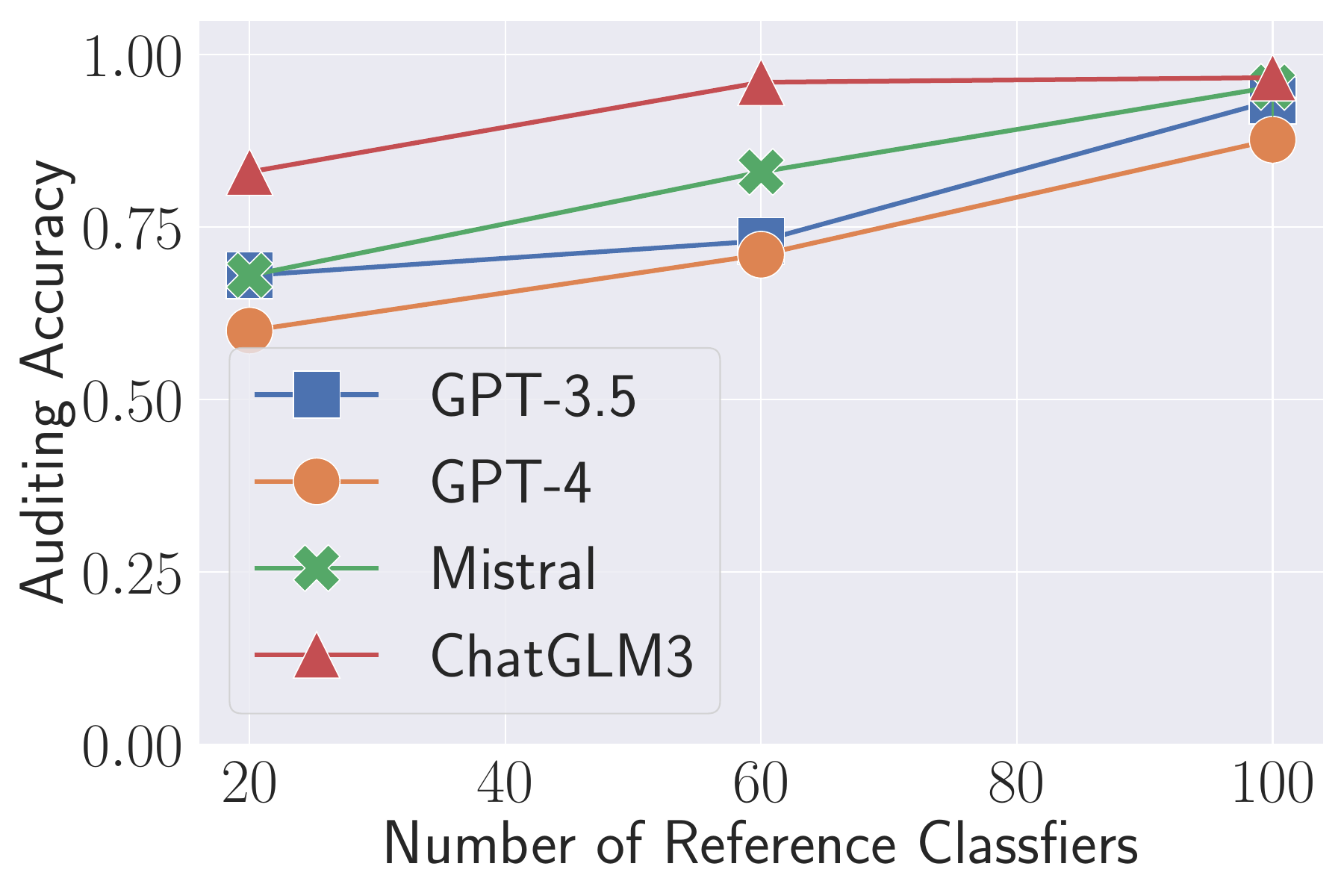}
\caption{\stwo}
\end{subfigure}
\caption{Tuning-based auditing performance for target classifiers fine-tuned on pre-trained DistilBERT with varying numbers of reference classifiers $\{20, 60, 100\}$ for \ttwo in (a) \sone and (b) \stwo.}
\label{figure:audit_classifier_x_ns_tuning_t2}
\end{figure}

\begin{figure}[ht]
\centering
\begin{subfigure}{0.45\columnwidth}
\includegraphics[width=\columnwidth]{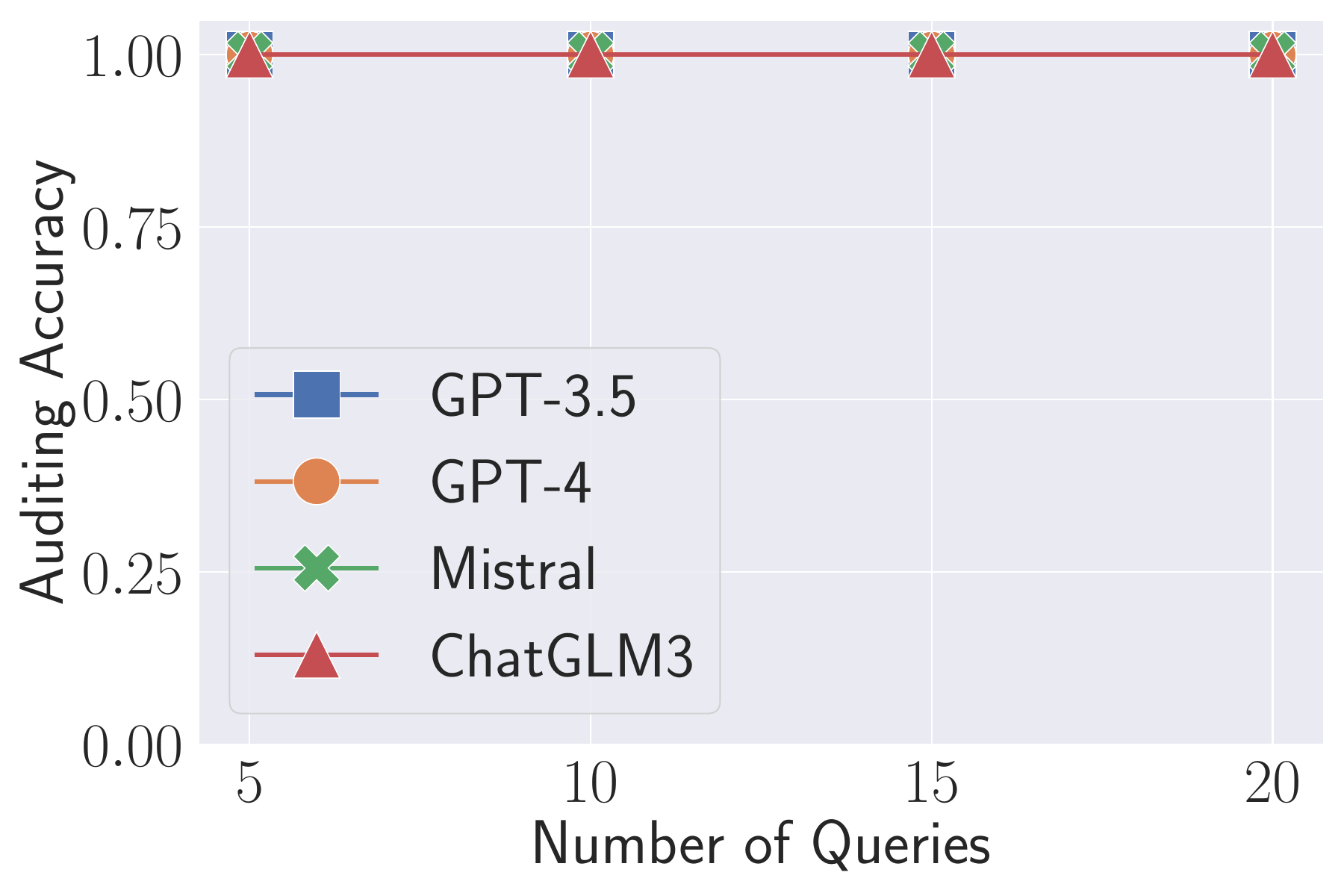}
\caption{\sone}
\end{subfigure}
\begin{subfigure}{0.45\columnwidth}
\includegraphics[width=\columnwidth]{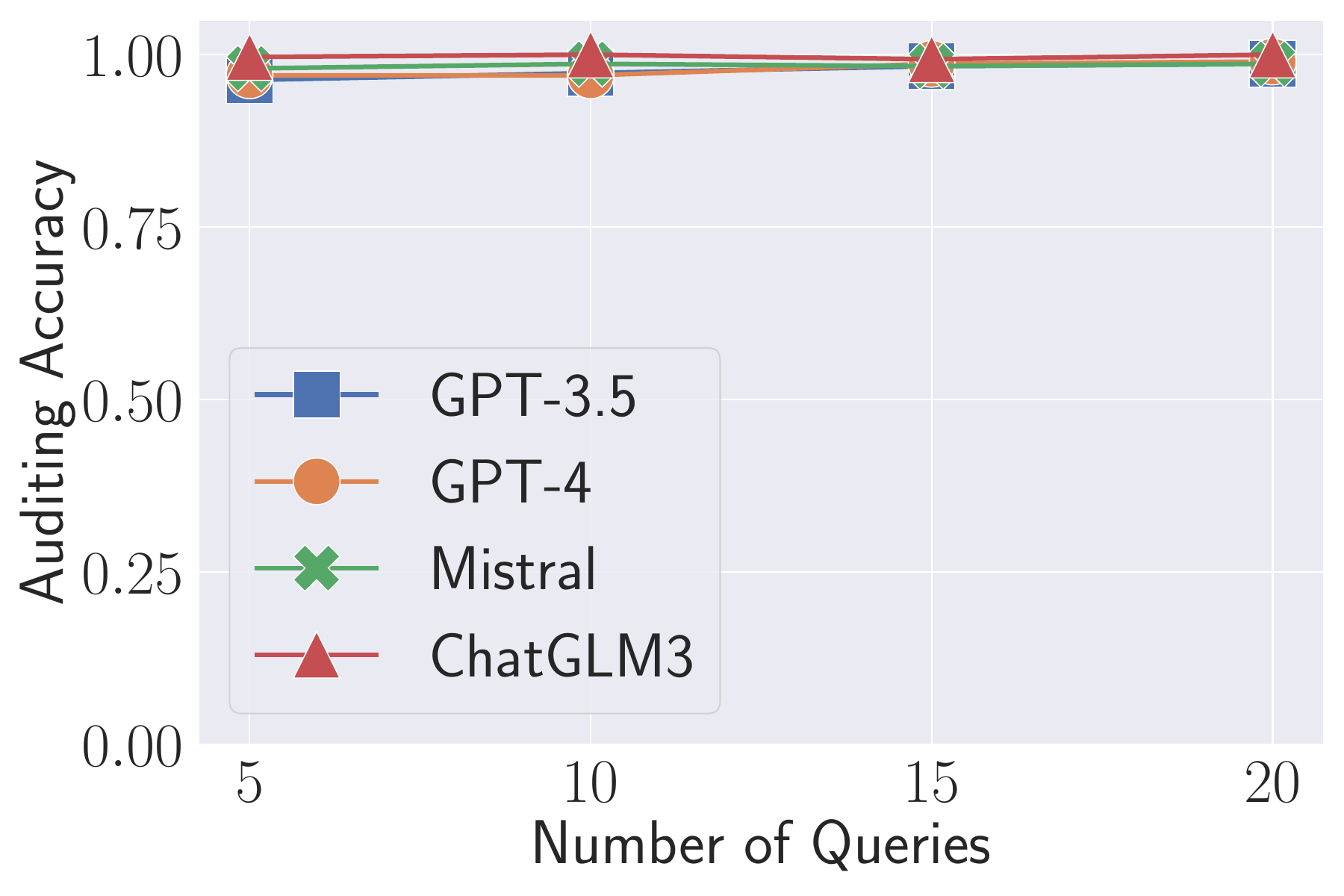}
\caption{\stwo}
\end{subfigure}
\caption{Tuning-based auditing performance for target classifiers fine-tuned on pre-trained DistilBERT with varying query budgets $\{5, 10, 15, 20\}$ for \tthree in (a) \sone and (b) \stwo.}
\label{figure:audit_classifier_x_query_tuning_t3}
\end{figure}

In~\autoref{figure:audit_classifier_x_ns_tuning_t1}, ~\autoref{figure:audit_classifier_x_ns_tuning_t2} we complement the auditing performance on other two tasks.
We learn a query set with $|\mathcal{Q}_{\phi}| = 5$.
We observe that tuning-based auditing also achieves strong performance on these two tasks, and more reference classifiers, meaning a larger training dataset leads to better tuning-based auditing performance.
Meanwhile, as illustrated in~\autoref{figure:audit_classifier_x_query_tuning_t3}, we demonstrate that five tuned queries are sufficient to achieve decent auditing performance, and the benefit of more query budgets for tuning-based auditing is minimal.

\subsection{Main Results on BERT}
\label{appendix:audit_classifier_bert_sup}

\begin{figure}[ht]
\centering
\begin{subfigure}{0.45\columnwidth}
\includegraphics[width=\columnwidth]{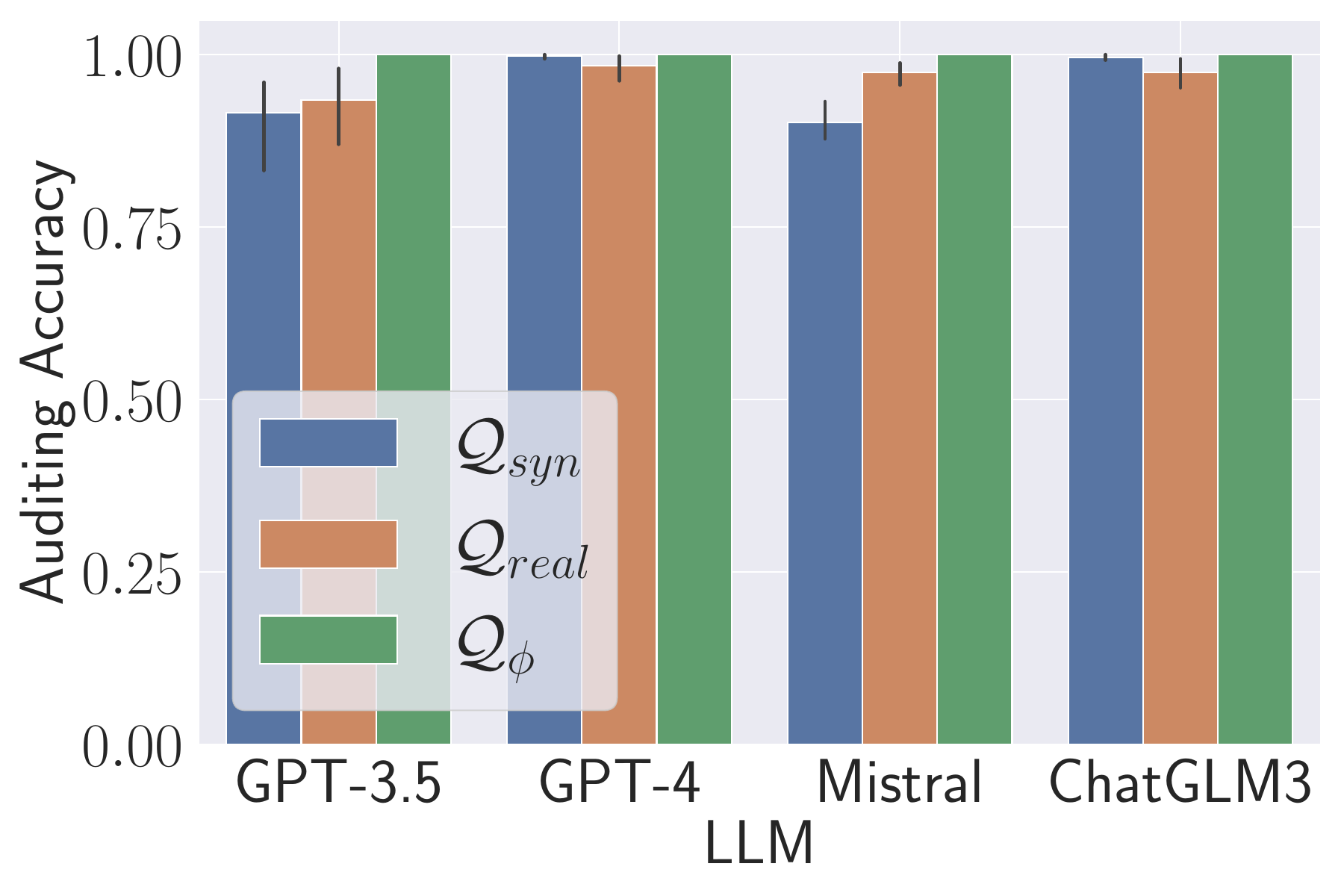}
\caption{\sone}
\end{subfigure}
\begin{subfigure}{0.45\columnwidth}
\includegraphics[width=\columnwidth]{figs/audit_classifier/barplot/s2/sentiment_analysis_distillbert_nt100.pdf}
\caption{\stwo}
\end{subfigure}
\caption{Auditing performance for target classifiers fine-tuned on BERT model in \tone.}
\label{figure:audit_classifier_barplot_t1_sup}
\end{figure}

\begin{figure}[ht]
\centering
\begin{subfigure}{0.45\columnwidth}
\includegraphics[width=\columnwidth]{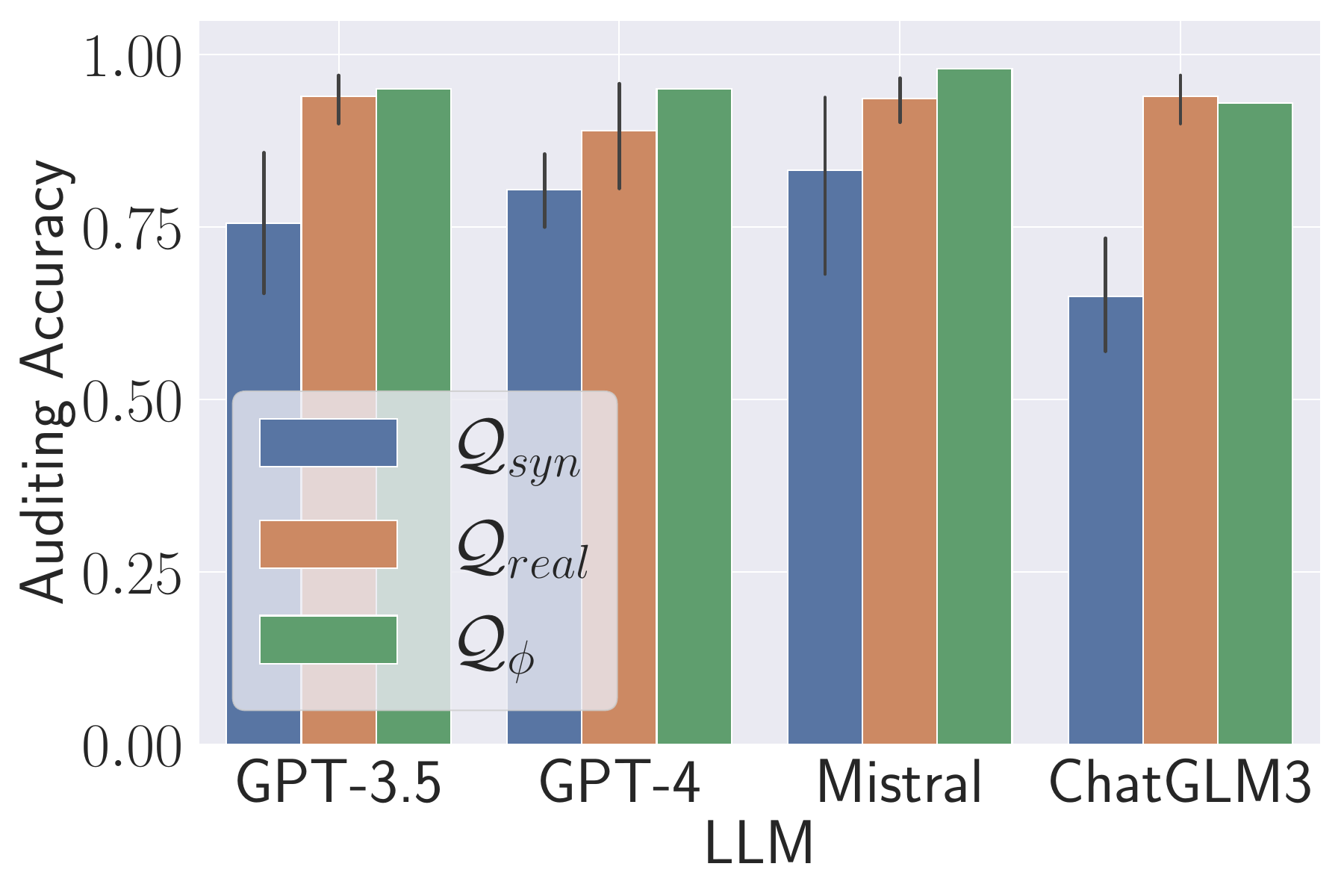}
\caption{\sone}
\end{subfigure}
\begin{subfigure}{0.45\columnwidth}
\includegraphics[width=\columnwidth]{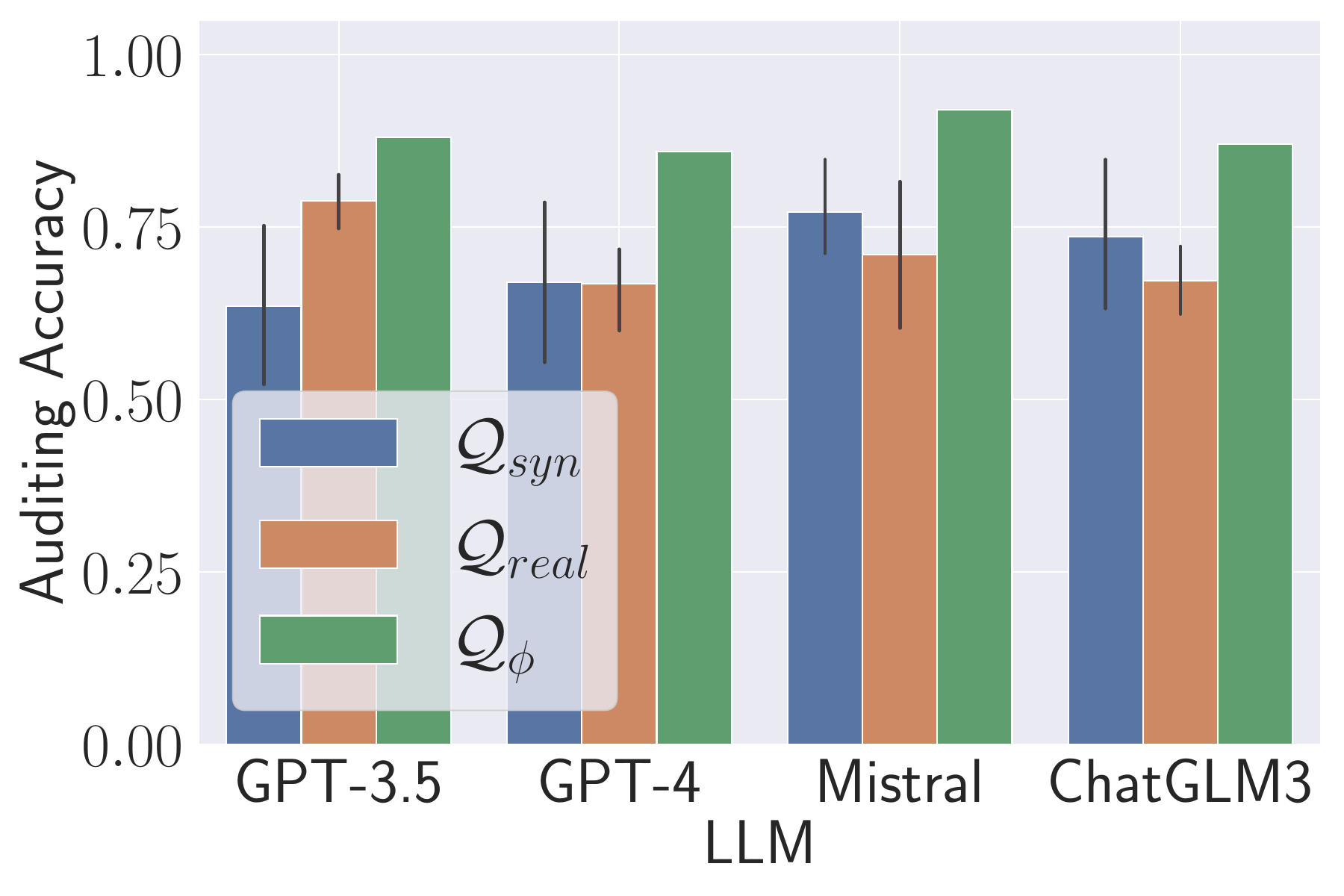}
\caption{\stwo}
\end{subfigure}
\caption{Auditing performance for target classifiers fine-tuned on BERT model in \ttwo.}
\label{figure:audit_classifier_barplot_t2_sup}
\end{figure}

\begin{figure}[ht]
\centering
\begin{subfigure}{0.45\columnwidth}
\includegraphics[width=\columnwidth]{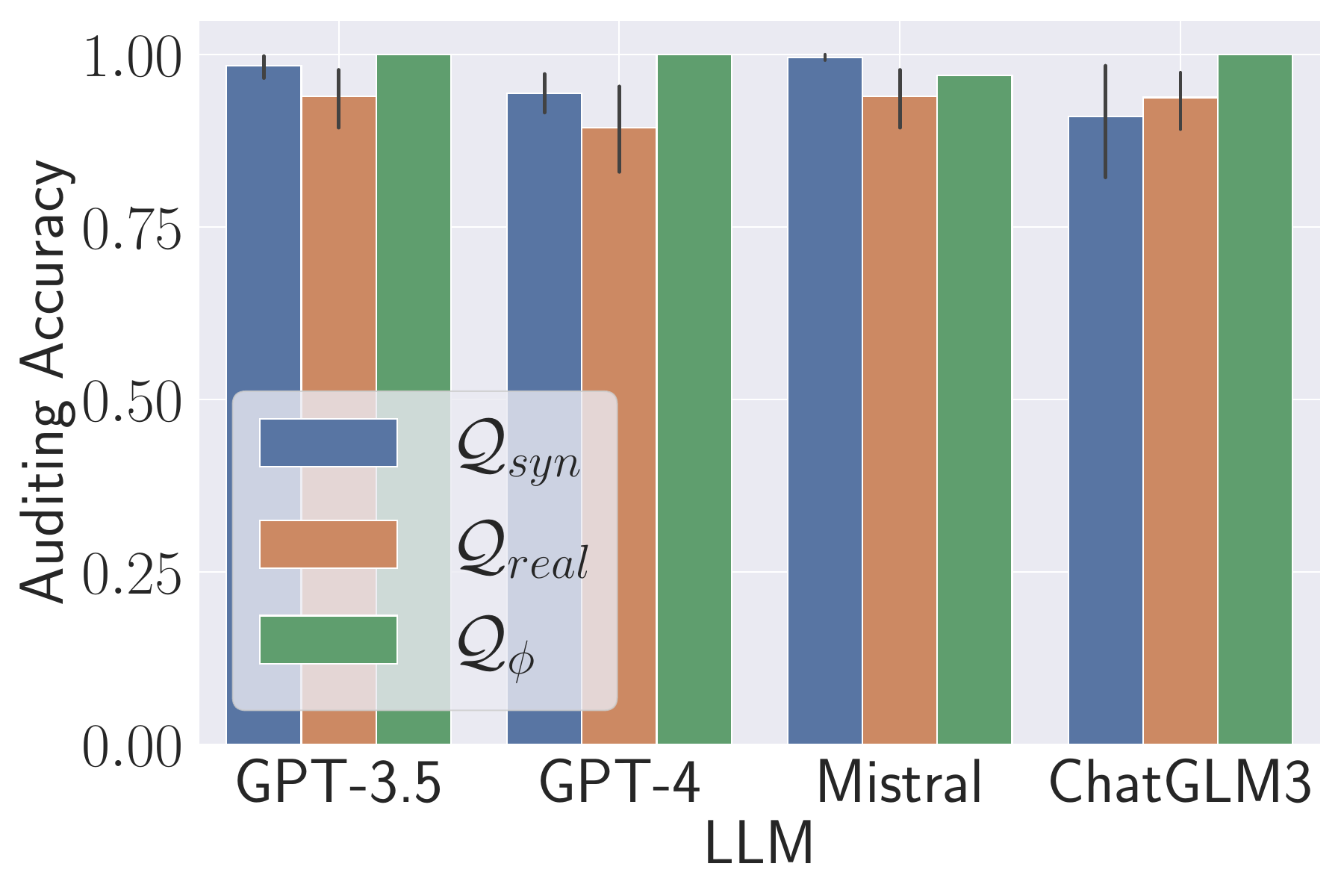}
\caption{\sone}
\end{subfigure}
\begin{subfigure}{0.45\columnwidth}
\includegraphics[width=\columnwidth]{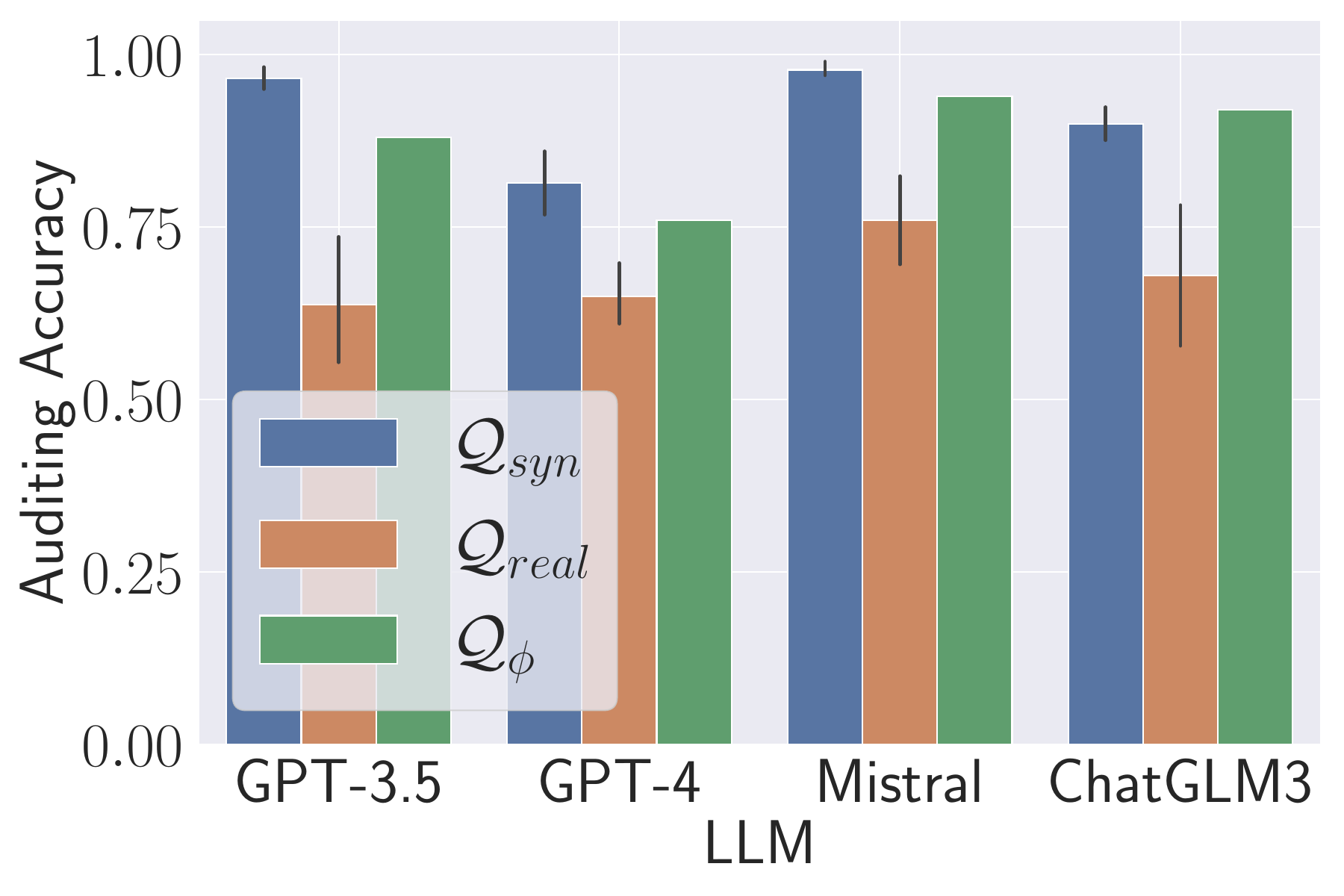}
\caption{\stwo}
\end{subfigure}
\caption{Auditing performance for target classifiers fine-tuned on BERT model in \tthree.}
\label{figure:audit_classifier_barplot_t3_sup}
\end{figure}

\begin{figure}[ht]
\centering
\includegraphics[width=0.75\columnwidth]{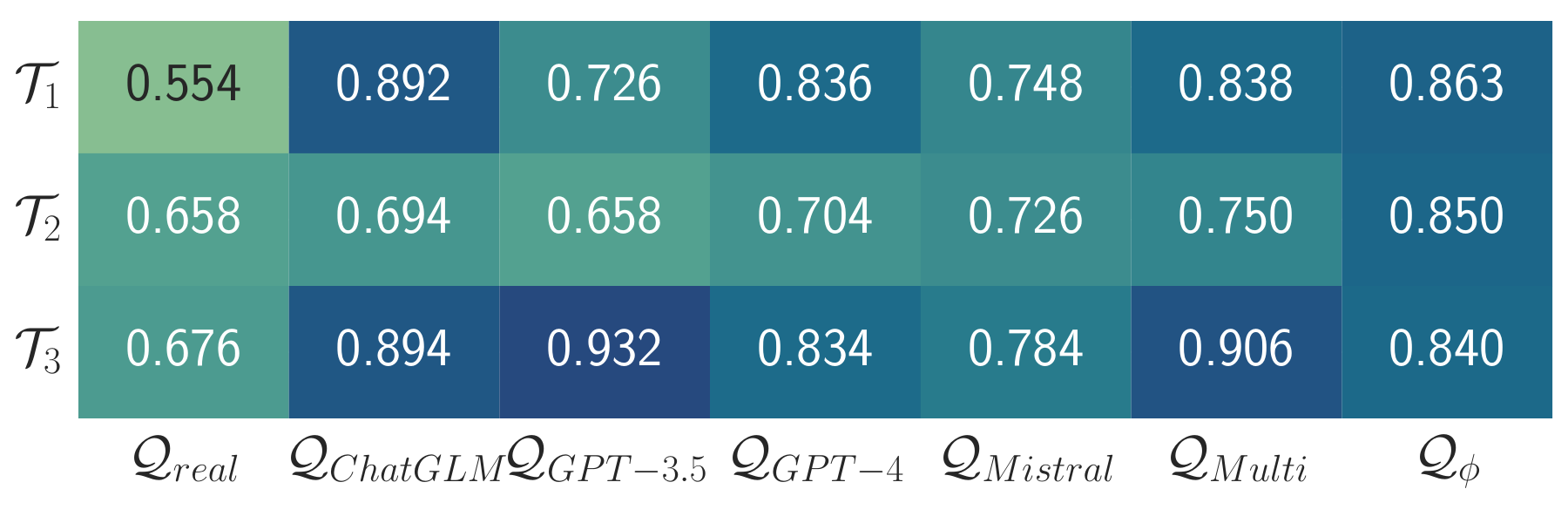}
\caption{Auditing performance for target classifiers fine-tuned on BERT using metric-based auditing with six different query sets and tuning-based auditing with \qt in \sthree.}
\label{figure:audit_classifier_heatmap_s3_sup}
\end{figure}

In~\autoref{figure:audit_classifier_barplot_t1_sup},~\autoref{figure:audit_classifier_barplot_t2_sup},~\autoref{figure:audit_classifier_barplot_t3_sup},and~\autoref{figure:audit_classifier_heatmap_s3_sup}, we complement the main results on the pre-trained BERT model, and we can conclude similar conclusions.

\section{Additional Results of Auditing Generator}
\label{appendix:audit_generator_sup}

\subsection{Target Performance of Generator}
\label{appendix:target_generator_performance}

\begin{figure*}[!t]
\centering
\begin{subfigure}{0.65\columnwidth}
\includegraphics[width=\columnwidth]{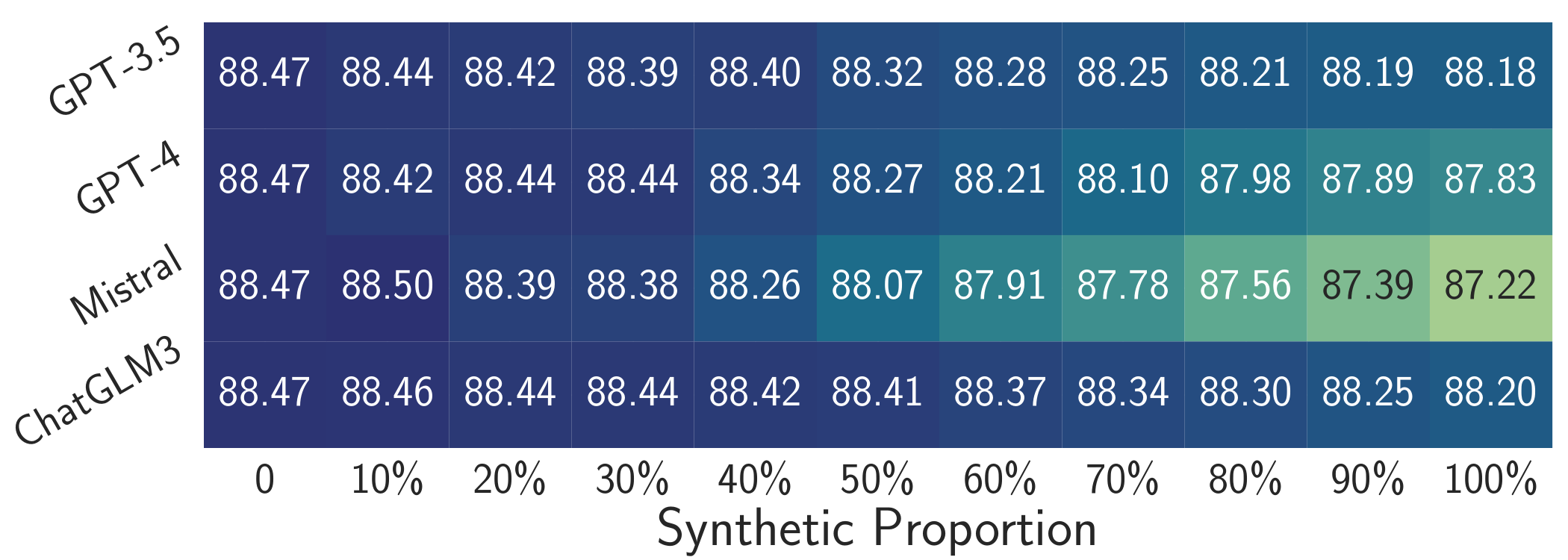}
\caption{\tfour}
\end{subfigure}
\begin{subfigure}{0.65\columnwidth}
\includegraphics[width=\columnwidth]{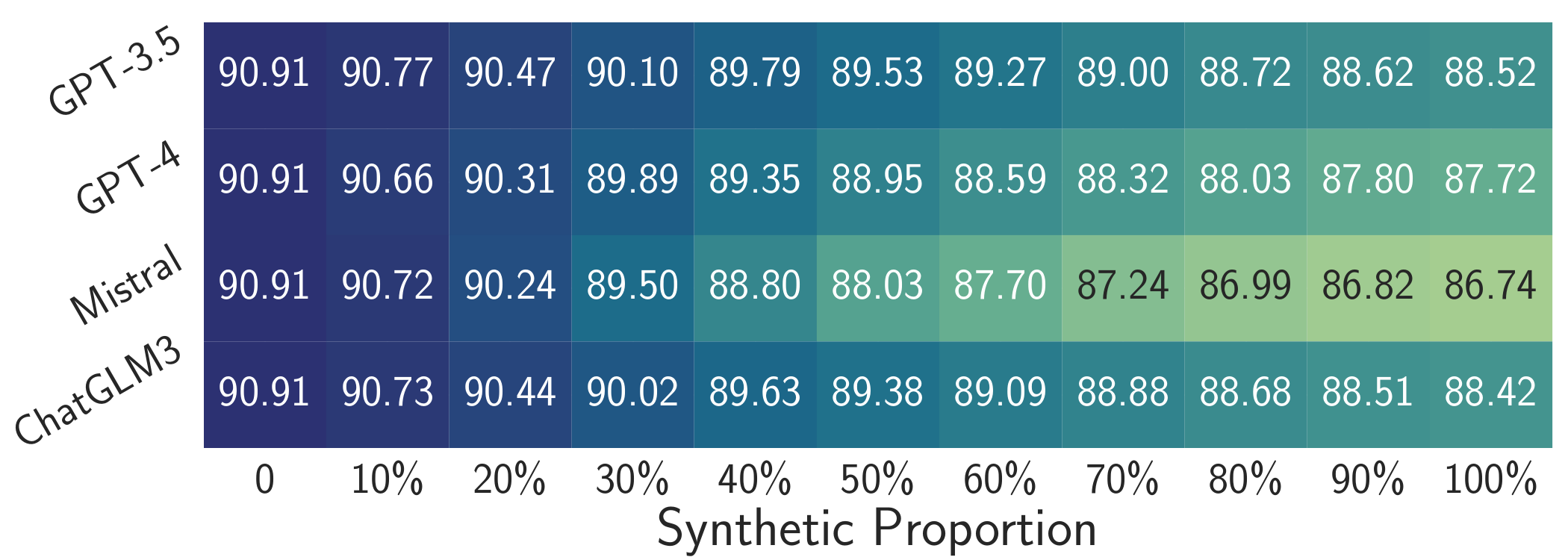}
\caption{\tfive}
\end{subfigure}
\caption{
Average target performance of real and synthetic generators fine-tuned on pre-trained BART for (a) \tfour and (b) \tfive evaluated on \dtest.}
\label{figure:target_performance_bart}
\end{figure*}

We present the average target performance of all real and synthetic generators, evaluated on \dtest using the BERTScore F1.
\dtest includes only real article and summary pairs.
As illustrated in~\autoref{figure:target_performance_bart}, synthetic generators, especially those trained on a mix of real data and synthetic data (\stwo) show comparable performance to the real synthetic generators.

\subsection{Impact of Reference Generators}
\label{appendix:reference_generator_sup}

\begin{figure}[ht]
\centering
\begin{subfigure}{0.45\columnwidth}
\includegraphics[width=\columnwidth]{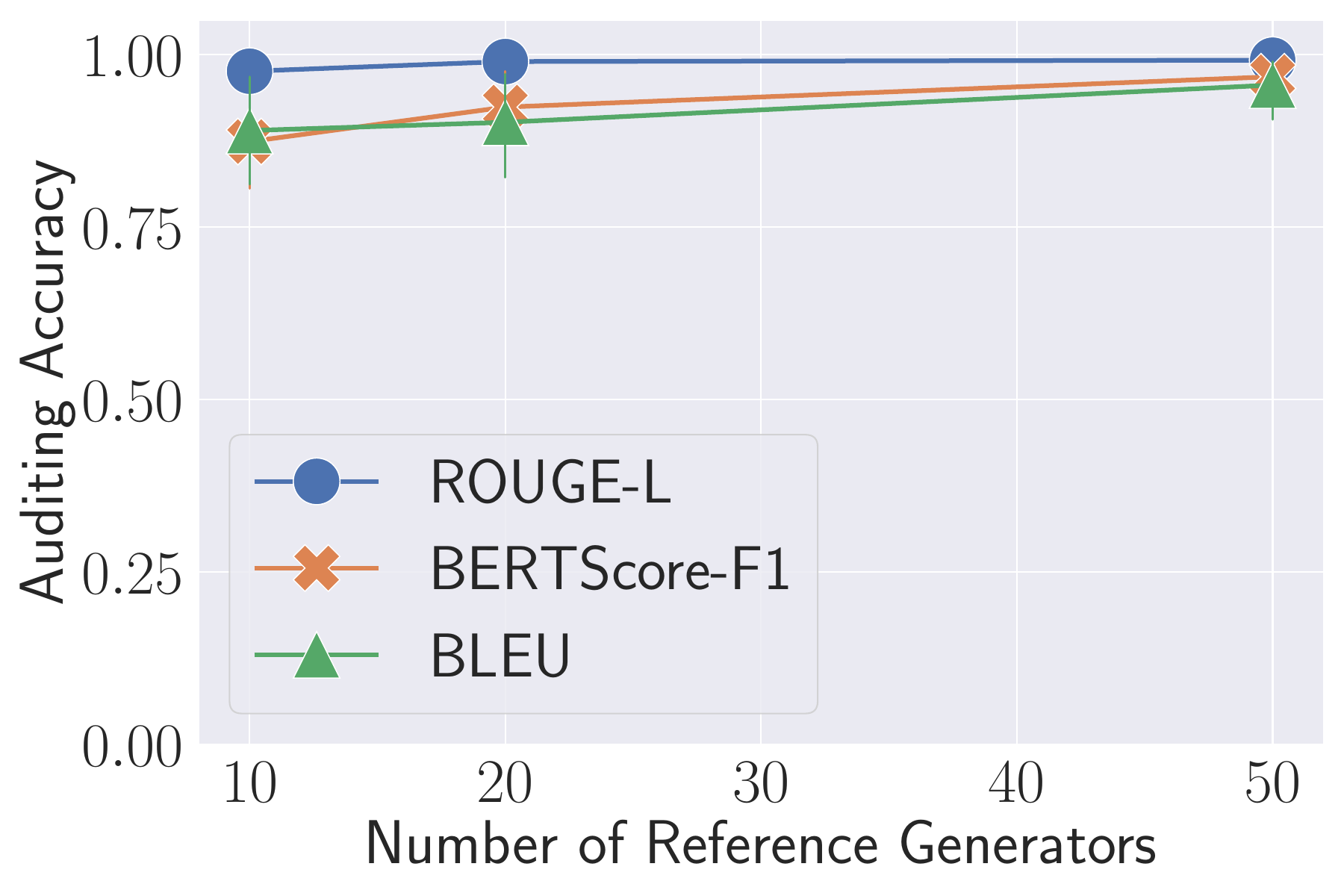}
\caption{\sone}
\end{subfigure}
\begin{subfigure}{0.45\columnwidth}
\includegraphics[width=\columnwidth]{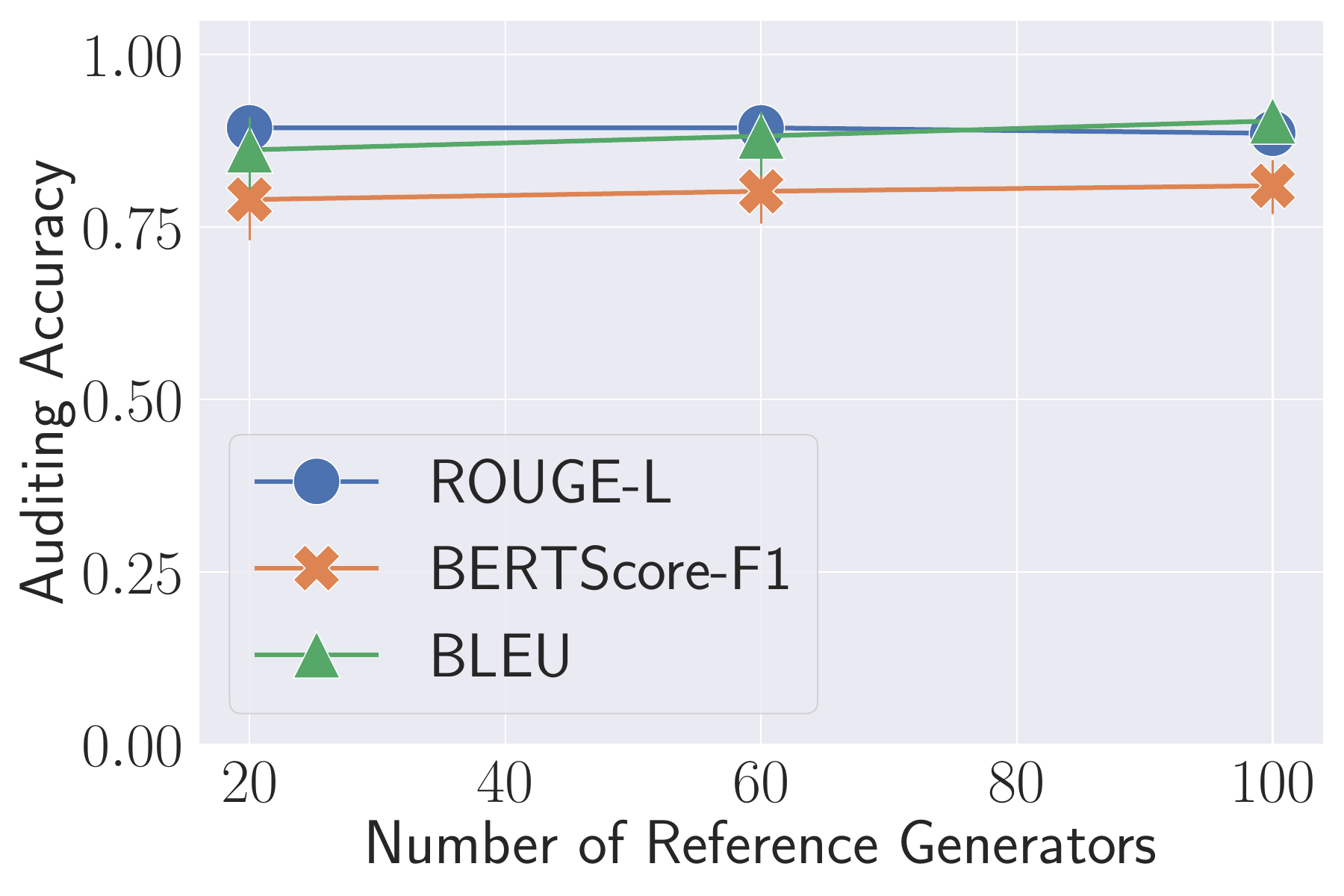}
\caption{\stwo}
\end{subfigure}
\caption{Metric-based auditing performance for target generators with varying numbers of reference generators for \tfour.
The numbers are $\{10, 20, 50\}$ in (a) \sone and $\{20, 60, 100\}$ in (b) \stwo.
The LLM is GPT-3.5.}
\label{figure:audit_generator_x_ns_sup}
\end{figure}

As illustrated in~\autoref{figure:audit_generator_x_ns_sup}, we demonstrate that 20 reference classifiers are sufficient to find a good threshold to achieve decent auditing performance, and the benefit of more reference generators for metric-based auditing with \qr is minimal.

\section{Additional Results of Auditing Plot}
\label{appendix:audit_plot_sup}

\begin{figure}[ht]
\centering
\begin{subfigure}{0.75\columnwidth}
\includegraphics[width=\columnwidth]{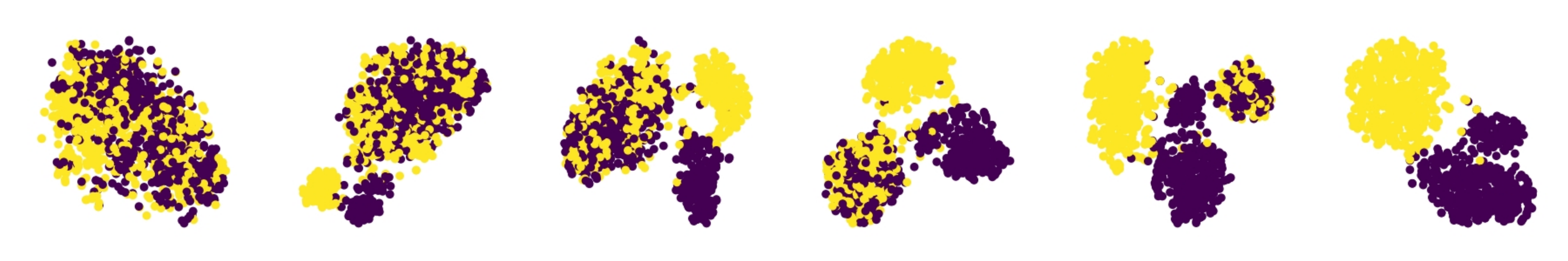}
\caption{\tsix}
\end{subfigure}
\begin{subfigure}{0.75\columnwidth}
\includegraphics[width=\columnwidth]{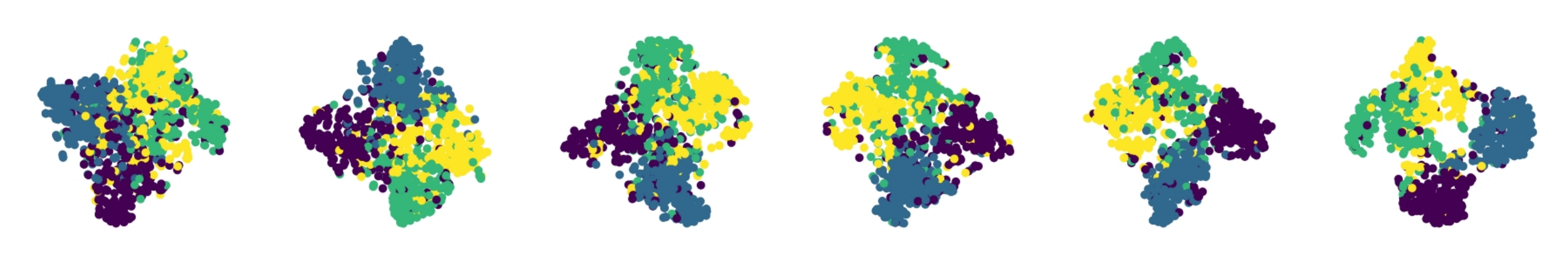}
\caption{\tseven}
\end{subfigure}
\caption{T-SNE plots using GloVe with different synthetic proportions of input data are set at intervals of 20\%, ranging from 0 to 100\% (left to right) in (a) \tsix and (b) \tseven.}
\label{figure:tsne_plot_glove}
\end{figure}

We present examples of t-SNE plots using GloVe embeddings in~\autoref{figure:tsne_plot_glove} and the auditing performance for these target plots in~\autoref{table:audit_plot_sup_glove}.
The results demonstrate the effectiveness of the proposed classification-based auditing method for t-SNE plots using different types of word embeddings.

\section{Training Details of Reference Classifiers on Rotten Tomatoes}
\label{appendix:ablation_study}

\begin{figure}[ht]
\centering
\includegraphics[width=0.75\columnwidth]{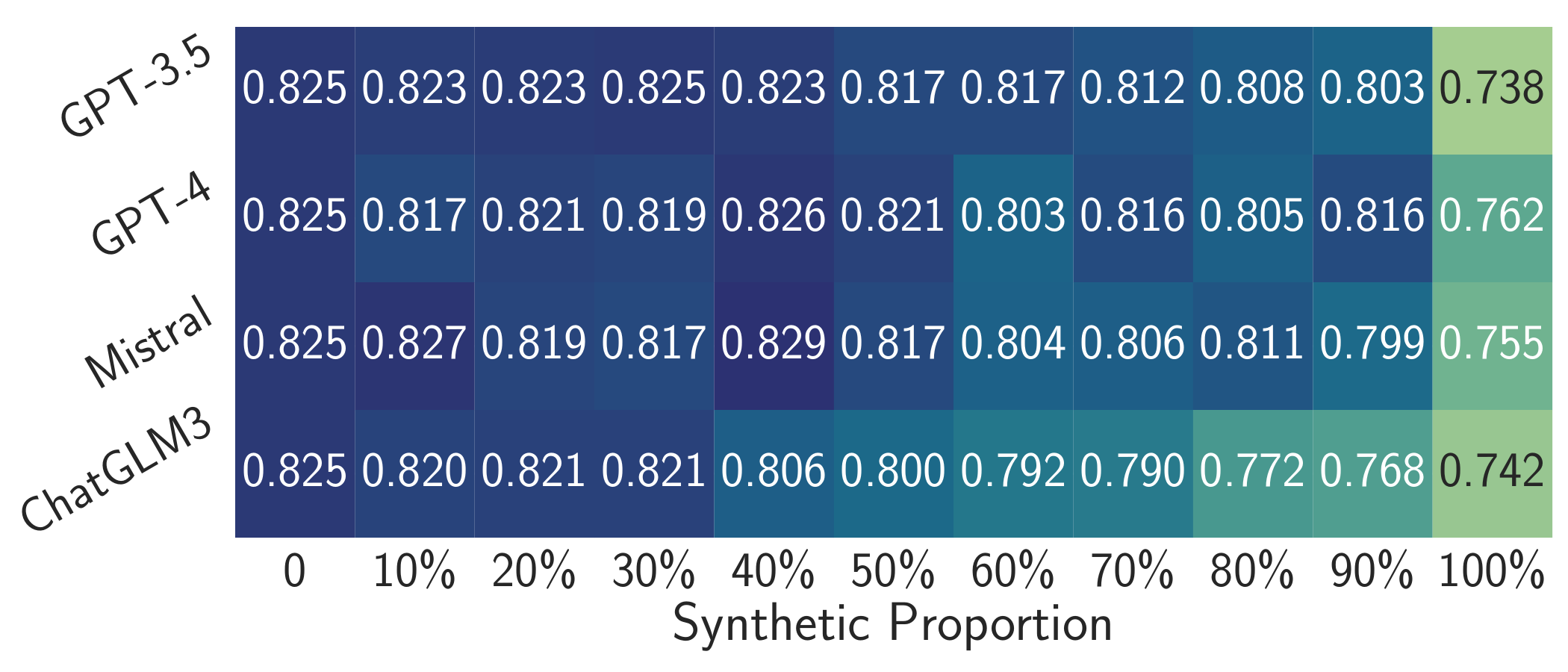}
\caption{Average classification performance of reference real and synthetic classifiers fine-tuned on pre-trained DistilBERT for \tone.
The reference real dataset is Rotten Tomatoes.}
\label{figure:tomato_performance}
\end{figure}

The Rotten Tomatoes dataset~\cite{PL05} contains 5,331 positive reviews and 5,331 negative reviews collected from the Rotten Tomatoes movie reviews.
Similar to the original \tone, we randomly sample 1,500 reviews per class using different seeds each time to train a classifier.
We show the classification performance, i.e., sentiment analysis (\tone) performance, of the reference classifiers trained on the Rotten Tomatoes dataset in~\autoref{figure:tomato_performance}.
The reference real classifiers are trained solely on the Rotten Tomatoes dataset.
The reference synthetic classifiers are trained on a mix of the Rotten Tomatoes dataset and synthetic data from different source LLMs.
The rest of the experimental settings are the same as those for \tone in~\autoref{section:classifier_eval_setup}.

\begin{table}[ht]
\caption{Auditing performance for target plots using GloVe embeddings across two tasks and four LLMs in three scenarios.}
\label{table:audit_plot_sup_glove}
\centering
\scalebox{0.65}{
\begin{tabular}{ c| c |c c c c}
\toprule
 \multirow{2}{*}{Scenario}  & \multirow{2}{*}{Task} & \multicolumn{4}{c}{LLMs} \\
 \cline{3-6}
  & & GPT-3.5 & GPT-4  & Mistral & ChatGLM3 \\
\midrule
 \multirow{2}{*}{\sone}  &  \tsix & $1.000 \pm 0.000$ &     $1.000 \pm 0.000$ & $1.000 \pm 0.000$ & $1.000 \pm 0.000$ \\
 &   \tseven & $1.000 \pm 0.000$ & $0.892 \pm 0.009$ & $1.000 \pm 0.000$ & $1.000 \pm 0.000$   \\
  \midrule
  \multirow{2}{*}{\stwo}  &  \tsix &    $0.999 \pm 0.001$ &     $0.999 \pm 0.001$ & $0.989 \pm 0.004$ &     $0.856 \pm 0.015$   \\
 &   \tseven &    $0.854 \pm 0.009$ &  $0.791 \pm 0.023$ & $0.883 \pm 0.009$ &     $0.935 \pm 0.005$  \\
 \midrule
  \multirow{2}{*}{\sthree}  &  \tsix &  \multicolumn{4}{c}{$0.936 \pm 0.004$}   \\
 &   \tseven &   \multicolumn{4}{c}{$0.781 \pm	0.008$}  \\
\bottomrule
\end{tabular}}
\end{table}

\clearpage

\begin{table*}[ht]
\caption{The prompts of synthetic data generation for \tone.}
\label{table:prompt_generation_sentiment_analysis}
\centering
\scalebox{0.65}{
\begin{tabular}{c | p{1.8\columnwidth}}
\toprule
\textbf{Target (\dts)} & Write a \{label\} review for the movie \{title\}, referring to the its outline: \{outline\} \\
\midrule
\textbf{Reference (\drs \& \qs)} &  Imagine you've just watched a movie titled `\{title\}.' Write a \{label\} review for it, referring to its outline: \{outline\} \\
\bottomrule
\end{tabular}}
\end{table*}

\begin{table*}[ht]
\caption{The prompts of synthetic data generation for \ttwo.}
\label{table:prompt_generation_topic_classification}
\centering
\scalebox{0.65}{
\begin{tabular}{c | p{1.8\columnwidth}}
\toprule
\multirow{3}{*}{\centering \textbf{Target (\dts)}} & Now you are a journalist writing news articles. You will be provided a reference article, and your task is to paraphrase the given article into a new news report on the topic of \{class\_dict[label]\}. You must be as creative as possible to diversify the new news article. Reference message: \{reference\_sample\} \\
\midrule
\multirow{4}{*}{\centering \textbf{Reference (\drs \& \qs)}} & You are now a journalist tasked with writing news articles. Given a specific topic and a reference news article, your job is to paraphrase the provided article into a new piece on the same topic \{class\_dict[label]\}. Please ensure that your news article is creative and unique. Reference article: \{reference\_sample\}. Remember to maintain the factual accuracy of the original article while infusing your unique perspective and style. \\
\bottomrule
\end{tabular}}
\end{table*}

\begin{table*}[ht]
\caption{The prompts of synthetic data generation for \tthree.}
\label{table:prompt_generation_spam_detection}
\centering
\scalebox{0.65}{
\begin{tabular}{c | c | p{1.6\columnwidth}}
\toprule
\multirow{14}{*}{ \textbf{Label = 0}} & \multirow{7}{*}{\centering \textbf{Target (\dts)}}  &  Now you are a person who is planning to send an email message. You will be given a reference message, and you need to paraphrase the reference message to a new email message. You must be as creative as possible to diversify the new message. Ensure your language is conversational and colloquial. Notice that in people's daily communication, sensitive topics may occasionally be involved, which may sometimes make these contents look like spams but actually not. You also need to imitate these contents. Now write a new email message based on the reference message as I required. Be creative and write a unique email message. Reference message: \{reference\_sample\} \\
\cline{2-3}
& \multirow{7}{*}{\centering \textbf{Reference (\drs \& \qs)}} & Imagine yourself about to craft an email. A reference message will be shared with you, and your task is to rewrite this message into a brand-new email. Your creativity will be key in adding variety to this new piece. Aim for a tone that's informal and friendly, mimicking everyday chat. It's important to remember that while everyday discussions might occasionally touch on delicate matters, making them seem spam-like, they're genuinely not. Your challenge includes echoing these nuances. Now, as requested, create an inventive and distinct email based on the provided sample. Let your creativity flow. Here's the message to start with: \{reference\_sample\} \\
\midrule
\multirow{13}{*}{ \textbf{Label = 1}} & \multirow{7}{*}{\centering \textbf{Target (\dts)}}   & Now you are a person who is planning to send a spam email message. You will be given a reference message, and you need to paraphrase the reference message to a new email message. You must be as creative as possible to diversify your messages. Ensure your language is conversational and colloquial. Notice that scammers, in order to make people believe them, will make their spam email messages look like people's daily conversations or very formal and serious content. You also need to imitate these contents. Now write a new email message based on the reference message as I required. Be creative and write a unique email message. Reference message: \{reference\_sample\} \\
\cline{2-3}
& \multirow{6}{*}{\centering \textbf{Reference (\drs \& \qs)}} &  Imagine yourself about to craft a spam email. A sample message will be shared with you, and your task is to rewrite this message into a brand-new email. Your creativity will be key in adding variety to this new piece. Aim for a tone that's informal and friendly, mimicking everyday chat. Note that scammers often design their spam emails to mimic everyday chatter or adopt a very formal and grave tone. Your task involves replicating this style. Now, as requested, create an inventive and distinct spam email based on the reference sample. Let your creativity flow. Here's the message to start with: \{reference\_sample\}\\
\bottomrule
\end{tabular}}
\end{table*}

\begin{table*}[ht]
\caption{The prompts of synthetic data generation for \tfour.}
\label{table:prompt_generation_text_summary_cnn}
\centering
\scalebox{0.65}{
\begin{tabular}{c | p{1.8\columnwidth}}
\toprule
\multirow{7}{*}{\centering \textbf{Target (\dts)}} & Below is an article followed by a reference summary. Your task is to generate a new summary of the article that captures all the essential information, themes, and insights. Your summary should be similar in performance to the reference summary, meaning it should be equally informative, concise, and capture the article's main points. However, it's crucial that your summary is unique and diverse in its wording and structure compared to the reference summary. Avoid repeating phrases or structuring your summary in the same way as the reference. Aim for originality in your expression while maintaining accuracy and succinctness. Additionally, your summary must match the length requirement. Article: \{article\} and the length requirement is \{num\_words\} words. Reference Summary: \{ref\_highlights\} \\
\midrule
\multirow{7}{*}{\centering \textbf{Reference (\drs \& \qs)}} & Enclosed is an article along with a reference summary. Your objective is to craft a new summary that encapsulates all vital information, themes, and insights from the article. This summary should match the reference in informativeness, conciseness, and coverage of key points. However, it is vital that your summary remains distinct in its wording and structure, steering clear of repeating any phrases or mimicking the structure found in the reference. Strive for a fresh and original expression while ensuring the summary remains accurate and succinct. Additionally, your summary must conform to the specified word count. Article: \{article\} and the required word count is \{num\_words\} words. Reference Summary: \{ref\_highlights\} \\
\bottomrule
\end{tabular}}
\end{table*}

\begin{table*}[ht]
\caption{The prompts of synthetic data generation for \tfive.}
\label{table:prompt_generation_text_summary_xsum}
\centering
\scalebox{0.65}{
\begin{tabular}{c | p{1.8\columnwidth}}
\toprule
\multirow{7}{*}{\centering \textbf{Target (\dts)}} & Below is an article followed by a reference summary. Your task is to generate a one-sentence summary of the article that captures all the essential information, themes, and insights. Your summary should be similar in performance to the reference summary, meaning it should be equally informative, concise, and capture the article's main points. However, it's crucial that your summary is unique and diverse in its wording and structure compared to the reference summary. Avoid repeating phrases or structuring your summary in the same way as the reference. Aim for originality in your expression while maintaining accuracy and succinctness. Additionally, your summary must match the length requirement. Article: \{article\} and the length requirement is \{num\_words\} words. Reference Summary: \{ref\_highlights\} \\
\midrule
\multirow{6}{*}{\centering \textbf{Reference (\drs \& \qs)}} & Provided below is an article along with a summary for reference. Your task is to create a one-sentence summary of the article that effectively encapsulates all the key information, themes, and insights. Ensure your summary is as informative and concise as the reference, but distinctly different in wording and structure. Strive for originality in your expression while still accurately and succinctly capturing the main points of the article. It is also essential that your summary adheres to the specified word count. Article: \{article\} and the word count requirement is \{num\_words\} words. Reference Summary: \{ref\_highlights\} \\
\bottomrule
\end{tabular}}
\end{table*}

\end{document}